\documentclass[twoside,11pt]{article}

\usepackage{blindtext}

\makeatletter
\newcommand\footnoteref[1]{\protected@xdef\@thefnmark{\ref{#1}}\@footnotemark}
\makeatother

\usepackage{multirow}
\usepackage[utf8]{inputenc}
\usepackage{amsmath,amssymb,array}
\usepackage{bbm}
\usepackage{xcolor}
\usepackage{graphicx}
\usepackage{xargs}
\usepackage{algorithmic}
\usepackage[font=small,skip=0pt]{caption}
\usepackage{wrapfig}
\usepackage{caption}
\usepackage{multicol}
\usepackage{booktabs}
\usepackage{todonotes}
\usepackage[american]{babel}

\usepackage[]{url}
\usepackage{colortbl}
\usepackage{csquotes} 
\usepackage{mathtools}
\usepackage{float} 
\usepackage{adjustbox}
\usepackage{dsfont}
\usepackage[ruled, vlined, linesnumbered]{algorithm2e}
\let\oldnl\nl
\newcommand{\nonl}{\renewcommand{\nl}{\let\nl\oldnl}}

\newcommand{\Xspace}{\mathcal{X}}                                           
\newcommand{\Yspace}{\mathcal{Y}} 

\def\argmin{\mathop{\sf arg\,min}}   

\renewcommand{\P}{\mathds{P}} 
\newcommand{\R}{\mathds{R}} 
\newcommand{\E}{\mathds{E}} 
 
\newcommand{\xivec}{\left(x^{(i)}_1, \ldots, x^{(i)}_p\right)^T}  
\newcommand{\xj}{\xv_j}
\newcommand{\D}{\mathcal{D}}                                 
\newcommand{\fx}{f(\mathbf{x})}
\newcommand{\eps}{\epsilon}   
\newcommand{\xv}{\mathbf{x}}
\renewcommand{\xi}[1][i]{\mathbf{x}^{(#1)}}                                          
\newcommand{\yi}[1][i]{y^{(#1)}}  

\newcommand{\fh}{\hat{f}}                                
\newcommand{\Dset}{\{ (\xi, \yi) \}_{i=1}^n}

\usepackage{jmlr2e}

\usepackage{lastpage}

\renewcommand\citet{\cite}
\usepackage{cleveref}
\usepackage{autonum}

\ShortHeadings{Decomposing Global Feature Effects Based on Feature Interactions}{Herbinger, Wright, Nagler, Bischl, and Casalicchio}
\firstpageno{1}

\begin{document}

\title{Decomposing Global Feature Effects Based on Feature Interactions}

\author{\name Julia Herbinger$^{1,2}$ \email julia.herbinger@gmail.com \\
\name Marvin N. Wright$^{3,4,5}$
\email wright@leibniz-bips.de\\
\name Thomas Nagler$^{1,2}$ \email nagler@stat.uni-muenchen.de \\
\name Bernd Bischl$^{1,2}$ \email bernd.bischl@stat.uni-muenchen.de \\
\name Giuseppe Casalicchio$^{1,2}$ \email giuseppe.casalicchio@stat.uni-muenchen.de \\
\addr $^1$ Department of Statistics, LMU Munich, Munich, Germany\\
\addr $^2$ Munich Center for Machine Learning (MCML), Munich, Germany\\
\addr $^3$ Leibniz Institute for Prevention Research and Epidemiology, Bremen, Germany\\
\addr $^4$ Faculty of Mathematics and Computer Science, University of Bremen, Bremen, Germany\\
\addr $^5$ Department of Public Health, University of Copenhagen, Copenhagen, Denmark
}

\editor{}
\maketitle

\begin{abstract}
Global feature effect methods, such as partial dependence plots, provide an intelligible visualization of the expected marginal feature effect. 
However, such global feature effect methods can be misleading, as they do not represent local feature effects of single observations well when feature interactions are present. 
We formally introduce \textit{generalized additive decomposition of global effects} (GADGET), which is a new framework based on recursive partitioning to find interpretable regions in the feature space such that the interaction-related heterogeneity of local feature effects is minimized. We provide a mathematical foundation of the framework and show that it is applicable to the most popular methods to visualize marginal feature effects, namely partial dependence, accumulated local effects, and Shapley additive explanations (SHAP) dependence. Furthermore, we introduce and validate a new permutation-based interaction detection procedure that is applicable to any feature effect method that fits into our proposed framework. We empirically evaluate the theoretical characteristics of the proposed methods based on various feature effect methods in different experimental settings. Moreover, we apply our introduced methodology to three real-world examples to showcase their usefulness. 
\end{abstract}

\begin{keywords}
  interpretable machine learning, feature interactions, partial dependence, accumulated local effect, SHAP dependence
\end{keywords}

\section{Introduction}

Machine learning (ML) models are increasingly used in various application fields such as medicine \citep{shipp2002diffuse} or social sciences \citep{Stachl2020ML}.
Their exceptional predictive performance stems from their ability to capture complex non-linear relationships and feature interactions in the data.
However, this ability also complicates explaining the inner workings of an ML model.
A lack of explainability might hurt trust or might even be a deal-breaker for high-stakes decisions \citep{lipton2018mythos}. 
Hence, ongoing research on model-agnostic interpretation methods to explain any ML model has grown quickly in recent years. 

\begin{figure}[htb]
    \centering
    \includegraphics[width=0.67\textwidth]{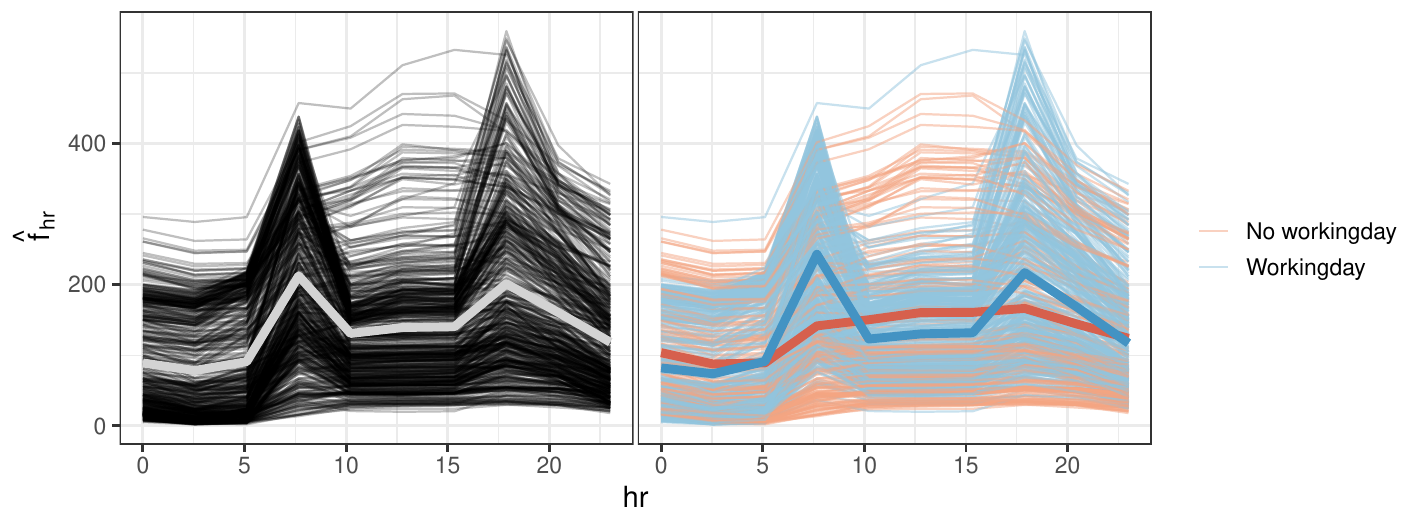}
    \caption{Left: ICE and global PD curves of feature \textit{hr} (hour of the day) of the bikesharing data set \citep{james2022ISLR2}. Right: ICE and regional PD curves of \textit{hr} depending on feature \textit{workingday}. The feature effect of \textit{hr} on predicted bike rentals is different on working days compared to non-working days, which is due to aggregation not visible in the global feature effect plot (white curve on the left).}
    \label{fig:bike_intro}
\end{figure}

One promising type of explanation is produced by feature effect methods, which explain how features influence the model predictions similar to interpreting the coefficients of a linear model or visualizing the non-linear effects of a generalized additive model (GAM) through splines.
We distinguish between local and global feature effect methods. Local feature effect methods---such as individual conditional expectation (ICE) curves \citep{goldstein_peeking_2015} or Shapley values / Shapley additive explanations (SHAP) \citep{strumbelj_2014, lundberg_2017_unified}---explain how each feature influences the prediction of a single observation.
In contrast, global feature effect methods explain the general model behavior. 
Since global feature effects of ML models are often non-linear, they are usually represented as marginal curves instead of single effect numbers; the most popular variants are partial dependence (PD) plots \citep{friedman_greedy_2001}, accumulated local effects (ALE) plots \citep{apley_visualizing_2020}, or SHAP dependence (SD) plots \citep{lundberg2020local}.
These global curves are expressible via an aggregation of local counterparts (e.g., ICE curves are aggregated to a PD curve). 
However, such an aggregation might cause information loss due to heterogeneity in local feature effects (e.g., see  Figure \ref{fig:bike_intro}).
This so-called aggregation bias is usually caused by feature interactions learned by the ML model, leading to a global feature effect that does not appropriately represent many individual observations in the data \citep{herbinger2022repid, mehrabi2021survey}. We refer to this heterogeneity as \textit{interaction-related heterogeneity}.
Consequently, global explanations can be misleading and cannot give a complete picture when feature interactions are present, e.g., see how the different shapes of ICE curves are not well represented by the global PD curve in Figure \ref{fig:bike_intro}.
This issue is particularly relevant when ML models are trained on biased data, as the model may, in turn, learn these biases \citep{mehrabi2021survey}. Although local explanations could reveal the learned bias, it remains hidden in global explanations due to aggregation (e.g., see the COMPAS example in Section \ref{sec:real_world}).

To bridge the gap between local and global effect explanations, so-called subgroup or regional explanations have recently been introduced 
\citep[e.g.,][]{hu2020surrogate,molnar_model-agnostic_2021, scholbeck_marginal_2022, herbinger2022repid}.
However, several pitfalls and difficulties arise in the realm of regional explanations, e.g., the REPID approach by \cite{herbinger2022repid} is limited to only PD plots and only to one feature of interest (see Section~\ref{sec:rel_work} for details).
Another common pitfall is the incomprehensibility in high-dimensional settings \citep{molnar2022pitfalls}, as it would require analyzing hundreds of feature effect plots, making it difficult for the user to understand the underlying relationships effectively.


\paragraph{\normalfont\textit{Contributions.}}
We introduce GADGET, a novel framework, which provides regional explanations based on feature effects by partitioning the feature space into interpretable subspaces. We prove that GADGET minimizes feature interactions for a predefined set of features using any feature effect method satisfying the \textit{local decomposability} axiom (Section \ref{sec:requirements} and \ref{sec:gadget}).
We show that popular feature effect methods (PD, ALE, and SD) satisfy the \textit{local decomposability} axiom (Section \ref{sec:pdp}-\ref{sec:shap}) and illustrate their applicability within GADGET in Section \ref{sec:illustr_comp} with instructions on how to estimate and visualize the corresponding regional feature effects in Appendix \ref{app:overview_estimates}. 
Moreover, we propose several measures to quantify feature interactions based on GADGET, which provide more insights into the learned effects and the remaining heterogeneity related to the interaction (Section \ref{sec:quant_feat_int}).
Furthermore, we introduce and validate the permutation-based interaction detection (PINT) procedure to detect feature interactions in a model-agnostic fashion based on the underlying feature effect method (Section \ref{sec:interact_pimp}).
We empirically evaluate the theoretical characteristics of the different methods based on several simulation settings (Sections \ref{sec:sim} and \ref{sec:sim_pint_valid}) and illustrate their usefulness for two real-world examples in Section \ref{sec:real_world}. 
We provide a third higher-dimensional example in Section \ref{sec:high_dim} and also propose a general filtering procedure for such scenarios.
All proposed methods and reproducible scripts for the experiments are available online via \url{https://github.com/JuliaHerbinger/gadget/}.


\section{Background}
\label{sec:background}


\subsection{General Notation}
\label{sec:notation}

We consider a feature space $\Xspace \subseteq \mathbb{R}^p$ and a target space $\Yspace$ (e.g., $\Yspace = \mathbb{R}$ for regression tasks). 
The random variables for the features are denoted by $X = (X_1, \hdots, X_p)$ and $Y$ for the target variable. The realizations of $X$ and $Y$ are sampled i.i.d. from an unknown joint probability distribution $\P_{X, Y}$ and are denoted by $\D = \Dset$. 
We denote by $\xi = \xivec$ the feature values of the $i$-th observation. 
Arbitrary observations $\xv$ and random variables $X$ we index as $\xv_W$ and $X_W$ (with $W \subseteq \{1,\ldots,p\}$) to refer to the feature values and random variables of the subset. Respectively, $\xv_j$ and $X_j$ refer to the $j$-th feature. 
Negated sets $-W$ (and indices such as in $\xv_{-W}$) denote the set complement. 
We assume that a true function $f$ maps $\Xspace$ to $\Yspace$ up to some label noise, e.g., for regression $Y = f(X) + \eps$. 
Under (independent) Gaussian noise $\eps$, or equivalently L2-loss minimization $f$ will be the conditional mean $\fx = \E[Y | X = \xv]$.
In supervised ML, we strive to approximate this relationship by a prediction model $\fh$ learned on $\D$. 


\subsection{Definition of Interactions and the Functional ANOVA Decomposition}
\label{sec:background_fanova}

\cite{friedman_predictive_2008} defined the presence of an interaction between two features $\xj$ and $\xv_k$ by 
    $\textstyle\E[\partial^2 \fh(X) /(\partial X_j \partial X_k)]^2 >0.
    $
This condition implies that when the feature $\xj$ does not interact with any other feature in $\xv_{-j}$, the prediction function $\fh(\xv)$ can be decomposed into two distinct functions: $h_j(\xj)$, which solely depends on $\xj$, and $h_{-j}(\xv_{-j})$, which solely depends on the remaining features $\xv_{-j}$, i.e.: $\fh(\xv) = h_j(\xj) + h_{-j}(\xv_{-j})$.

This definition of feature interactions is closely related to the concept of \textbf{functional ANOVA}\footnote{This should not be confused with the \textbf{classical ANOVA (Analysis of Variance)}, which is a statistical method that tests for significant differences among group means in a population.
} decomposition, which has already been studied by \cite{sobol1993sensitivity, stone1994fanova} to explain black box functions. 
The functional ANOVA decomposition aims to decompose a square-integrable prediction function $\hat f (\xv)$ into a sum of lower-order functions:
\begin{equation}
\hat{f}(\xv) = 
 g_0 + \sum_{j = 1}^p g_j(\xv_j) + \sum_{j \neq k} g_{jk}(\xv_j,\xv_k) + \ldots + g_{12\ldots p}(\xv)
=g_0 +\sum_{k = 1}^p \; \sum_{\substack{W \subseteq \{1,\ldots,p\},\\ |W| = k}} g_W(\xv_W),
\label{eq:fANOVA}
\end{equation}
where $g_0$ is a constant,
$g_j(\xv_j)$ denotes the main effect of the $j$-th feature, $g_{jk}(\xv_j,\xv_k)$ is the pure two-way interaction effect between features $\xv_j$ and $\xv_k$, and so forth. The last component $g_{12\ldots p}(\xv)$ always allows for an exact decomposition.



The standard functional ANOVA decomposition by \cite{hooker_discovering_2004} assumes independent features and ensures that the component functions $g_W(\xv_W)$ can be optimally and uniquely defined
by imposing the so-called \textit{vanishing condition}\footnote{\label{note1}See Appendix \ref{app:fanova} for more details.
}  \citep{li_general_2012, rahman_generalized_2014}.
This decomposition is widely used in sensitivity analysis to further uniquely decompose the variance of each component from Eq.~\eqref{eq:fANOVA} and derive a feature importance measure \citep[see Sobol indices in][]{sobol1993sensitivity, Owen_2013}. 
\cite{hooker_generalized_2007} introduced the generalized functional ANOVA decomposition with a \textit{relaxed vanishing condition}\footnoteref{note1} that leads to a unique decomposition in case of dependent features.
These conditions guarantee orthogonality between the components representing different feature subsets and ensure that each component is distinct without redundancy. 
In this paper, we do not attempt to compute the exact decomposition presented in Eq.~\eqref{eq:fANOVA}; instead, we rely on the existence of such a decomposition for our theoretical proofs in Section \ref{sec:new_framework}.
We refer to Appendix \ref{app:fanova} for more details on the functional ANOVA decomposition and related research regarding its estimation.

\subsection{Feature Effect Methods}
\label{sec:background_fe}
Below, we summarize global feature effect methods used in our framework in Section~\ref{sec:gadget}.

\paragraph{\normalfont\textit{Partial Dependence.}}
The \emph{PD plot} introduced by \cite{friedman_greedy_2001} visualizes the marginal effect of a set of features $W \subseteq \{1,\ldots, p\}$ for a fitted model $\fh$ by integrating over the joint distribution over all other features in $-W$. 
The number of features in $W$ is typically one or two. 
The theoretical PD function is defined by \footnote{The notation here is used consistently with \cite{friedman_predictive_2008}. Please note: While the right-hand side of the definition always refers to the ML model $\fh$, the $f$ in $f^{PD}_W$ does not refer to the true relationship (see Section \ref{sec:notation}); $f^{PD}_W$ is rather to be read as a new symbol. In the same manner, the ``hat'' of $\hat{f}^{PD}_W$ is not introduced to refer to $\fh$ now -- but expresses the fact that we now estimate based on data.}
\begin{equation}
\label{eq:pdp}
f_W ^{PD}(\xv_W) = E_{X_{-W}} [\fh(\xv_W, X_{-W}) ] = \int \fh(\xv_W, \xv_{-W}) d\P_{X_{-W}} (\xv_{-W}),
\end{equation}
which is estimated using Monte-Carlo integration: $\textstyle \fh_W^{PD}(\tilde\xv_W) = \frac{1}{n} \sum_{i = 1}^n \fh(\tilde\xv_W, \xi_{-W})$ at a specific grid point $\tilde\xv_W$. For constructing a PD curve, we typically use $m$ grid points\footnote{Instead of using all observed feature values $\xv_W$, an equidistant grid or a grid based on randomly selected feature values or quantiles of $\xv_W$ are commonly chosen \citep{molnar2022pitfalls}.} 
denoted by $\tilde\xv_W^{(1)}, \hdots, \tilde\xv_W^{(m)}$ and visualize the pairs $\{ (\tilde\xv_W^{(k)}, \fh_W^{PD}(\tilde\xv_W^{(k)}))\}_{k=1}^m$. 
\citet{goldstein_peeking_2015} introduced \emph{ICE curves}, which, for an observation $\xv$, fixes the values $\xv_{-W}$ to the observed ones, and varies the values $\xv_{W}$ yielding the ICE function $\fh^{\xv}_W(\tilde\xv_W) = \fh(\tilde\xv_W, \xv_{-W})$, where the superscript $\xv$ denotes the observation whose values $\xv_W$ are replaced with $\tilde\xv_W$. 
ICE plots visualize the pairs $\{ (\tilde\xv_W^{(k)}, \fh^{\xi}_W(\tilde\xv_W) \}_{k=1}^m$, representing prediction changes of each observation as the values of features in $W$ vary.
Notably, the PD curve is the point-wise average over all the ICE curves, i.e., $\fh_W^{PD}(\tilde\xv_W) = \frac{1}{n} \sum_{i = 1}^n \fh^{\xi}_W(\tilde\xv_W)$. 
ICE curves can reveal heterogeneous effects caused by interactions between features $\xv_W$ and other features, which remain hidden in the PD curve due to averaging (see Figure \ref{fig:bike_intro}). 
However, they do not reveal which features interact with $\xv_W$. In addition, if the model contains complex interactions, it becomes more difficult to identify clear patterns with interacting features (even if we color-code ICE curves as in Figure \ref{fig:bike_intro}).

\paragraph{\normalfont\textit{ALE.}} 
The PD function uses the marginal distribution and is likely to \emph{extrapolate} in empty or sparse regions of the feature space when features are correlated.
This means that the prediction function is evaluated at unrealistic combinations of feature values in Eq.~\eqref{eq:pdp}, which can lead to inaccurate PD estimates, especially for models with high flexibility in capturing complex feature interactions, such as neural networks \citep{apley_visualizing_2020}.
Hence, ALE uses the conditional distribution to avoid extrapolation. 
The uncentered ALE ${f}^{ALE}_{W}(x)$ for $|W| = 1$ at feature value $x_W \sim \P_{X_W}$ and with $z_0 = \min(\xv_W)$ is defined by 
\begin{equation}
\begin{aligned}
\hspace*{-0.7cm} 
{f}^{ALE}_{W}(x_W) &= \int\limits_{z_{0}}^{x_W} \E\Bigg[  \frac{\partial \fh(X)}{\partial X_W} \bigg \vert X_W = z_W \Bigg] dz_W\\ 
&= \int\limits_{z_{0}}^{x_W} \int  \frac{\partial \fh(z_W, \xv_{-W})}{\partial z_W} d\P_{X_{-W} \vert X_W = z_W}(\xv_{-W}|z_W)  dz_W.
\label{eq:ale}
\end{aligned}
\end{equation}
ALE first calculates local derivatives weighted by the conditional distribution $\P_{X_{-W}\vert X_W = z_W}$ and then accumulates the local effects to generate a global effect curve.
For interpretability reasons, ALE curves are usually centered by the average of the uncentered ALE curve to obtain $\textstyle f^{ALE, c}_{W}(x_W) = {f}^{ALE}_{W}(x_W) - \int{f}^{ALE}_{W}(\xv_W) \, d\P_{X_W}(\xv_W)$.

Eq.~\eqref{eq:ale} is usually estimated by splitting the value range of $\xv_W$ in intervals and calculating the partial derivatives for all observations within each interval. The partial derivatives are estimated by the differences between the predictions of the upper ($z_{k, W}$) and lower ($z_{k-1, W}$) bounds of the $k$-th interval for each observation. 
The accumulated effect up to the feature value $x_W$ is then calculated by summing up the average partial derivatives (weighted by the number of observations $n(k)$ within each interval) over all intervals until the interval that includes $x_W$, which is denoted by $k(x_W)$:
\begin{equation}
  \hat{{f}}^{ALE}_{W}(x_W) = \sum_{k = 1}^{k(x_W)}\frac{1}{n(k)}\sum_{i: \; x_W^{(i)} \in \; ]z_{k-1, W}, z_{k, W}]}\left[\fh(z_{k, W}, \xi_{-W}) -\fh(z_{k-1, W}, \xi_{-W})\right]. 
  \label{eq:ale_est}
\end{equation}

\paragraph{\normalfont\textit{SHAP Dependence.}}
Shapley values \citep{shapley_1951} have been adapted from game theory to ML to quantify feature effects on a local level by fairly distributing the prediction of a single observation $\xv$ among all features \citep{strumbelj_2014}.
The Shapley value for the $j$-th feature is calculated by assessing its contribution to every possible feature combination $W \subseteq \{1,\ldots, p\} \setminus \{j\}$ using the formula
$$
\phi_j (\xv) = \sum_{W \subseteq \{1,\ldots,p\}\setminus \{j\}} \frac{|W|!(p - |W| - 1)!}{p!}(f_{W \cup \{j\}}(\xv_{W \cup \{j\}}) - f_{W}(\xv_W)),
$$
where $f_{W}(\xv_W) = E_{X_{-W}} [\fh(\xv_W, X_{-W}) ]$ is typically the PD function \citep{casalicchio2019visualizing}. Since estimating the above equation is computationally expensive, various efficient approximation algorithms have emerged \citep{aas_explaining_2021, chen_algorithms_2023}, including SHAP \citep{lundberg_2017_unified}, which conceptualizes Shapley values as a solution to a quadratic program with equality constraints.
To visualize the global feature effect of a feature on the prediction, \cite{lundberg2020local} proposed the SHAP dependence (SD) plot as an alternative to PD plots. 
SD plots visualize the feature values of the $j$-th feature on the x-axis and its corresponding SHAP values on the y-axis in a scatter plot, either for all or for a representative sample of observations (see Figure \ref{fig:repid_extrapol}).
Feature interactions between the feature of interest and other features influence the shape of the resulting point cloud. 
To highlight patterns in the point cloud that may indicate interactions with another feature, \cite{lundberg2020local} suggest color-coding the points by other (potentially interacting) features. 
However, identifying a clear trend or pattern in the color-coded point cloud becomes challenging in the presence of complex feature interactions (similar to ICE plots). 

Shapley values can be quantified using either a marginal-based (similar to PDs) or a conditional-based approach. While the former might encounter extrapolation issues in the presence of feature correlations, the latter presents the challenge of estimating the conditional distribution. For a more thorough discussion on this topic, see Appendix \ref{app:marg_vs_cond}.

\subsection{Quantification of Interaction Effects}
\label{sec:background_int}
Visualization methods for feature effects are usually limited to a maximum of two features. To understand which feature effects depend on other features due to interactions, we are interested in detecting and quantifying the strength of feature interactions.
Building on the mean-centered PD function $\textstyle \fh_j^{PD,c}(\tilde x_j) = \fh_j^{PD}(\tilde x_j) - \frac{1}{m} \sum_{k = 1}^m \fh_j^{PD}(\tilde x_j^{(k)})$ at each grid point $\tilde x_j \in \{\tilde x_j^{(1)}, \dots, \tilde x_j^{(m)}\}$, \cite{friedman_predictive_2008} introduced the H-Statistic as a global interaction measure that quantifies the interaction strength between a feature of interest $\xj$ and all other features $\xv_{-j}$ by
\begin{equation} \textstyle
    \mathcal{\hat H}_{j}^2 = \frac{\sum\nolimits_{i=1}^n \left( 
    \fh^c(\tilde\xv^{(i)}) - \hat f_{j}^{PD,c}(\tilde x_j^{(i)}) -  \hat f_{-j}^{PD,c}(\tilde\xv^{(i)}_{-j}) 
    \right)^2}{\sum_{i=1}^n \left( \fh^c(\tilde\xv^{(i)}) \right)^2 }.
\label{eq:hstatistic}
\end{equation}
Here, $\fh^c(\tilde\xv) = \fh(\tilde\xv) - \frac{1}{n} \sum_{i=1}^n \fh(\tilde\xv^{(i)})$ is the mean-centered prediction function and thus the denominator in Eq.~\eqref{eq:hstatistic} corresponds to the prediction function's variance. Hence, $\mathcal{\hat H}_{j}^2$ quantifies how much of the prediction function's variance can be attributed to the interaction between feature $\xj$ and all other features $\xv_{-j}$.\footnote{If the feature $\xj$ does not interact with any other feature in $\xv_{-j}$, the mean-centered prediction function can be decomposed as $\fh^c(\tilde\xv) = \fh_j^{PD,c}(\tilde\xv_j) + \fh_{-j}^{PD,c}(\tilde\xv_{-j})$ (see Section~\ref{sec:background_fanova}), leading to $\mathcal{\hat H}_{j}^2 = 0$.} 
Thus, the H-Statistic can be used to rank features according to the global interaction strength. Despite its standardization between $0$ and $1$, it is unclear which value is generally considered a high or a low interaction. Furthermore, being derived from PD functions, the H-Statistic inherits the same disadvantages---particularly regarding extrapolation---as the PD plot itself.

\section{Related Work}
\label{sec:rel_work}
Below, we outline related work and define research gaps concerning our main contributions: GADGET (Section \ref{sec:new_framework}) for regional effects and PINT (Section \ref{sec:interact_pimp}) for testing interactions.

\paragraph{\normalfont\textit{Related Work on Regional Effects.}}

Regional explanations~\citep{molnar_model-agnostic_2021, britton_vine_2019, hu2020surrogate, scholbeck_marginal_2022, herbinger2022repid}---sometimes also referred to cohort explanations \citep{sokol2020explainability}---explain subspaces of the entire feature space, thereby bridging the gap between local and global explanations. 
In general, we can distinguish between unsupervised and supervised methods to define these subspaces.
Concerning the former, \cite{britton_vine_2019} suggests the use of an unsupervised clustering approach to group ICE curves based on their partial derivatives and apply a decision tree post hoc for explanation purposes. A similar approach has been introduced by \cite{zhang_interpreting_2021}. However, \cite{herbinger2022repid} showed that such unsupervised approaches can produce misleading interpretations. To overcome these downsides, other works such as \cite{molnar_model-agnostic_2021, scholbeck_marginal_2022, herbinger2022repid} propose methods that are based on a supervised decision tree approach to define regions that are interpretable and distinct. The first two references focus on finding regions with few feature correlations and few non-linearities, respectively.
Instead, the REPID method \citep{herbinger2022repid} minimizes feature interactions within each region and decomposes the global PD into regional PDs, ensuring the homogeneity of ICE curves within each region. However, REPID has its limitations: First, REPID generates feature-specific regions by minimizing interactions between a particular feature and all other features. Applying the method to a different feature of interest typically results in different regions, complicating interpretations when multiple features are of interest. 
Second, REPID relies on ICE and PD plots and thus inherits also their issues, such as extrapolation, potentially providing misleading interpretations if both feature interactions and correlations are present \citep{molnar2022pitfalls}.
To overcome these limitations, we propose a more flexible framework in Section \ref{sec:new_framework} that is capable of producing regions where interactions between multiple features are minimized and where different feature effect methods other than ICE and PD plots can be used. 

\paragraph{\normalfont\textit{Related Work on Detecting Feature Interactions.}}

ANOVA is one of the most prominent statistical testing procedures to detect significant (two-way) interaction terms in statistical models. 
An algorithm to detect important feature interactions without a formal definition of a statistical test was introduced along with the standard functional ANOVA decomposition in \cite{hooker_discovering_2004}.
Later, \citet{mentch_hooker_2017} introduced a statistical test for tree-based methods to detect additive structures and, thus, feature interactions. 
\citet{henelius2016finding} introduced a statistical test for classification tasks to detect additive structures between different subsets of features based on their earlier work \cite{henelius2014peek}. In their approach, feature values of the considered feature subset are permuted jointly within each class to break possible interactions with other feature subsets while preserving interactions within the subset. However, this approach is limited to classification tasks and breaks correlations with features outside of the regarded subset. Consequently, learned spurious interactions are detected by the algorithm as true feature interactions. 
Furthermore, statistical tests for two-way interaction detection have been introduced by \cite{loh2002regression} or \cite{sorokina_detecting_2008}. While the former provides a formal statistical test statistic, the latter is based on a heuristic to detect two-way interactions by comparing the performance of a full and restricted model.
To the best of our knowledge, a model-agnostic procedure designed for classification and regression tasks to detect interactions between a feature of interest $\xv_j$ and the remaining features $\xv_{-j}$, as presented in Section~\ref{sec:interact_pimp}, has not yet been introduced.

\section{Generalized Additive Decomposition of Global EffecTs (GADGET)}
\label{sec:new_framework}
Our proposed framework, GADGET, additively decomposes global feature effects by minimizing feature interactions to produce regional feature effects. One or more features can be used to derive interpretable subregions where aggregated regional feature effects better represent individual observations within each region. Below, we will introduce the GADGET methodology, applicable to various feature effect methods, by formally defining the required axiom and demonstrating its satisfaction by popular methods.

\subsection{Measuring Interaction-Related Heterogeneity}
\label{sec:requirements}

\begin{figure}[htb]
    \centering
    \includegraphics[width=\textwidth]{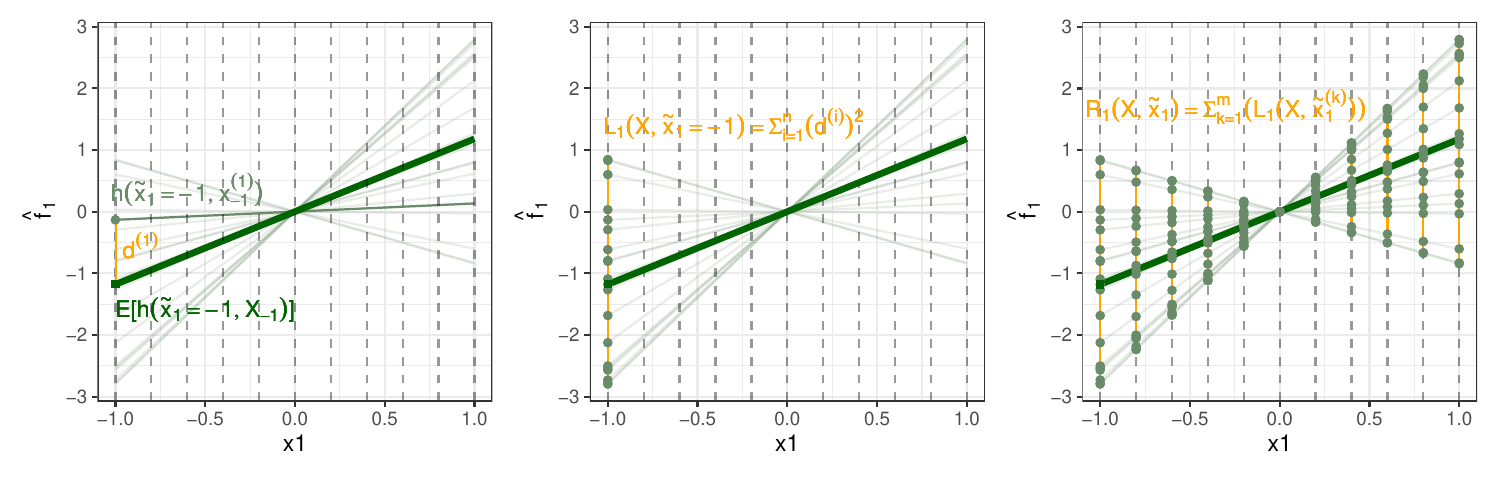}
    \caption{Mean-centered ICE and PD curves of feature $\xv_1$ of the described simulation example. Left: Illustrates the distance $d^{(1)}$ that is calculated within Eq.~\eqref{eq:loss} between the local feature effect (here: $h(x_j,\xv_{-j}^{(i)}) := \fh^c(x_j,\xv_{-j}^{(i)})$, i.e., the mean-centered ICE curve) and the expected effect within region $\mathcal{A}_g$ (here: $\E[h (x_j, X_{-j})|\mathcal{X}] = \E[\fh^c (x_j, X_{-j})|\mathcal{X}]$, i.e. the global mean-centered PD) at the first grid point $\tilde{x}_1 = -1$ for the first observation. Middle: The distances $d$ are calculated for all mean-centered ICE curves at the first grid point and the squared values are summed up to obtain the loss function value for the first grid point as defined in Eq.~\eqref{eq:loss}. Right: The risk function value is calculated according to Eq.~\eqref{eq:risk} by aggregating the loss function values over the valid grid points. This measures the heterogeneity of the mean-centered ICE curves of feature $\xv_1$, which quantifies the interaction-related heterogeneity between $\xv_1$ and $\xv_{-1}$ based on Theorem~\ref{theorem:interact_rel}.}
    \label{fig:expl_heterogeneity}
\end{figure}

Let $h(x_j, \xv_{-j}^{(i)})$ be a function that quantifies the local effect of the $j$-th feature on the $i$-th observation at a specific value $x_j \in \mathcal{X}_j$, calculated by a function $h: \mathbb{R}^{p} \rightarrow \mathbb{R}$. We keep this function generic for now, but discuss in Sections \ref{sec:pdp} to \ref{sec:shap} several specific choices, including the underlying local effect functions of PD, ALE, and SHAP dependence (SD) curves. For example, in the case of PD, $h$ is defined by the $i$-th mean-centered ICE curve of the $j$-th feature at one specific grid point $x_j$ (see Figure~\ref{fig:expl_heterogeneity}).

Let $\E[h(x_j, X_{-j})|\mathcal{A}_g]$ be the expected effect of the $j$-th feature at a specific value $x_j$ regarding $X_{-j}$ given the event $X \in \mathcal{A}_g $ where $\mathcal{A}_g = \iota_{g,1} \times \dots \times \iota_{g,p} \subseteq \mathcal{X}$ is a hypercube and the $\iota_{g,j} \subseteq \R$ are (potentially unbounded) intervals. 
Our goal is to find a partition of the feature space into subsets $\mathcal{A}_g$ with little interaction-related heterogeneity.
The deviation between the local feature effect and the expected feature effect at a specific value $x_j$ for subspace $\mathcal{A}_g$ is defined by a point-wise loss function, e.g., the squared distance:
\begin{eqnarray} 
    \mathcal{L}_j\left(\mathcal{A}_g, x_j\right) &=&
    \sum\limits_{i: \xi \in \mathcal{A}_g}
     \left(h(x_j, \mathbf{x}_{-j}^{(i)}) - \E[h (x_j, X_{-j})|\mathcal{A}_g] \right)^2.
     \label{eq:loss}
\end{eqnarray}
Eq.~\eqref{eq:loss} measures the heterogeneity of the local feature effects of feature $\xv_j$ at a specific value $x_j$ by calculating the L2 loss as demonstrated in the left and middle plots of Figure~\ref{fig:expl_heterogeneity}. Here, the local feature effects are the mean-centered ICE curves and the respective mean-centered PD curve represents the expected effect of the feature of interest.

The risk function $\mathcal{R}_j$ for the $j$-th feature and subspace $\mathcal{A}_g$ is defined by aggregating the point-wise losses of Eq.~\eqref{eq:loss} over a given $m$-dimensional grid of feature values $\tilde\xv_j$: 

\begin{eqnarray} 
    \mathcal{R}_j\left(\mathcal{A}_g, \tilde\xv_j\right) &=&
    \sum\limits_{k: k \in \{1,\ldots, m\} \wedge \tilde x_j^{(k)} \in \iota_j} \mathcal{L}_j\left(\mathcal{A}_g, \tilde x_j^{(k)}\right).
    \label{eq:risk}
\end{eqnarray}

For instance, $\tilde\xv_j$ can be $m$ values sampled from the realizations of the $j$-th feature, an equidistant grid or quantiles, but we restrict the summation in Eq.~\eqref{eq:risk} to values which fall inside the considered subspace $\mathcal{A}_g$, so $\iota_{g,j}$.
It follows that Eq.~\eqref{eq:risk} summarizes the heterogeneity of local feature effects of the feature $\xv_j$ for its entire feature range within the region $\mathcal{A}_g$ in one number and therefore quantifies how much the feature's influence of individual observations varies from the expected (average) feature's influence on the predictions (e.g., right plot of Figure~\ref{fig:expl_heterogeneity} in the case of mean-centered ICE and PD curves).

The loss in Eq.~\eqref{eq:loss} measures the heterogeneity of the local effects, which can be decomposed into the main and interaction effects of involved features (see Eq.~\eqref{eq:fANOVA}). Since we are only interested in measuring the interaction-related heterogeneity of the feature of interest, disregarding variability caused by any other effects, the local feature effect function $h$ needs to be defined in such a way that the heterogeneity of local feature effects measured by Eq.~\eqref{eq:loss} and ~\eqref{eq:risk} are solely based on feature interactions between the feature of interest $\xv_j$ and the remaining feature in the data set.
This is possible when Axiom \ref{axiom:feat_rel} holds.\footnote{We 
show in Sections \ref{sec:pdp} to \ref{sec:shap} that common feature effect methods satisfy this general decomposition based on the functional ANOVA decomposition in Eq. \eqref{eq:fANOVA}.} 
\begin{axiom}[Local Decomposability] A local feature effect function $h: \mathbbm{R}^p \rightarrow \mathbbm{R}$ satisfies the \textit{local decomposability axiom} if and only if $\forall i \in \{1,\dots, n\}$ the decomposition of the $i$-th local effect $h(x_j, \xv_{-j}^{(i)})$ of feature $\xv_j$ at $x_j$ solely depends on main and higher-order effects of feature $\xv_j$: 
$$h(x_j, \xv_{-j}^{(i)}) = g_j(x_j) + \sum\limits_{k = 1}^{p-1} \; \sum\limits_{\substack{W \subseteq -j,\\ |W| = k}} g_{W \cup j}(x_j,\xv_W^{(i)}).$$
\label{axiom:feat_rel}
\end{axiom}

In Sections \ref{sec:pdp} to \ref{sec:shap} we show that this axiom is fulfilled by the most popular feature effect methods, at least when properly centered.
If Axiom \ref{axiom:feat_rel} is satisfied, then Theorem \ref{theorem:interact_rel} guarantees that the loss function defined in Eq.~\eqref{eq:loss} quantifies the interaction-related heterogeneity of local effects (feature interactions) of feature $\xv_j$ at $x_j$ within the regarded subspace $\mathcal{A}_g$. 
Specifically, Theorem~\ref{theorem:interact_rel} shows that the main effect $g_j(x_j)$ no longer affects the loss $\mathcal{L}_j$, rendering $\mathcal{L}_j$ a pure measure of interaction-related heterogeneity.
\begin{theorem}
   If the local feature effect function $h(x_j, \xv_{-j}^{(i)})$ satisfies Axiom \ref{axiom:feat_rel}, then the loss function $\mathcal{L}_j\left(\mathcal{A}_g, x_j\right)$ defined in Eq.~\eqref{eq:loss} only depends on feature interactions between the feature $\xv_j$ at $x_j$ and features in $-j$: 
   $$
    \mathcal{L}_j\left(\mathcal{A}_g, x_j\right) = \sum\limits_{i: \xi \in \mathcal{A}_g}
     \left(\sum\limits_{k = 1}^{p-1} \; \sum\limits_{\substack{W \subseteq -j,\\ |W| = k}} g_{W \cup j}(x_j,\xv_W^{(i)}) - \E[ g_{W \cup j}(x_j,X_W)|\mathcal{A}_g]\right)^2.
   $$
   The proof can be found in Appendix \ref{app:proof_theorem2}.
 \label{theorem:interact_rel}  
\end{theorem}

\paragraph{\normalfont\textit{Illustration.}}
To better understand Axiom~\ref{axiom:feat_rel}, Theorem ~\ref{theorem:interact_rel}, Eq.~\ref{eq:loss} and ~\ref{eq:risk}, we provide an illustrative example. 
Let $X_1, X_2, X_3 \sim \mathbbm{U}(-1,1)$ be independent, with the true underlying relationship defined by $Y = X_1 + X_2 + X_3 + X_1 X_2 + \epsilon$, where $\epsilon \sim \mathbb{N}(0,0.0025)$. We draw 20 observations and fit a correctly specified linear model to the data. We define the local feature effect function as the mean-centered ICE curves (Section \ref{sec:pdp}) and use $\xv_1$ as the feature of interest. The $i$-th ICE curve value at grid point $\tilde{x}_1$ is given by $\fh(\tilde{x}_1, \xv_{-1}^{(i)}) = \tilde{x}_1 + x_2^{(i)} + x_3^{(i)} + \tilde{x}_1 x_2^{(i)}$, which corresponds to the decomposition in Eq.~\eqref{eq:fANOVA}. Since this decomposition includes effects independent of feature $\xv_1$ (i.e., $g_3(x_3^{(i)}) = x_3^{(i)}$), Axiom \ref{axiom:feat_rel} does not hold. Therefore, the ICE curve needs to be centered by its mean: $\fh^c(\tilde{x}_1, \xv_{-1}^{(i)}) = \fh(\tilde{x}_1, \xv_{-1}^{(i)}) - \E[\fh(X_1, \xv_{-1}^{(i)}] = \tilde{x}_1 - \E[X_1] + (\tilde{x}_1- \E[X_1]) x_2^{(i)} 
 \approx \tilde{x}_1 +  \tilde{x}_1 x_2^{(i)}$ which satisfies Axiom \ref{axiom:feat_rel} with main effect $g_j(x_j) = \tilde{x}_1$ and the respective higher-order effect $g_{W\cup j}(x_j, \xv_W^{(i)}) = \tilde{x}_1 x_2^{(i)}$.
 Figure \ref{fig:expl_heterogeneity} shows the mean-centered ICE curves and the mean-centered PD curve, which is calculated as the average over all mean-centered ICE curves. The loss function in Eq.~\eqref{eq:loss} aggregates the squared distance between each mean-centered ICE curve and the PD curve at a specific grid point as demonstrated in the left and middle plot of Figure \ref{fig:expl_heterogeneity} for $\tilde{x}_1 = -1$. Since mean-centered ICE curves satisfy Axiom \ref{axiom:feat_rel}, this distance only measures interaction-related heterogeneity as stated in Theorem \ref{theorem:interact_rel}: $h(x_1, \mathbf{x}_{-1}^{(i)}) - \E[h (x_1, X_{-1})|\mathcal{X}] = \tilde{x}_1 + \tilde{x}_1 x_2^{(i)} - \tilde{x}_1 -  \tilde{x}_1 \E[ X_2] = \tilde{x}_1 x_2^{(i)} - \tilde{x}_1 \E[ X_2] \approx \tilde{x}_1 x_2^{(i)}$. Thus, the loss function and therefore also the risk function in Eq.~\eqref{eq:risk}, which aggregates over all valid grid points of the feature $\xv_j$ within the respective region (e.g., right plot Figure \ref{fig:expl_heterogeneity}), solely depend on the interaction effect between the feature of interest ($\xv_1$) and other features in the data set (here: $\xv_2$). 

\subsection{The GADGET Algorithm}
\label{sec:gadget}

The goal of the GADGET algorithm is to find regions with little interaction-related heterogeneity. A user specifies two sets: $S \subseteq \{1,\ldots, p\}$ containing the features of interest for which we want to receive representative regional feature effect plots by minimizing their interaction-related heterogeneity, and $Z \subseteq \{1,\ldots, p\}$ containing features assumed to interact with those in $S$.
Features in $Z$ serve as split features when partitioning the feature space to minimize interactions with features in $S$. 
Features in $-S$ are usually assumed not to interact with other features and are thus only calculated and visualized on a global level.
Algorithm \ref{alg:tree} defines a single partitioning step of GADGET, which is inspired by the CART algorithm \citep{breiman:1984} and is recursively repeated until a certain stop criterion is met (see Appendix \ref{app:recom_stop_crit} for more details). We greedily search for the best split point within the feature subset $Z$ such that the interaction-related heterogeneity of feature subset $S$ is minimized in the two resulting subspaces. The interaction-related heterogeneity is measured by the variance of the local feature effects of $\xv_j$ within the new subspace (see Eq.~\ref{eq:risk}). 
Hence, for each candidate split feature $z$ and candidate split point $t$, we sum up the risks of the two resulting subspaces for all features in $S$ (line 6 in Algorithm \ref{alg:tree}) and then choose the split $(\hat t,\hat z)$ that minimizes the interaction-related heterogeneity of all features $j \in S$ (line 9 in Algorithm \ref{alg:tree}).
Note that Algorithm \ref{alg:tree} is exemplarily defined for numeric features, however, GADGET can also handle categorical features which is described in detail in Appendix \ref{app:categ_feat}. 

\begin{algorithm}[tbh]
    \caption{Partitioning algorithm of GADGET}
    \label{alg:tree}
    \begin{algorithmic}[1]
      \STATE \textbf{input: } subspace $\mathcal{A} \subseteq \mathcal{X}$, risk $\mathcal{R}_j$, features $S$, splitting set $Z$
      , grids $\tilde\xv_j \;\forall j \in S$
      \STATE \textbf{output: } subspaces $\mathcal{A}_l^{\hat{t},\hat{z}}$ and $\mathcal{A}_r^{\hat{t}, \hat{z}}$
      \FOR{each feature indexed by $z \in Z$}{
        \FOR{every split $t$ on feature $\xv_z$ }{
            \STATE $\mathcal{A}_l^{t,z} = \{\xv \in \mathcal{A} | \xv_z \leq t\}$ ;
            $\mathcal{A}_r^{t,z} = \{\xv \in \mathcal{A} | \xv_z > t\}$
            \STATE $\mathcal{I}(t,z) = \sum_{j \in S} \left( \mathcal{R}_j(\mathcal{A}_l^{t,z}, \tilde\xv_j) + \mathcal{R}_j(\mathcal{A}_r^{t,z}, \tilde\xv_j) \right)$
        }\ENDFOR
      } \ENDFOR   
      \STATE Choose $(\hat{t}, \hat{z}) \in \arg\min\nolimits_{t,z} \mathcal{I}(t,z)$
    \end{algorithmic}
  \end{algorithm}
  \footnotetext{$S$ and $Z$ can be distinct or partially or fully overlap with each other, see also Section \ref{sec:interact_pimp}.}

If the set of splitting features $Z$ contains all features interacting with features in $S$, the GADGET algorithm eventually finds regions with no interaction-related heterogeneity (see Theorem \ref{corollary:objective}). 
This implies that any feature interactions inherent in the feature effects of the features within $S$ can be minimized, leaving only their main effects within each subspace. 
Thus, the joint regional feature effect $f_{S|\mathcal{A}_g}$ of all features in $S$ within each subspace $\mathcal{A}_g$ can be uniquely and additively decomposed into the respective univariate regional feature effects $f_{j|\mathcal{A}_g}$ of all features in $S$: 
\begin{equation}
   f_{S|\mathcal{A}_g}(\xv_S) = \sum_{j \in S} f_{j|\mathcal{A}_g} (\xv_j).
   \label{eq:decomp_fe}
\end{equation}
While the global feature effects are estimated on the entire feature space $\mathcal{X}$, the regional feature effects are determined by conditioning on the respective subspace, i.e., on $X \in \mathcal{A}_g$.

\begin{theorem} Suppose the local feature effect function $h$ satisfies Axiom \ref{axiom:feat_rel}. If the feature subset $-Z$ does not contain any features interacting with any features indexed by $j \in S$, i.e., $\forall j \in S$ and $\forall L \subseteq -Z$ with $-Z \subset \{1,\ldots, p\}\setminus j$ and $\forall K \subseteq Z := \{\emptyset, 1,\ldots, p\} \setminus \{-Z\cup j\}$ it holds: $g_{L \cup K \cup j}(X_j,X_L, X_K) = 0$, then after sufficiently many iterations, GADGET returns regions with $\mathcal{R}_j(\mathcal{A}_g, \tilde\xv_j$) = 0 for every final region $\mathcal{A}_g$, i.e., the theoretical minimum of the objective function $\mathcal{I}$ in Algorithm~\ref{alg:tree} is $0$. 
The proof can be found in Appendix \ref{app:proof_cor3}.
\label{corollary:objective}
\end{theorem}


In practice, one rarely wants to run the algorithm to convergence.
The question of how many partitioning steps should be performed depends on the underlying research question. 
If the user is more interested in reducing the interaction-related heterogeneity as much as possible, they might split rather deeply, depending on the complexity of interactions learned by the model. However, this might lead to many regions that are more challenging to interpret. If a small number of regions is of interest, one might prefer a shallow tree, thus reducing only the heterogeneity of the features that interact the most. More detailed recommendations on specifying these stop criteria are provided in Appendix \ref{app:recom_stop_crit}.

\subsection{GADGET-PD}
\label{sec:pdp}
The PD plot is based on ICE curves (local feature effects). 
However, ICE curves do not satisfy Axiom \ref{axiom:feat_rel}, since the decomposition of the $i$-th ICE curve (i.e., $\fh^\xv_j(x_j) = \fh(x_j, \xv_{-j}^{(i)})$) also contains main or interaction effects of the $i$-th observation that are independent of feature $\xv_j$ (see Appendix \ref{app:proof_pd}). 
These effects can be canceled out by centering ICE curves w.r.t. the mean of each curve, yielding the following local effect function in GADGET-PD:
$$h(x_j, \xv_{-j}^{(i)}) = \fh^c(x_j, \xv_{-j}^{(i)}) = \fh(x_j, \xv_{-j}^{(i)}) - \E[\fh(X_j, \xv_{-j}^{(i)})|\mathcal{A}_g].$$

\begin{theorem}
    Mean-centered ICE curves satisfy the local decomposability axiom (Axiom \ref{axiom:feat_rel}).
    The proof is given in Appendix \ref{app:proof_pd}.
    \label{theorem:axiom_pd}
\end{theorem}

\begin{figure}[tbh]
    \centering
      \includegraphics[width=0.49\linewidth]{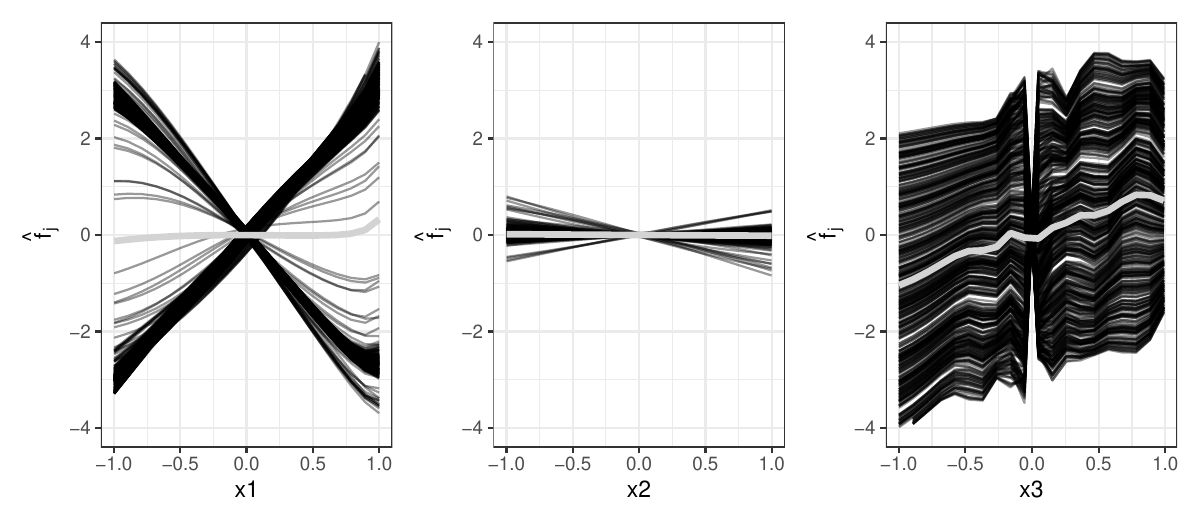}
      \scalebox{1}{
      \hspace{15pt} 
      \begin{tikzpicture}
      \usetikzlibrary{arrows}
        \usetikzlibrary{shapes}
         \tikzset{treenode/.style={draw}}
         \tikzset{line/.style={draw, thick}}
        \node [treenode](a0) {} ; [below=1pt,at=(4,0)]  {};
         \node [treenode, below=0.3cm, at=(a0.south), xshift=-3.0cm]  (a1) {};
         \path [line] (a0.south) -- + (0,-0.2cm) -| (a1.north) node [midway, above] {};
         \node[at=(a0.south), xshift=-1.5cm]{$x_3 \leq -0.003$};
      \end{tikzpicture}
      \hspace{35pt}
      \begin{tikzpicture}
      \usetikzlibrary{arrows}
        \usetikzlibrary{shapes}
         \tikzset{treenode/.style={draw}}
         \tikzset{line/.style={draw, thick}}
        \node [treenode] (a01) {};[below=5pt,at=(node1.south) , xshift=3.5cm]
         \node [treenode, below=0.3cm, at=(a01.south), xshift=3.0cm]  (a2) {};
         \path [line] (a01.south) -- +(0,-0.2cm) -|  (a2.north) node [midway, above] {};
         \node[at=(a01.south), xshift=1.5cm]{$x_3 > -0.003$};
      \end{tikzpicture}
      }
    \includegraphics[width=0.49\linewidth]{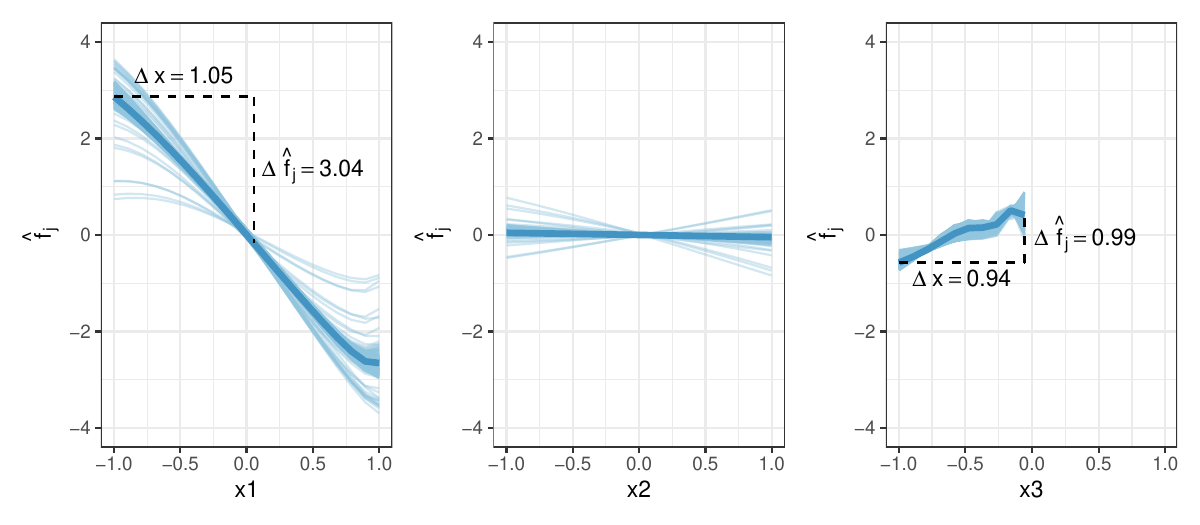}
    \includegraphics[width=0.49\linewidth]{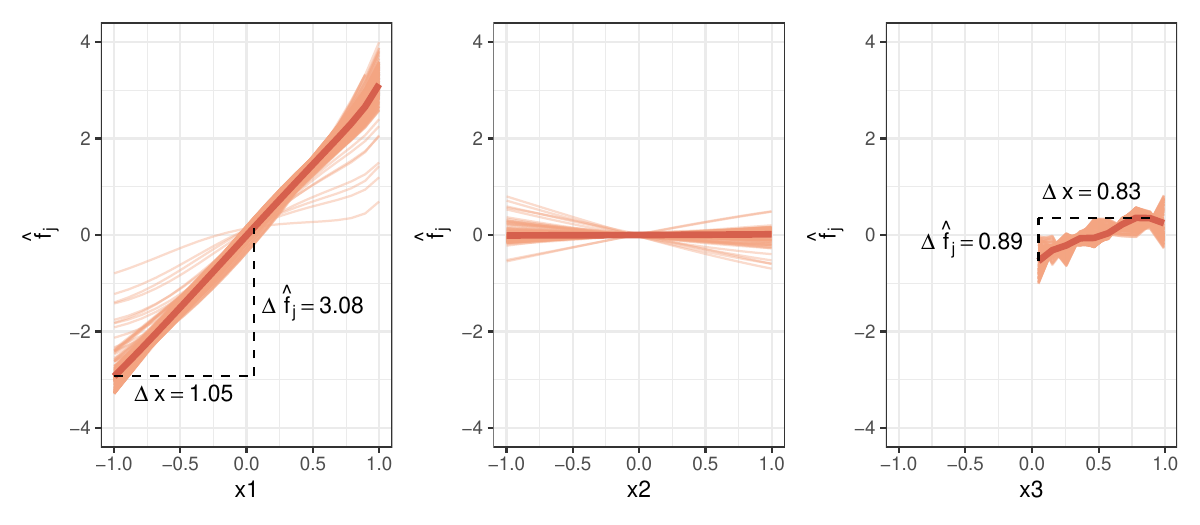}
    \vspace{.2in}
    \caption{Visualization of applying GADGET with $S = Z = \{1,2,3\}$ to mean-centered ICE curves of the uncorrelated simulation example with $Y = 3X_1\mathbbm{1}_{X_3>0} - 3X_1\mathbbm{1}_{X_3\leq 0}  + X_3 + \epsilon$ where $\epsilon \sim \mathbb{N}(0,0.09)$. Plots show mean-centered ICE and PD curves on the entire feature space (upper) and within regions after partitioning the feature space w.r.t. $\xv_3 = -0.003$ (lower).
    }
    \label{fig:pdp_dt}
\end{figure}

Note that the mean-centering constant depends on the region $\mathcal{A}_g$.
In Figure \ref{fig:pdp_dt}, for example, we use feature $\xv_3$ as a feature of interest in $S$ and as a split feature in $Z$. Consequently, the visualized range of feature $\xv_3$ is partitioned based on the split point identified by the GADGET algorithm and the mean-centering constant changed within the new subspace (i.e., $\E[\fh(X_j, \xv_{-j}^{(i)})|\mathcal{A}_g] \neq \E[\fh(X_j, \xv_{-j}^{(i)})]$), necessitating adjustment to avoid additive (non-interaction) effects in the new subspace.

The loss function used within GADGET-PD to minimize the interaction-related heterogeneity is defined by the variability of mean-centered ICE curves: 
\begin{equation}
   \mathcal{L}_j^{PD}\left(\mathcal{A}_g, x_j\right) =
    \sum\limits_{i: \xi \in \mathcal{A}_g}
     \left(\fh^c(x_j, \mathbf{x}_{-j}^{(i)}) - \E[\fh^c (x_j, X_{-j})|\mathcal{A}_g] \right)^2. 
     \label{eq:loss_pd}
\end{equation}
In Appendix \ref{app:repid}, we show that the REPID method discussed in Section \ref{sec:rel_work} is a special case of GADGET-PD for the loss function in Eq.~\eqref{eq:loss_pd} with only one feature of interest (i.e., $S = j$) and where we consider all other features to be potential split features (i.e., $Z = -j$).

\paragraph{\normalfont\textit{Illustration.}}
To provide a better understanding of GADGET-PD, we consider a commonly used simulation example in a slightly modified form in the context of feature interactions \citep[see e.g.,][]{goldstein_peeking_2015, herbinger2022repid}: Let $X_1, X_2, X_3 \sim \mathbbm{U}(-1,1)$ be independently distributed and the true underlying relationship be defined by $Y = 3X_1\mathbbm{1}_{X_3>0} - 3X_1\mathbbm{1}_{X_3\leq 0} + X_3 + \epsilon$, with $\epsilon \sim \mathbb{N}(0,0.09)$. We then draw 500 observations from these random variables and fit a feed-forward neural network (NN) with a single hidden layer of size 10 and weight decay of $0.001$.\footnote{We tuned the size of the hidden layer over ($3, 5, 10, 20, 30$) and the weight decay value over ($0.5, 0.1, 0.01, 0.001, 0.0001, 0.00001$) via 5-fold cross-validation using grid search.} 
The $R^2$ measured on a separately drawn test set of size $10000$ following the same distribution is $0.94$. 

Figure \ref{fig:pdp_dt} visualizes the global PD and ICE curves along with the regional ones after applying GADGET-PD. 
The global plots for features $\xv_1$ and $\xv_3$ show high heterogeneity in the underlying local effects, indicating that the global PD curve does not represent the respective ICE curves well. For example, the nearly horizontal global PD curve of $\xv_1$ (grey line) indicates no influence on the model's predictions. Consequently, it is not representative of the underlying local feature effects (ICE curves), which vary highly due to the feature interaction with $\xv_3$.
We applied GADGET-PD with $S = Z = \{1, 2, 3\}$ to visualize the (regional) feature effect of all available features by considering all features as possible interaction (split) features. GADGET-PD performed one split 
with regard to $\xv_3 \approx 0$. Thus, the correct split feature and its corresponding split point are found by GADGET-PD such that the interaction-related heterogeneity of all features in $S$ is almost completely reduced and only the main effects within the subspaces remain. Hence, the joint mean-centered PD function $f^{PD, c}_{S|\mathcal{A}_g}$ within each final subspace $\mathcal{A}_g$ can be decomposed into the respective 1-dimensional mean-centered PD functions as defined in Eq.~\eqref{eq:decomp_fe}. The decomposition is explained in more detail in Appendix \ref{app:decomp_pd}.

The main issue with local ICE curves and resulting global PD plots is the extrapolation problem when features are correlated. This problem and how it affects GADGET for different feature effect methods, will be explained in Section \ref{sec:illustr_comp}.

\subsection{GADGET-ALE}
\label{sec:ale}

While ALE curves avoid the issue of extrapolation seen in PD curves, they do not directly entail a local feature effect visualization that shows the heterogeneity induced by feature interactions, as seen in ICE curves for PD plots. 
However, ALE plots are based on derivatives of the prediction function regarding the feature of interest $\xv_j$ (see Eq.~\ref{eq:ale}), which serve as local feature effects to capture the interaction-related heterogeneity. We use those derivatives within GADGET-ALE to receive more representative ALE curves in the final regions, yielding the following choice for the local feature effects:
$\textstyle h(x_j,  \xv_{-j}^{(i))}) = \frac{\partial \fh(x_j, \xv_{-j}^{(i)})}{\partial x_j}$.

\begin{theorem}
    The local feature effects of GADGET-ALE satisfy the local decomposability axiom (Axiom \ref{axiom:feat_rel}). The proof is given in Appendix \ref{app:proof_ale}.
    \label{theorem:axiom_ale}
\end{theorem}

The derivatives in ALE are typically estimated by defining $m$ intervals for feature $\xv_j$ and quantifying for each interval the prediction difference between the upper and lower boundary for observations lying in this interval (see Eq.~\ref{eq:ale_est}). Hence, the loss function in Eq.~\eqref{eq:loss} measures the variance of the derivatives of observations where the feature values of $\xv_j$ lie within the boundaries of the regarded interval and, thus, measures the interaction-related heterogeneity of local effects of feature $\xv_j$ within this interval. The risk function in Eq.~\eqref{eq:risk} then aggregates this loss over all relevant intervals.
Equivalently to PD plots, ALE plots can be decomposed additively into main and interaction effects. Furthermore, if all feature interactions of features in $S$ can be completely minimized (see Theorem \ref{corollary:objective}), the joint mean-centered ALE function $f^{ALE, c}_{S|\mathcal{A}_g}$ within each final subspace $\mathcal{A}_g$ can be decomposed into the 1-dimensional mean-centered ALE functions of features in $S$ (see Eq.~\ref{eq:decomp_fe}). More details on the decomposition including an illustrative example can be found in Appendix \ref{app:decomp_ale}.

\subsection{GADGET-SD}
\label{sec:shap}
While PD and ALE plots visualize the global feature effect for which we define the appropriate local feature effect function for the GADGET algorithm, the SD plot (Section \ref{sec:background_fe}) is comparable to an ICE plot in the sense that it shows all local feature effects instead of an aggregated global feature effect. Here, we provide an estimate for the global effect and show the applicability of SD within GADGET which we term GADGET-SD.

\cite{herren_statistical_2022} showed that the Shapley value $\phi_j^{(i)}(x_j)$ of observation $i$ for feature value $x_j$ can be decomposed to 
\begin{equation}
    \phi_j^{(i)}(x_j) = g_j^c(x_j) + \sum_{k=1}^{p-1}\frac{1}{k+1}\sum_{W\subseteq -j, |W| = k} g_{W\cup j}^c(x_j,\xv_W^{(i)}),
    \label{eq:shap_decomp}
\end{equation}
with $\textstyle g_{W\cup j}^c(x_j,\xv_W^{(i)}) = \E[\hat f(x_j, X_{-j})|X_W = \xv_W^{(i)}] + \sum\limits_{k=0}^{|W|-1} \sum\limits_{V \subset \{W\cup j\}, |V|=k} (-1)^{|W|-k} \E[\hat f(X)|X_V = \xv_V^{(i)}]$. These terms correspond to the pure feature interaction effects of feature $\xv_j$ as defined by the standard functional ANOVA decomposition \citep{hooker_discovering_2004}. For more details regarding the functional ANOVA representation of Shapley values, the interested reader is referred to \cite{herren_statistical_2022}.
This decomposition satisfies Axiom \ref{axiom:feat_rel} and thus we choose $h = \phi_j^{(i)}$ for GADGET-SD.
\begin{theorem}
    The Shapley values satisfy the local decomposability axiom (Axiom \ref{axiom:feat_rel}). The proof can be found in Appendix \ref{app:proof_sd}.
    \label{theorem:axiom_sd}
\end{theorem}

It is important to note that the Shapley value is defined via expected values and thus its definition differs from those of ICE curves and derivatives for ALE, which are based on local predictions. Similarly to estimating the mean-centering constants for ICE curves, it follows that the expected values within the Shapley value estimation must consider the regarded subspace $\mathcal{A}_g$
to acknowledge the full heterogeneity reduction due to interactions within each subspace. This means that we must recalculate the Shapley values after each partitioning step for each new subspace to receive regional SD effects in the final subspaces that are representative of the underlying main effects within each subspace. However, without recalculating the conditional expected values, we still minimize the unconditional expected value (i.e., the feature interactions on the entire feature space). 
The two different approaches lead to a different acknowledgment of interaction effects. In general, it can be said that the faster approach without recalculation will be less likely to detect feature interactions of higher order compared to the exact approach with recalculation. The differences of the two approaches are explained in Appendix \ref{app:decomp_shap} and further discussed on a more complex simulation example in Section \ref{sec:sim_higher}.

Based on Eq.~\eqref{eq:shap_decomp}, the global feature effect for the SD plot of feature $\xv_j$ at the feature value $x_j$ can be defined by
\begin{equation}
f^{SD}_j(x_j) = \E_{X_W}[\phi_j(x_j)]   = g_j^c(x_j) + \sum_{k=1}^{p-1}\frac{1}{k+1}\sum_{W\subseteq -j: |W| = k} \E[g_{W\cup j}^c(x_j,X_W)].
\label{eq:shap_global}
\end{equation}

While there exist estimators for the global effect in PD and ALE plots, a pendant for the SD plot has not been introduced yet. Hence, a suitable estimator to estimate the expected value of Shapley values of feature $\xv_j$ in Eq.~\eqref{eq:shap_global} must be chosen. Here, we use univariate GAMs with splines to estimate the expected value.
Thus, the estimated GAM for feature $\xv_j$ represents the regional SD feature effect of feature $\xv_j$ within subspace $\mathcal{A}_g$ (see Section~\ref{sec:illustr_comp}).

The decomposition of Shapley values differs from that of mean-centered ICE curves in the way that feature interactions in Shapley values are fairly distributed between involved features, which leads to decreasing weights the higher the order of the interaction effect (see Eq.~\ref{eq:shap_decomp}). In contrast, all interaction terms receive the same weight as the main effects in ICE curves. Hence, interactions of high order lead to less heterogeneity in SD plots compared to ICE plots. 
However, by using the marginal-based approach to calculate Shapley values, they can be decomposed by weighted PD functions \citep[see Eq.~\ref{eq:shap_decomp} and][]{herren_statistical_2022}. Hence, the same decomposition rules as defined for PD plots apply for the SD feature effect, as defined in Eq.~\eqref{eq:shap_global}.
Meaning, 
the regional joint SD effect of features in $S$ can be decomposed into 1-dimensional regional SD effect functions, as in Eq.~\eqref{eq:decomp_fe} if all interaction-related heterogeneity is reduced. In this special case and if Shapley values are estimated by the marginal-based approach, it can be shown that $f^{SD,c}_{j|\mathcal{A}_g} = f^{PD,c}_{j|\mathcal{A}_g}$ (see Appendix \ref{app:decomp_shap} for more details and an illustration in a simulated example).  

\subsection{Illustrative Comparison of Feature Effect Methods Within GADGET}
\label{sec:illustr_comp}

\begin{figure}[t]
    \centering
    \includegraphics[width=0.65\textwidth]{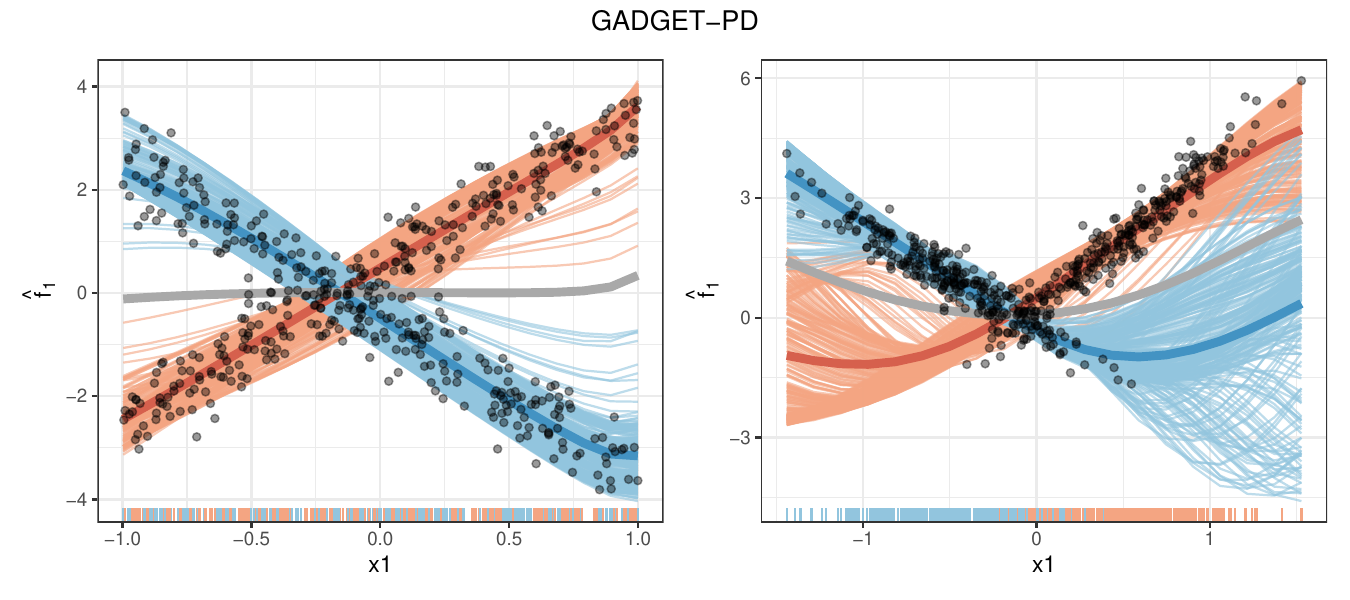}
    \includegraphics[width=0.65\textwidth]{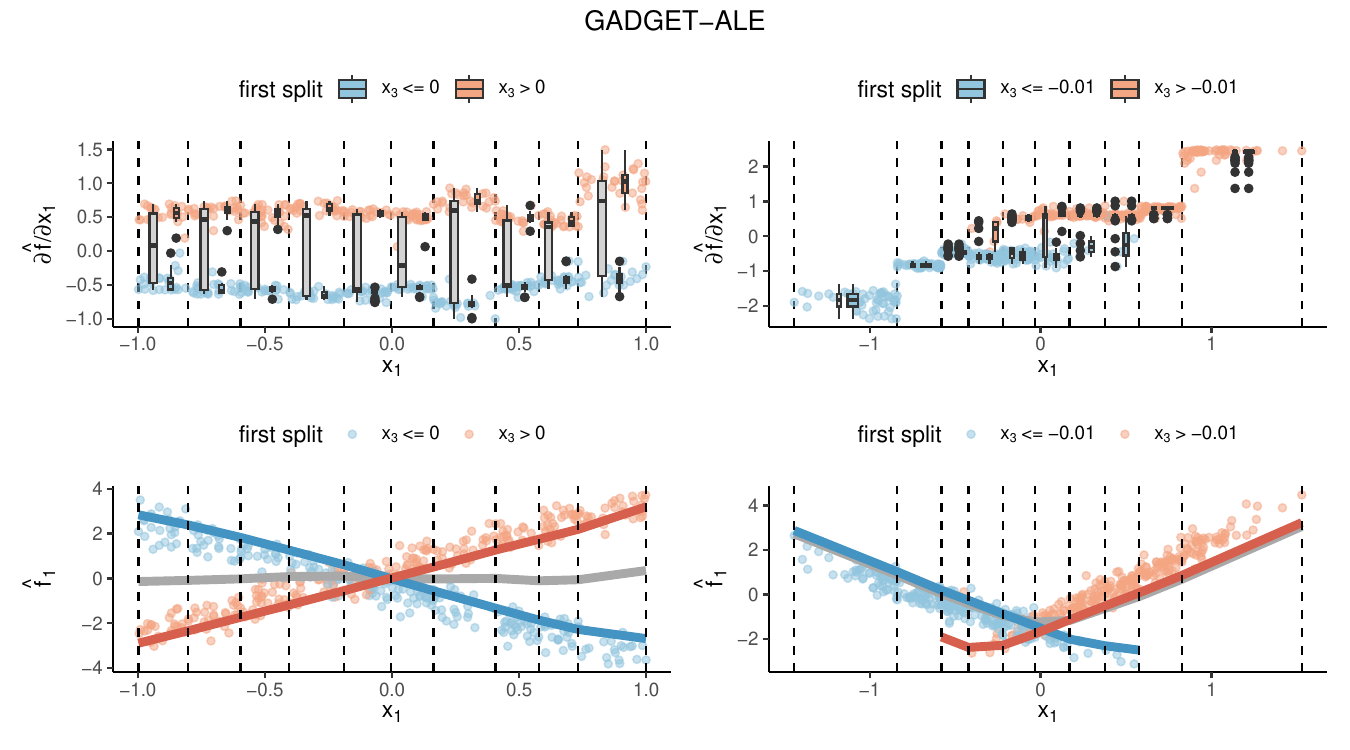}
    \includegraphics[width=0.65\textwidth]{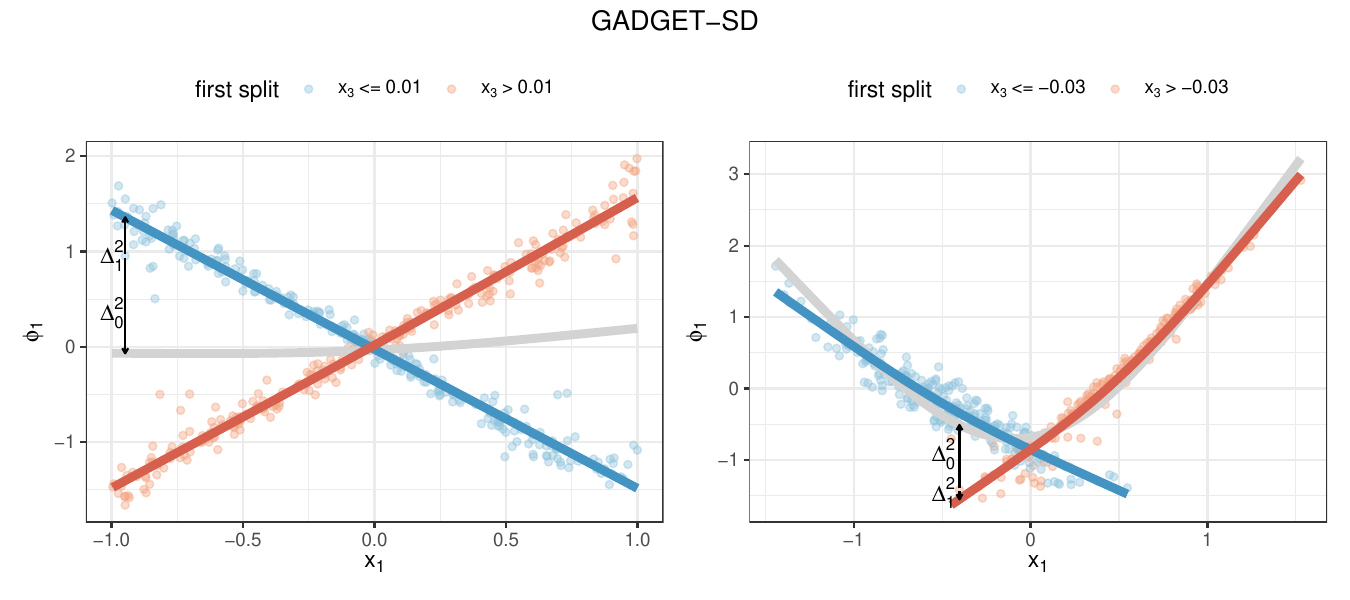}
    \caption{Local, global (grey), and regional effects for uncorrelated (left) and correlated (right) simulation settings. Local effects are colored w.r.t. the first split when GADGET is applied. The split feature is in all cases $\xv_3$. The blue color represents the left and orange the right region according to the split point. The thicker lines represent the respective regional PD curves. The rug plot and the black points in the upper plots show the distribution of $\xv_1$ according to the split point and the underlying observational values, respectively. The two upper ALE plots visualize the heterogeneity of local feature effects (derivatives) while the two lower plots show the mean-centered global and regional ALE-curves.}
    \label{fig:repid_extrapol}
    
\end{figure}

We now illustrate the differences between the three discussed feature effect methods and how their underlying assumptions affect the resulting regional effects when GADGET is applied. 
We consider the simulation example from Section \ref{sec:pdp} with the true underlying relationship: $Y = 3X_1\mathbbm{1}_{X_3>0} - 3X_1\mathbbm{1}_{X_3\leq 0} + X_3 + \epsilon$. Here, all three features are jointly independent. The upper left plot in Figure \ref{fig:repid_extrapol} shows the ICE, global PD, and regional PD curves for feature $\xv_1$ similar to Figure \ref{fig:pdp_dt}. The split found at $\xv_3 \approx 0$ is optimal according to the true functional relationship of the data-generating process. 
The regional PD plots reflect the contradicting influence of feature $\xv_1$ on the predictions due to the underlying feature interaction with $\xv_3$ compared to the global PD plot. The respective split reduces the interaction-related heterogeneity almost completely (by $98\%$). Hence, the resulting regional marginal effects of $\xv_1$ can be approximated well by the respective main effects (regional PD) and are more representative of the individual observations within each region.

One main disadvantage of ICE and PD plots is the extrapolation problem in the presence of feature interactions and correlations, as described in Section \ref{sec:background_fe}. To illustrate how this problem affects GADGET-PD, we consider the same simulation example as before, but with $X_1 = X_3 + \delta$ and $\delta \sim \mathbb{N}(0, 0.0625)$. We again draw 500 observations and train an NN with the same specification and receive an $R^2$ of $0.92$ on a separately drawn test set of size $10000$ following the same distribution. Thus, the high correlation between $\xv_1$ and $\xv_3$ barely influences the model performance. However, the upper right plot of Figure \ref{fig:repid_extrapol} clearly shows that ICE curves are extrapolating in low-density regions. The rug plot on the bottom indicates the distribution of feature $\xv_1$ depending on feature $\xv_3$. Thus, the model was not trained on observations with small $\xv_1$ values and simultaneously large $\xv_3$ values, and vice versa. However, since we integrate over the marginal distribution, we also predict in these ``out-of-distribution'' areas.
This leads to the so-called extrapolation problem and, hence, to uncertain predictions in extrapolated regions that must be interpreted with caution. 

The four plots in the middle of Figure \ref{fig:repid_extrapol} illustrate the results for the same uncorrelated and correlated settings using GADGET-ALE. 
The upper plots show that the heterogeneity of the local effects (derivatives) before applying GADGET is very high, spanning across negative to positive values (grey boxplots) within each interval. With GADGET, we then partition the feature space w.r.t. one of the features in $Z$ such that this interaction-related heterogeneity is minimized.
GADGET-ALE finds the correct split feature and approximately the correct split point for both the uncorrelated and the correlated cases, and this split strongly reduces the heterogeneity of the local effects. 
While the shapes of the centered ALE curves look very similar to PD plots for the uncorrelated case, the ALE curves for the correlated case do not extrapolate and, thus, are more representative of the feature effect with regard to the underlying data distribution.

The lower plots in Figure \ref{fig:repid_extrapol} visualize the SD plots for feature $\xv_1$. 
Similarly to the least-square estimate in linear regression, we search for the GAM that minimizes the squared distance (see $\Delta^2$ in two bottom plots in Figure \ref{fig:repid_extrapol}) of the Shapley values of $\xv_1$.
With GADGET, we now split such that the fitted GAMs within the two new subspaces minimize the squared distances between them and the Shapley values within the respective subspace. Since the GAMs are fitted on the Shapley values (local feature effects), in contrast to PDs, they do not extrapolate regarding $\xv_1$ in the correlated scenario. However, Shapley values are based on expected values that must be estimated. If they are calculated using the marginal-based approach (as we do here), it is still possible that the predictions considered in the Shapley values extrapolate into sparse regions.
Another difference to ICE curves that becomes visible in Figure \ref{fig:repid_extrapol} is the allocation of feature interactions. While the interaction effect of size $3$ is fully attributed to both involved features for ICE curves, it is fairly distributed between $\xv_1$ and $\xv_3$ for Shapley values and thus the slopes of the regional effects for feature $\xv_1$ are halved compared to the regional PD curves. In which way this difference affects the GADGET algorithm will be analyzed in Section \ref{sec:sim_higher}.

Definitions and descriptions of the mean and interaction-related heterogeneity estimates and visualization techniques for each feature effect method can be found in Appendix \ref{app:overview_estimates}.

\subsection{Quantifying Feature Interactions}
\label{sec:quant_feat_int}
In addition to 
visualizing regional effects, the GADGET algorithm quantifies feature interactions, for which we introduce several measures \citep[inspired by][]{herbinger2022repid}.
%
We quantify how much interaction-related heterogeneity has been reduced in a single split of GADGET for a considered feature $j \in S$,
\begin{equation}
    I(\mathcal{A}_P, \tilde\xv_j) = \frac{\mathcal{R}_j(\mathcal{A}_P, \tilde\xv_j) - \mathcal{R}_j(\mathcal{A}_l, \tilde\xv_j) - \mathcal{R}_j(\mathcal{A}_r, \tilde\xv_j)}{\mathcal{R}_j(\mathcal{X}, \tilde\xv_j)},
    \label{eq:int_index_node}
\end{equation}
where $P$ is the parent node and $l$ and $r$ are its left and right children, respectively.
We express risk reduction via the split
relative to the risk on the entire feature space (root node).
%
%
Instead of considering only a single partitioning step, one might be more interested in how much interaction-related heterogeneity reduction a specific split feature $z \in Z$ is responsible for w.r.t. all performed partitioning steps of an entire tree. 
For a given splitting feature $z$ we now simply sum up the reduction terms for all splits in which $z$ occurs, which we denote by $\mathcal{B}_z \subset \{\mathcal{A}_1,\ldots, \mathcal{A}_G\}$, where $G$ is the number of splitting nodes in the tree:

\begin{equation}
    I_{z,j}(\tilde\xv_j) = \sum_{\mathcal{A}_P \in \mathcal{B}_z} I(\mathcal{A}_P, \tilde\xv_j).
    \label{eq:int_index_feat_j}
\end{equation}

We obtain the overall interaction-related heterogeneity reduction $I_z$ of the $z$-th split feature by relating the risk reduction of all its splits over all considered features in $S$ to the total risk of the root node: 
\begin{equation}
    I_z = \frac{ \sum_{\mathcal{A}_P \in \mathcal{B}_z}\sum_{j\in S} (\mathcal{R}_j(\mathcal{A}_P, \tilde\xv_j) - \mathcal{R}_j(\mathcal{A}_l, \tilde\xv_j) - \mathcal{R}_j(\mathcal{A}_r, \tilde\xv_j))}{\sum_{j\in S}\mathcal{R}_j(\mathcal{X}, \tilde\xv_j)}.
    \label{eq:int_index_feat}
\end{equation}
For example, in our previous example in Figure~\ref{fig:pdp_dt}, the interaction-related heterogeneity reduction of the first split ($I(\mathcal{X}, \tilde\xv_1)$) for feature $\xv_1$ is $0.986$, which means that almost all of the interaction-related heterogeneity of $\xv_1$ is reduced after the first split.
Furthermore, we obtain $I_{3,1}(\tilde\xv_1) = I(\mathcal{X}, \tilde\xv_1) = 0.986$, since $z = 3$ was only used once for splitting, while the interaction-related heterogeneity reduction of all three features is $I_3 = 0.99$ for $z = 3$.

\paragraph{\normalfont\textit{Goodness of Fit.}} A further aggregation level would be to sum up $I_{z,j}(\tilde\xv_j)$ over all $z \in Z$ and, thus, obtain the interaction-related heterogeneity reduction for feature $\xv_j$ between the entire feature space and the terminal nodes. This is related to the concept of $R^2$, which is a well-known measure in statistics to quantify the goodness of fit. We apply this concept here to quantify how well the terminal regional effect curves (in the terminal subspaces $\mathcal{B}_t \subset \{\mathcal{A}_1,\ldots, \mathcal{A}_G\}$) fit the underlying local effects compared to the global feature effect curve on the entire feature space. We distinguish between the feature-related $R^2_j$, which represents the goodness of fit for the feature effects of feature $\xv_j$:
\begin{equation}
    R^2_j = \sum_{z \in Z} I_{z,j}(\tilde\xv_j) = 1- \frac{\sum_{\mathcal{A}_t \in \mathcal{B}_t} \mathcal{R}_j(\mathcal{A}_t, \tilde\xv_j)}{\mathcal{R}_j(\mathcal{X}, \tilde\xv_j)},
    \label{eq:Rsquared_feat}
\end{equation}
and the $R^2_{Tot}$, which quantifies the goodness of fit for the feature effects of all features in $S$: 
\begin{equation}
    R^2_{Tot} =  \sum_{z \in Z} I_z = 1- \frac{\sum_{\mathcal{A}_t \in \mathcal{B}_t} \sum_{j \in S} \mathcal{R}_j(\mathcal{A}_t, \tilde\xv_j)}{\sum_{j \in S}\mathcal{R}_j(\mathcal{X}, \tilde\xv_j)}.
    \label{eq:Rsquared_total}
\end{equation}
The right-hand versions of both formulas follow from the cancellation of parent vs. children terms (except for root and terminal nodes) when the left-hand version is summed across the whole tree. 
Both $R^2$ measures take values between 0 and 1, with values close to 1 signaling that almost all heterogeneity in the final subspaces compared to the entire feature space has been reduced---either for a specific feature of interest ($R^2_j$) or for all features of interest ($R^2_{Tot})$.
In our example, $R_1^2$ for feature $\xv_1$ is the same as for $I_{3,1}(\xv_1)$, since GADGET performed only one split. The total interaction-related heterogeneity reduction over all features in $S$ is $R_{Tot}^2 = I_3 = 0.99$. Hence, the interaction-related heterogeneity of all features in $S$ has been reduced by $99\%$ after the first split.

\section{A Procedure to Detect Global Feature Interactions}
\label{sec:interact_pimp}

%
As shown in Section \ref{sec:quant_feat_int}, GADGET quantifies feature interactions between the feature subsets $S$ and $Z$. 
Choosing the features to include in $S$ and $Z$ depends on the underlying research question. 
If the user is interested in how a specific set of features ($S$) influences the model's predictions depending on another user-defined feature set ($Z$), 
then $S$ and $Z$ are chosen based on domain knowledge. However, the user may not know which relationships the ML model has learned. Thus, choosing $S$ and $Z$ based on domain knowledge does not ensure consideration of all interacting features. 
%
If our goal is to minimize feature interactions between all features to potentially additively decompose the prediction function into univariate effects, we can define $S = Z = \{1,\ldots,p\}$. With that choice, we aim to reduce the overall interaction-related heterogeneity in all features ($S$), since we also consider all features to be possible interacting features ($Z$). 
For high-dimensional settings, this has the disadvantage of being computationally expensive and we might ``detect'' spurious interactions and therefore produce less stable regional splits.
Hence, we must define beforehand the subset of features that actually interact. 
Given that features typically interact with each other and all involved features will usually show heterogeneity in their local effects while being responsible for the heterogeneity of other involved features, we will choose $S = Z$. 
Thus, we will only use $S$ as the globally interacting feature subset to be defined.


\subsection{The PINT Procedure}
We introduce a new permutation-based interaction detection (PINT) procedure to define the interacting feature subset $S$. PINT is based on the idea of a non-parametric permutation test \citep{ernst2004permutation}. 
It can be applied to any feature effect method satisfying Axiom~\ref{axiom:feat_rel}. 
%

With the risk function defined in Eq.~\eqref{eq:risk} and based on the chosen local feature effect method, we can quantify the interaction-related heterogeneity of each feature $\xv_j$ within the feature space $\mathcal{X}$. Due to correlations between features, the estimated heterogeneity might also include spurious interactions. Thus, we must define a null distribution to determine which heterogeneity is actually due to feature interactions (i.e., significant w.r.t. the null distribution) and which heterogeneity is due to other reasons, such as correlations or noise. 
Thus, the null distribution needs to retain the underlying properties of the data. 

Generally, we do not know the true functional relationship and there are infinitely many distributions compliant with a null hypothesis of no interaction-related heterogeneity: $\mathcal{R}_j(\mathcal{X}, \tilde\xv_j) = 0$. We use the prediction model and a permutation-based approach to approximate the variability of the risk $\mathcal R_j$ under this null hypothesis. We follow a similar approach to \cite{altmann_2010_pimp} who suggested a permutation-based null distribution to detect important features. To obtain the null distribution, the target variable $y$ is permuted (line 4 in Algorithm \ref{alg:pint}), which breaks the association between the target and the features while keeping the feature distribution intact. Hence, no more interactions are present between $\xv_j$ and $\xv_{-j}$, but the association between the features remains. Thus, when performing a refit on the data set $\tilde{\mathcal{D}}$ with the permuted target variable, potential heterogeneity of local effects is only due to randomly learned effects of the prediction model depending on the feature distribution and noise and not caused by true feature interactions. Therefore, calculating $\tilde{\mathcal{R}}_j$ based on $\tilde{\mathcal{D}}$ measures the risk of $\xv_j$ under the null (lines 5-8 in Algorithm \ref{alg:pint}). To obtain the null distribution of risk values for $\xv_j$, this procedure is repeated $s$ times.\footnote{An alternative approach would be to randomly permute $\xv_j$ instead of $y$ to break the interaction effects between $\xv_j$ and $\xv_{-j}$. However, this approach is computationally more expensive and does not only break the respective interaction effects but also the correlations between $\xv_j$ and $\xv_{-j}$, and thus spurious interactions will not be detected. Using conditional samples \citep{molnar_model-agnostic_2021} or knock-offs \citep{watson2021testing} instead of random permutations solves the second problem. However, this approach is even more computationally expensive and becomes infeasible with a higher number of features.}

Since we are interested in defining the feature subset $S$ of all interacting features, we calculate $\tilde{\mathcal{R}}_j$ for all features $j \in \{1,\ldots, p\}$. Since all feature interactions between $y$ and $\xv$ are broken by permuting $y$, the null distribution of risk values can be estimated for all features by the described $s$ permuting and refitting steps. To determine the interacting feature subset $S$, the respective risk value $\mathcal{R}_j(\mathcal{X}, \tilde\xv_j, \fh_\mathcal{D})$, is calculated based on the unpermuted data set $\mathcal{D}$ and the originally fitted model on this data set. The value of $\mathcal{R}_j(\mathcal{X}, \tilde\xv_j, \fh_\mathcal{D})$ is then compared to the $(1-\alpha)$-quantile, i.e., $z^{1-\alpha}_j$, of the non-parametric null distribution. 
If the risk value of $\xv_j$ exceeds the respective quantile, then feature interactions between $\xv_j$ and $\xv_{-j}$ are considered to be highly relevant (lines 11-13 in Algorithm \ref{alg:pint}).

 
\begin{algorithm}[t]
    \caption{PINT}
    \label{alg:pint}
    \begin{algorithmic}[1]
      \STATE \textbf{input:} data set $\mathcal{D}$, prediction function $\fh$, number of permutations $s$,\\ risk function $\mathcal{R}_j$, significance level $\alpha$ 
      \STATE \textbf{output: } feature subset $S$
      \FOR{$k \in \{1,\ldots, s\}$}{
      \STATE permute target $y$ of $\mathcal{D}$ and obtain permuted $\tilde{y}^{k}$ and $\tilde{\mathcal{D}}^{k}$
      \STATE refit model on $\tilde{\mathcal{D}}^{k}$ to obtain the prediction function $\tilde{f}^{k}_{\tilde{\mathcal{D}}^{k}}$
        \FOR{$j \in \{1,\ldots, p\}$}{
            \STATE calculate risk $\tilde{\mathcal{R}}_j^{k} =\mathcal{R}_j^{k}(\mathcal{X}, \tilde\xv_j, \tilde{f}^{k}_{\tilde{\mathcal{D}}^{k}})$ for the $j$-the feature achieved by $\tilde{f}^{k}_{\tilde{\mathcal{D}}^{k}}$
        }\ENDFOR
      } \ENDFOR   
       \FOR{$j \in \{1,\ldots, p\}$}{
            \STATE calculate risk $\mathcal{R}_j(\mathcal{X}, \tilde\xv_j, \fh_\mathcal{D})$ for $j$-the feature based on $\mathcal{D}$ and $\fh$
            \STATE sort (increasing) all $\tilde{\mathcal{R}}_j^{k}$ , set $(1-\alpha)$-quantile  $z^{1-\alpha}_j = \tilde{\mathcal{R}}_j^{\lceil (s + 1) (1-\alpha) \rceil}$
            \STATE add $j$ to $S$, if and only if $\mathcal{R}_j(\mathcal{X}, \tilde\xv_j, \fh_\mathcal{D}) > z^{1-\alpha}_j$
        }\ENDFOR
    \end{algorithmic}
\end{algorithm}

\subsection{Theoretical Properties}
\label{sec:pint_validation}

We emphasize that PINT is a detection \emph{procedure}, not a formal statistical test. This distinction is necessary because the permutation approach generates data from only one of the many possible distributions compliant with the null hypothesis. We may nevertheless ask how the procedure's type I and type II errors behave. In the following, we give conditions under which PINT is valid as a statistical test. An empirical validation is given in Section \ref{sec:sim_pint_valid} and Appendix \ref{app:pint}.

A difficulty in formally analyzing the procedure is that it depends on the unspecified ML algorithm that induces $\hat f$. We can emphasize this by writing $\mathcal{R}_j(\mathcal{X}, \tilde\xv_j, \fh_{\mathcal{D}_n})$ for the risk for feature value 
$\xv_j$ achieved by the algorithm $\fh$ run on a data set\footnote{With a slight abuse of notation, we use $\mathcal{D}_n$ to define the data set as a random variable.} $\mathcal{D}_n = \{(X^{(i)}, Y^{(i)})\}_{i = 1}^n$, generated i.i.d. from $\P_{X, Y}$. Define $f(\xv) = \E[Y \mid X = \xv]$ and  $\mathcal{D}^\Pi_n = \{(X^{(i)}, Y_{\Pi(i)}^{(i)})\}_{i = 1}^n$ where $\Pi$ is a random permutation of $\{1, \dots, n\} $ independent of $\mathcal{D}_n$. We can now impose abstract, but interpretable conditions on the ML algorithm under which the test is valid. 
\begin{theorem}  \label{thm:typeI}
	Suppose $\P_{X, Y}$ is a probability measure where $\mathcal{R}_j( \mathcal{X}, \tilde\xv_j, f) = 0$.
	The type I error probability of PINT is upper bounded by $\alpha + \epsilon$,  
	where
	\begin{align}
		\epsilon = \max_{t > 0} \biggl[\P\biggl(\mathcal{R}_j(\mathcal{X}, \tilde\xv_j, \fh_{\mathcal{D}_n}) > t\biggr) - \P\biggl( \mathcal{R}_j( \mathcal{X}, \tilde\xv_j, \fh_{\mathcal{D}_n^\Pi})  > t\biggr)\biggr].
	\end{align}
  The proof can be found in Appendix \ref{app:pint_proof}.
\end{theorem}
Intuitively,  $\mathcal{D}_n^\Pi$ is a data set with no feature effects. The extra term $\epsilon$ is zero if the algorithm $\fh$ is less likely to find a spurious interaction in unpermuted data compared to a case where no effects are present. Often, this is natural: most learners detect less of the spurious effect when there are other effects to focus on. In this case, the PINT procedure is a valid statistical test. In fact, we conjecture that the test is conservative (i.e., type I error is strictly smaller than $\alpha$) in most cases, which is in line with our empirical observations (see Section \ref{sec:sim_pint_valid} and Appendix \ref{app:pint}).
However, $\epsilon$ may be strictly positive when features are strongly dependent, and the spurious effect is easier to learn than the true one. An example of such a case is given in Appendix \ref{app:sim_pint_invalid}.
\begin{theorem} \label{thm:typeII}
	Suppose $\P_{X, Y}$ is a probability measure where $\mathcal{R}_j(\mathcal{X}, \tilde\xv_j, f) \neq 0$. 
	If the algorithm $\fh$ is such that for every $\epsilon > 0$
	\begin{align} \label{cond:2-1}
		\P\biggl(\sup_x|\fh_{\mathcal{D}_n^\Pi}(\xv) - \E[Y]| > \epsilon \biggr) \to 0
	\end{align}
	and there exists $c > 0$ such that
	\begin{align}  \label{cond:2-2}
	\P\biggl(\mathcal{R}_j(\mathcal{X}, \tilde\xv_j, \fh_{\mathcal{D}_n}) > c \biggr) \to 1,
	\end{align}
	the type II error probability tends to 0. The proof can be found in Appendix \ref{app:pint_proof}.
\end{theorem}
Recall that $\mathcal{D}_n^\Pi$ is a data set with no feature effects, in which case we would have $f(\xv) = \E[Y]$ for all $\xv$.
Condition $\eqref{cond:2-1}$ essentially requires the inducing algorithm of $\fh$ to perfectly learn constant functions. When there are feature effects, the algorithm does not have to be perfect, though. Condition \eqref{cond:2-2} merely requires it to capture at least some of the interaction effect between $x_j$ and other features.

\section{Empirical Validation and Analyses of GADGET}
\label{sec:sim}

We now investigate different hypotheses to (1) empirically validate that GADGET generally minimizes feature interactions and (2) show how different characteristics of the underlying data---like feature correlations and true functional relationships---might influence GADGET, depending on the chosen feature effect method.
The structure of the following sections and the concrete definition of the hypotheses is based on (2). However, the simulation examples themselves are designed in such a way that we know the underlying ground-truth of feature interactions for each example. Thus, we are able to empirically validate that GADGET generally minimizes feature interactions.

\paragraph{\normalfont\textit{Definition of Stop Criteria for GADGET.}}
In all our experiments and real-world applications, we control the number of partitioning steps performed by GADGET based on (1) the tree depth and the minimum number of observations per leaf node which are hyperparameters typically used by decision trees, and (2) an early stopping mechanism where a further split is only performed if the relative improvement of the split to be performed is at least $\gamma \in [0,1]$ times the total relative interaction-related heterogeneity reduction of the previous split:
$\textstyle \gamma\times\frac{\sum_{j \in S} \left(\mathcal{R}_j(\mathcal{A}_P, \tilde\xv_j)\right) -\mathcal{I}(\hat t,\hat z)}{\sum_{j \in S} \left(\mathcal{R}_j(\mathcal{X}, \tilde\xv_j)\right)}$.
The higher we choose $\gamma$, the fewer partitioning steps will be performed. See Appendix \ref{app:recom_stop_crit} for more details and recommendations.

\subsection{Influence of Correlated Features}
\label{sec:sim_extrapol}
\paragraph{\normalfont\textit{Hypothesis.}}
With increasing correlation between features, results become less stable due to extrapolation for methods using the marginal distribution for integration (especially PD, as shown in Section \ref{sec:illustr_comp}, but also SD) compared to methods using the conditional distribution (such as ALE). 
In this context, ``stability'' denotes the ability of the different GADGET variants to find the correct split feature and split point over various repetitions.

\paragraph{\normalfont\textit{Experimental Setting.}}
We use the (simple) simulation example of Section \ref{sec:new_framework} for four different correlation coefficients $\rho_{13}$ between $X_1$ and $X_3$: $0, 0.4, 0.7,$ and $0.9$.
The data is generated as follows: 
Let $X_2, X_3 \sim \mathbbm{U}(-1,1)$ be independently distributed and $X_1 = c \cdot X_3 + (1-c)\cdot Z$ with $Z \sim \mathbbm{U}(-1,1)$, where $c$ is chosen between $0$ and $0.7$, to achieve the $\rho_{13}$ values. The true underlying relationship is defined as before by $Y = 3X_1\mathbbm{1}_{X_3>0} - 3X_1\mathbbm{1}_{X_3\leq 0} + X_3 + \epsilon$ with $\epsilon \sim \mathbb{N}(0,0.09)$. We draw $1000$ observations and fit a GAM with correctly specified main and interaction effects and an NN with the previously defined specifications. We repeat the experiment $30$ times.
We apply GADGET to each setting and model within each repetition using PD, ALE, and SD as feature effect methods and set $S = Z = \{1, 2, 3\}$. As stopping criteria, we choose a maximum tree depth of 6, a minimum number of observations per leaf of 40, and set the improvement parameter $\gamma$ to $0.2$.

\begin{figure}
    \centering
    \includegraphics[width=0.9\textwidth]{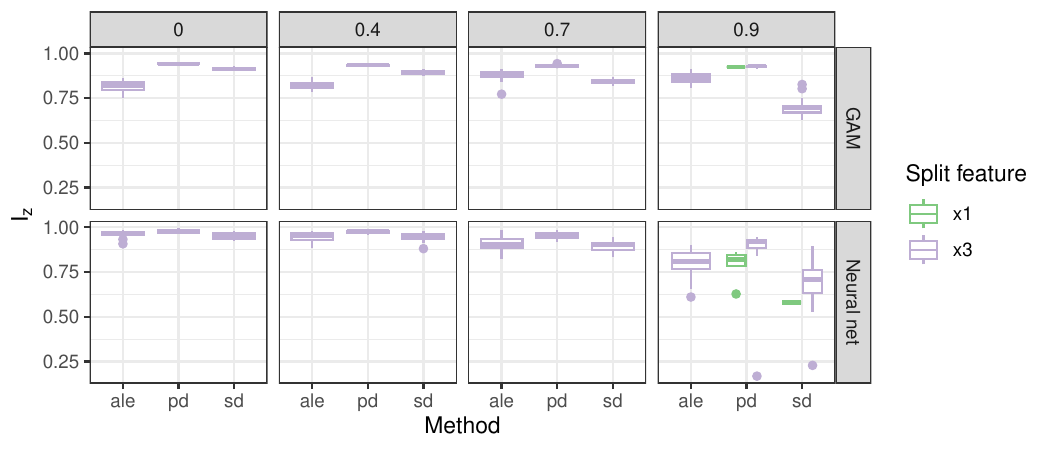}
    \caption{Boxplots showing the interaction-related heterogeneity reduction $I_z$ per split feature over $30$ repetitions when PD, ALE, or SD is used in GADGET. Columns refer to correlation $\rho_{13}$, rows refer to fitted ML model.}
    \label{fig:sim_xor}
\end{figure}

\paragraph{\normalfont\textit{Results.}}
Figure \ref{fig:sim_xor} shows that
independent of the model or correlation degree, $\xv_2$ has (correctly) never been considered as split feature. For correlation strengths $\rho_{13}$ between $0$ and $0.7$, $\xv_3$ is always chosen as the only split feature, with $I_3$ taking values between $0.75$ and $1$ and thus reducing most of the interaction-related heterogeneity of all features with one split. Thus, for low to medium correlations, there are only minor differences between the various feature effect methods and models. However, the observed behavior changes substantially for $\rho_{13} = 0.9$. For the correctly specified GAM, we still receive fairly consistent results, apart from one repetition where $\xv_1$ is chosen as the split feature when PD is used. In contrast, there is more variation of $I_3$ for the NN. For SD, $\xv_1$ is chosen once for splitting, and for PD, this is also the case for $30\%$ of all repetitions (see Table \ref{tab:xor}). Using ALE, GADGET always correctly performs one split with respect to $\xv_3$. Additionally, GADGET performs a second split once when PD is used and 7 times when SD is used.
Thus, for high correlations between features, we already observe in this very simple setting that the extrapolation problem influences the splitting within GADGET for effect methods based on marginal distributions, while methods based on conditional distributions such as ALE are less affected. This is particularly relevant for learners that model very locally (e.g. NNs) and, thus, tend to have wiggly prediction functions and oscillate in extrapolating regions. This is also notable in the split value range, which increases for the NN in the case of high correlations for PD and SD but not for ALE (see Table \ref{tab:xor}).

\begin{table}[ht]
\centering
\begin{tabular}{|ll|rrr|rrr|c|}
\hline
& &\multicolumn{3}{c|}{Split feature $\xv_3$ in \%} & \multicolumn{3}{c|}{Split value range} & \multicolumn{1}{c|}{MSE}\\
  \hline
Model & $\rho_{13}$ & ALE & PD & SD & ALE & PD & SD & mean (sd)\\ 
  \hline
GAM & 0 & 1.00 & 1.00 & 1.00 & 0.10 & 0.10 & 0.12 & 0.329 (0.008)\\ 
  GAM & 0.4 & 1.00 & 1.00 & 1.00 & 0.13 & 0.10 & 0.10 & 0.201 (0.003)\\ 
  GAM & 0.7 & 1.00 & 1.00 & 1.00 & 0.09 & 0.10 & 0.09 & 0.137 (0.002)\\ 
  GAM & 0.9 & 1.00 & 0.97 & 1.00 & 0.11 & 0.11 & 0.12 & 0.107 (0.001)\\ \hline
  NN & 0 & 1.00 & 1.00 & 1.00 & 0.10 & 0.09 & 0.09 & 0.152 (0.018)\\ 
  NN & 0.4 & 1.00 & 1.00 & 1.00 & 0.08 & 0.08 & 0.07 & 0.124 (0.007)\\ 
  NN & 0.7 & 1.00 & 1.00 & 1.00 & 0.11 & 0.10 & 0.10 &0.111 (0.005)\\ 
  NN & 0.9 & 1.00 & 0.70 & 0.97 & 0.10 & 0.19 & 0.15 & 0.104 (0.003)\\ 
   \hline
\end{tabular}
\caption{
Overview of the frequency of $\xv_3$ as the first split feature over $30$ repetitions, along with the corresponding split value range. The last column presents the model's test performance as the mean (standard deviation) of the mean squared error (MSE).
}
\label{tab:xor}
\end{table}

\subsection{Influence of Higher-Order Interaction Effects}
\label{sec:sim_higher}
\paragraph{\normalfont\textit{Hypothesis.}}
While SD puts less weight on interactions with increasing order, all interactions (regardless of the order) receive the same weight in PD and ALE. Hence, when using GADGET-SD, we may not be able to detect high order interactions---especially when using the approximation without recalculation after each split (see Section \ref{sec:shap} and Appendix \ref{app:decomp_shap}). However, it should be more likely to detect these interactions with PD and ALE.

\paragraph{\normalfont\textit{Experimental Setting.}}
To investigate this hypothesis, we consider $5$ independent features with $X_1 \sim \mathbbm{U}(0,1)$ and $X_2, X_3, X_4, X_5 \sim \mathbbm{U}(-1,1)$ and draw 1000 samples. The true functional relationship of the data-generating process is defined by a series of interactions between different features: $y = f(\xv) + \epsilon$ with $f(\xv) = \xv_1\cdot\mathbbm{1}_{\xv_3\leq 0}\mathbbm{1}_{\xv_4 > 0} + 4 \xv_1\cdot\mathbbm{1}_{\xv_3\leq 0}\mathbbm{1}_{\xv_4 \leq 0} - \xv_1\cdot\mathbbm{1}_{\xv_3 > 0}\mathbbm{1}_{\xv_5 \leq 0}\mathbbm{1}_{\xv_2 > 0} - 3 \xv_1\cdot\mathbbm{1}_{\xv_3 > 0}\mathbbm{1}_{\xv_5 \leq 0}\mathbbm{1}_{\xv_2 \leq 0} - 5 \xv_1\cdot\mathbbm{1}_{\xv_3 > 0}\mathbbm{1}_{\xv_5 > 0}$ and $\epsilon \sim \mathbbm{N}(0, 0.01\cdot var(f(\xv)))$. These interactions can be seen as one hierarchical structure between all features, where the slope of $\xv_1$ depends on the subspace defined by the interacting features (see Figure \ref{fig:tree_expl_higher}).

We fit an XGBoost (XGB) model with correctly specified feature interactions and a random forest (RF) on the data set.
Subsequently, we apply GADGET to each model using PD, ALE, SD with, and SD without recalculation after each split.
In 30 repetitions of the experiment, XGB achieved an MSE of $0.068$ ($0.009$ standard deviation) and the RF $0.121$ ($0.012$ standard deviation) on a separate test set with the same distribution.
For GADGET, we consider one feature of interest $S = 1$ and all other features as potential interacting features $Z = \{2,3,4,5\}$. As stop criteria, we set the maximum tree depth to $7$, the minimum number of observations to $40$, and $\gamma = 0.1$.
If the underlying model learned the effects of the true functional relationship of the data-generating process correctly, GADGET should split as shown in Figure \ref{fig:tree_expl_higher} to minimize the interaction-related heterogeneity of $\xv_1$. 

\paragraph{\normalfont\textit{Results.}}
All methods used $\xv_3$ as the first split feature in all repetitions.
Table \ref{tab:higher_second} shows that a second-level split was performed in only 10\% of the repetitions when GADGET-SD without recalculation is applied on the XGB model, compared to about 90\% for all other methods. A similar pattern appears with the RF model, but with more variation in the frequencies, possibly due to different learned effects. If a second-level split is performed, all methods always choose the correct split features $\xv_4$ and $\xv_5$, indicating that GADGET generally minimizes feature interactions (see Figure \ref{fig:tree_expl_higher}).
Table \ref{tab:higher_third} shows similar differences in the third-level split frequencies between SD without recalculation and other methods.
Table \ref{tab:higher_nodes} confirms that SD without recalculation shows high variation in final subspaces, with a median of 2, indicating it stops after the first split (with $\xv_3$). 
Other GADGET variants, like PD and SD with recalculation, typically reach the correct number of subspaces (5), while ALE tends to split slightly deeper.
\begin{table}[ht]
\begin{minipage}{.4\linewidth}
    \centering
    \includegraphics[width=\textwidth]{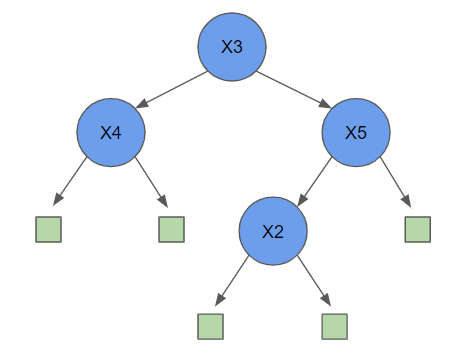}
    \captionof{figure}{Explanation of the true functional relationship of the data-generating process. The green squares represent the 5 final subspaces which contain linear effects of feature $\xv_1$.}
    \label{fig:tree_expl_higher}
\end{minipage}
\hfill
\begin{minipage}{.58\linewidth}
\centering
\begin{tabular}{|ll|rr|rr|}
\hline
&&\multicolumn{2}{c|}{$\mathcal{A}_l^{t,z}$}&\multicolumn{2}{c|}{$\mathcal{A}_r^{t,z}$}\\
  \hline
Model & Method & $z$ & Freq. &  $z$ & Freq. \\ 
  \hline
XGB & ALE & x4 & 0.93 & x5 & 0.93 \\ 
  XGB & PD & x4 & 0.90 & x5 & 0.90 \\ 
  XGB & SD\_not\_rc & x4 & 0.10 & x5 & 0.10 \\ 
  XGB & SD\_rc & x4 & 0.87 & x5 & 0.87 \\ 
  RF & ALE & x4 & 0.80 & x5 & 0.90 \\ 
  RF & PD & x4 & 0.63 & x5 & 0.73 \\ 
  RF & SD\_not\_rc & x4 & 0.03 & x5 & 0.13 \\ 
  RF & SD\_rc & x4 & 0.73 & x5 & 0.90 \\ 
   \hline
\end{tabular}
\caption{Frequencies of splits relative to root node by split feature $z$ over $30$ repetitions in the left $\mathcal{A}_l^{t,z}$ and right $\mathcal{A}_r^{t,z}$ subspaces after the first split when GADGET with ALE, PD, SD with recalculation (rc) and without recalculation (not\_rc) is used.}
\label{tab:higher_second}
\end{minipage}
\end{table}

To summarize, when SD without recalculation is used, the two-way interaction with $\xv_3$ is primarily detected,
while features of a third-order ($\xv_4$ and $\xv_5$) or fourth-order ($\xv_2$) interaction are rarely considered for splitting. In contrast, the other three methods frequently detect higher-order interactions.
This supports our hypothesis regarding higher-order effects and the theoretical differences among the feature effect methods. 
Note that recalculating the Shapley values after each split reduces the order of interactions. For example, recalculating Shapley values in the subspace $\{\mathcal{X}|\xv_3 \leq 0\}$ reduces the three-way interaction $\xv_1\cdot\mathbbm{1}_{\xv_3\leq 0}\mathbbm{1}_{\xv_4 > 0}$ to the two-way interaction $\xv_1\cdot\mathbbm{1}_{\xv_4 > 0}$ and thus the weight of the interaction increases for the next split.
As a result, the outcomes are similar to those for PD. However, recalculation can be computationally expensive.

\begin{table}[ht]
\begin{minipage}{.48\linewidth}
\centering
\begin{tabular}{|llrr|}
  \hline
Model & Method & $z$ & Rel. Freq. \\ 
  \hline
XGB & ALE & x2 & 0.93 \\ 
  XGB & PD & x2 & 0.90 \\ 
  XGB & SD\_not\_rc & x2 & 0.10 \\ 
  XGB & SD\_rc & x2 & 0.87 \\ 
  RF & ALE & x2 & 0.83 \\ 
  RF & ALE & x4 & 0.03 \\ 
  RF & PD & x2 & 0.67 \\ 
  RF & SD\_not\_rc & x2 & 0.10 \\ 
  RF & SD\_rc & x2 & 0.83 \\ 
   \hline
\end{tabular}
\caption{Frequencies of splits relative to root node by split feature $z$ over all $30$ repetitions in the subspace $\{\mathcal{X}|\xv_3>0 \cap \xv_5 \leq 0\}$ on third tree depth when GADGET is applied with ALE, PD, SD\_not\_rc and SD\_rc.}
\label{tab:higher_third}
\end{minipage}
\hfill
\begin{minipage}{.48\linewidth}
\centering
\begin{tabular}{|ll|rrr|}
\hline
&&\multicolumn{3}{c|}{No. of Subspaces}\\
  \hline
Model & Method & Min & Max & Med \\ 
  \hline
XGB & ALE &   3 &  15 & 7 \\ 
  XGB & PD &   2 &   7 & 5 \\ 
  XGB & SD\_not\_rc &   2 &   7 & 2 \\ 
  XGB & SD\_rc &   2 &   8 & 5 \\ 
  RF & ALE &   3 &  16 & 9 \\ 
  RF & PD &   2 &  11 & 5 \\ 
  RF & SD\_not\_rc &   2 &  11 & 2 \\ 
  RF & SD\_rc &   3 &  12 & 5 \\ 
   \hline
\end{tabular}
\caption{Minimum, maximum and median number of final subspaces over $30$ repetitions after GADGET is applied with ALE, PD, SD\_not\_rc and SD\_rc.}
\label{tab:higher_nodes}
\end{minipage}
\end{table}

\subsection{Influence of Spurious Interactions}
\label{sec:sim_pint}
\paragraph{\normalfont\textit{Hypothesis.}}
When using PINT to pre-select $S = Z \subseteq \{1,\ldots,p\}$, we are more likely to actually split according to feature interactions compared to when choosing all features ($S = Z = \{1,\ldots,p\}$) or using the values of the H-Statistic---which was defined in Section \ref{sec:background_int}---for pre-selection, especially when potential spurious interactions are present.

\paragraph{\normalfont\textit{Experimental Setting.}}
We reuse the simulation example described in Section \ref{sec:interact_pimp}: Consider four features with $X_1, X_2, X_4 \sim U(-1,1)$
and $X_3 = X_2 + \epsilon$ with $\epsilon \sim N(0,0.09)$. For $30$ repetitions, we draw $n = \{300, 500\}$ observations and create the dependent variable, including a potential spurious interaction between $\xv_1$ and $\xv_3$ (as $\xv_1$ and $\xv_2$ interact, but $\xv_2$ and $\xv_3$ are correlated)\footnote{While the true functional relationship is here defined without noise to purely measure the effect of the spurious interaction, we analyze the effect of adding noise in Appendix \ref{app:further_examples_pint}}: $\mathbf{y} = \xv_1 + \xv_2 + \xv_3 - 2\xv_1 \xv_2$. 
For each sample size $n$ and each repetition, we fit an SVM with an RBF kernel, cost parameter $C=1$, and choose the inverse kernel width based on the data. 
The performance of the model, measured by the mean (standard deviation) of the MSE on a test set of size 100000 across all repetitions, is \(0.028\ (0.010)\) for \(n = 300\) and \(0.027\ (0.010)\) for \(n = 500\).
We calculate PINT using PD, ALE, and SD for each repetition and sample size with $s = 100$ and $\alpha = 0.05$ by approximating the null distribution as described in Section \ref{sec:interact_pimp}.
We apply GADGET using PD, ALE, and SD with recalculation, where $S=Z$ is based on the feature subset chosen by PINT for each method. We compare these results by considering all features as features of interest and potential split features (i.e. $S = Z = \{1,\ldots,p\}$).
We use a maximum tree depth of $6$, minimum number of observations of $40$, and $\gamma = 0.15$ as stop criteria.

\begin{figure}[tbh]
    \centering
    \includegraphics[width=0.8\textwidth]{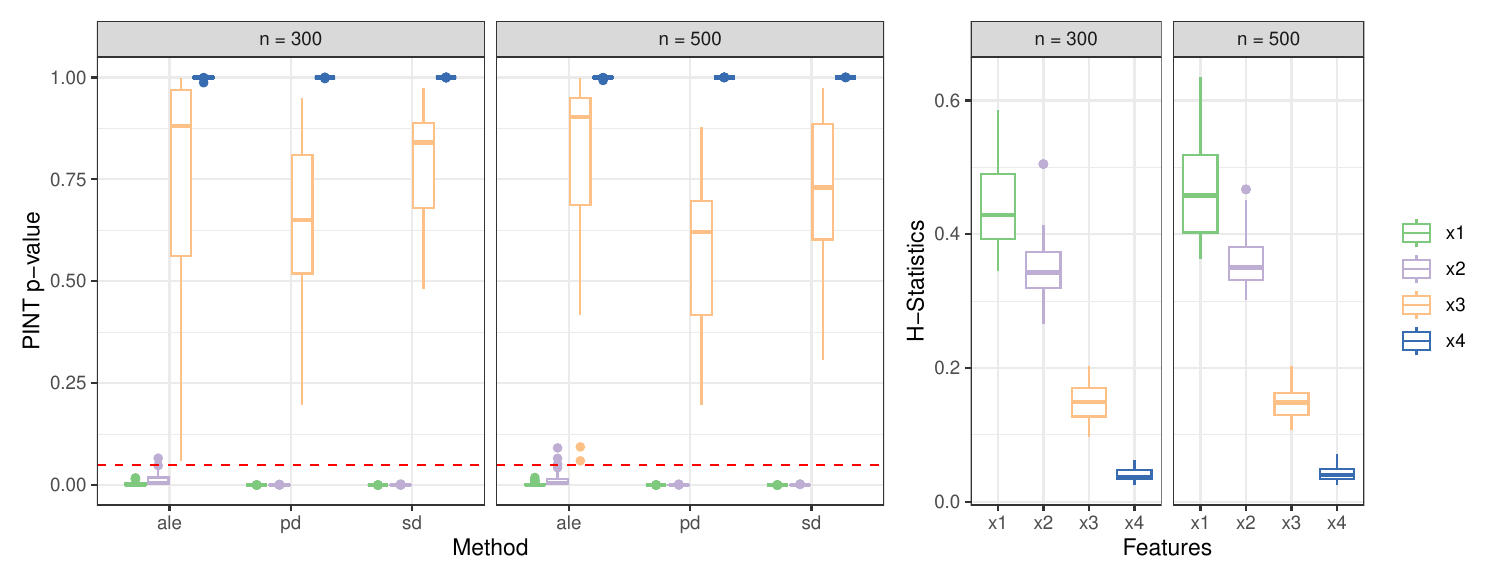}
    \caption{Boxplots showing the distribution of p-values of each feature for different sample sizes and effect methods over all repetitions when PINT is applied (left) and the distribution of feature-wise H-Statistic values for both sample sizes (right).}
    \label{fig:sim_pint_hstat}
\end{figure}

\paragraph{\normalfont\textit{Results.}}
The two left plots in Figure \ref{fig:sim_pint_hstat} show that $\xv_3$ and $\xv_4$ are always correctly identified as insignificant (according to the chosen $\alpha$ level), while the interacting features $\xv_1$ and $\xv_2$ are always significant and considered in $S$ (apart from a few exceptions for ALE). The sample size does not seem to have a clear influence on PINT in this setting. The right plots in Figure \ref{fig:sim_pint_hstat} show that even the H-Statistic values of the non-influential and uncorrelated feature $\xv_4$ are larger than $0$ for both sample sizes. The H-Statistic values of $\xv_3$ could support its inclusion in $S$, depending on the chosen threshold. Since this choice is not clear for the H-Statistic, it is not very suitable as a pre-selection method for GADGET.

\begin{figure}[tbh]
    \centering
    \includegraphics[width=0.8\textwidth]{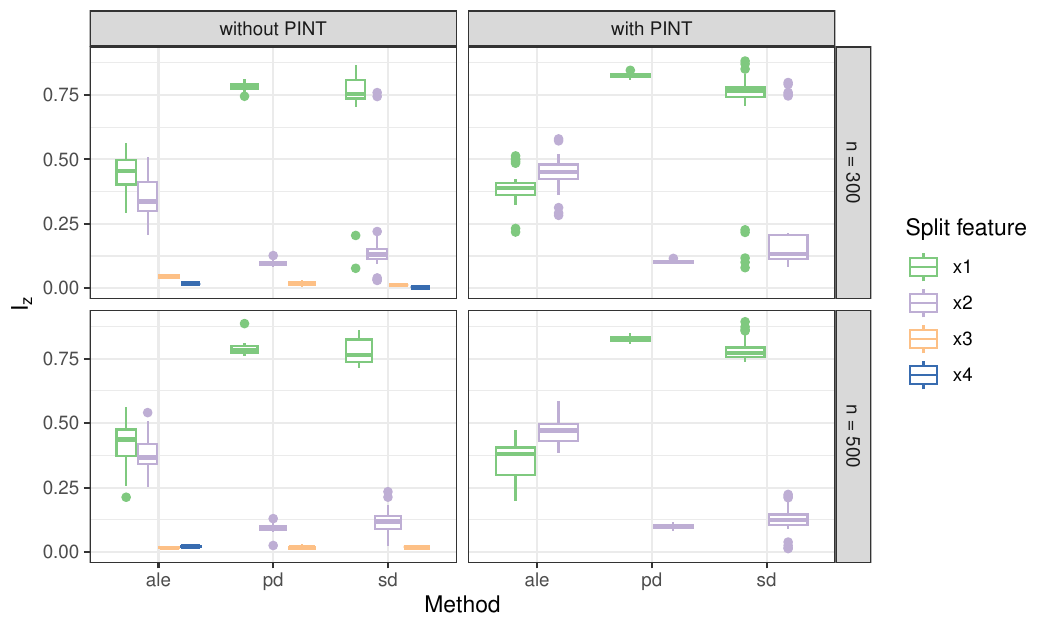}
    \caption{Boxplots showing the interaction-related heterogeneity reduction $I_z$ per split feature over $30$ repetitions when PD, ALE, or SD is used within GADGET. The rows show the results for the two different sample sizes, and the columns indicate if GADGET is used based on all features without using PINT upfront (left) or GADGET is used based on the feature subset resulting from PINT (right).}
    \label{fig:sim_pint}
\end{figure}

Figure \ref{fig:sim_pint} illustrates that we correctly only consider $\xv_1$ and $\xv_2$ when PINT is applied upfront, while GADGET also splits w.r.t. $\xv_3$ for all settings and (in some cases) even w.r.t. $\xv_4$ if PINT is not applied upfront. The influence of these two features appears to be stronger for smaller sample sizes, as indicated by $I_z$. Among the various effect methods used in GADGET, PD and SD attribute most of the reduction in heterogeneity to $\xv_1$ and a smaller part to $\xv_2$, while ALE attributes the heterogeneity more equally between the two split features. This discrepancy might be explained by the correlation between $\xv_2$ and $\xv_3$, which particularly affects the two methods based on marginal distributions (that is, PD and SD).
Furthermore, Table \ref{tab:pint} shows that we tend to obtain shallower trees when we use PINT upfront, while we retain the level of heterogeneity reduction by obtaining similar $R^2_j$ values for the two interacting features $\xv_1$ and $\xv_2$.

Thus, PINT reduces the number of features to consider in the interacting feature subset for GADGET depending on the regarded feature effect method. This leads to better results in GADGET in the sense that we only split w.r.t. truly interacting features defined by PINT and receive shallower (and thus more interpretable) trees.


\begin{table}[tbh]
\centering
\begin{tabular}{|rll|rr|rrr|}
\hline
& & & \multicolumn{2}{c}{$R^2_j$} & \multicolumn{3}{|c|}{Number of Subspaces}\\
  \hline
n & Method & PINT & $R^2_1$ mean(sd)  & $R^2_2$ mean(sd) & Min & Max & Median\\ 
  \hline
300 & ALE & no & 0.83 (0.05) & 0.83 (0.05) &   4 &  10 & 7.50\\ 
  300 & ALE & yes & 0.83 (0.07) & 0.82 (0.05)  &   1 &  10 & 7.00\\ 
  300 & PD & no & 0.94 (0.02) & 0.84 (0.06) &   2 &   7 & 4.00\\ 
  300 & PD & yes & 0.92 (0.03) & 0.80 (0.08) &   2 &   5 & 3.00 \\ 
  300 & SD & no & 0.93 (0.04) & 0.90 (0.05) &   2 &  11 & 5.00 \\ 
  300 & SD & yes & 0.93 (0.04) & 0.90 (0.05) &   2 &  11 & 4.50 \\ 
  500 & ALE & no & 0.83 (0.04) & 0.83 (0.06)&   4 &  10 & 7.00  \\ 
  500 & ALE & yes & 0.86 (0.04) & 0.82 (0.05) &   1 &  11 & 6.50 \\ 
  500 & PD & no & 0.94 (0.03) & 0.84 (0.06) &   2 &   7 & 4.00\\ 
  500 & PD & yes & 0.93 (0.03) & 0.82 (0.08) &   2 &   5 & 4.00\\ 
  500 & SD & no & 0.93 (0.04) & 0.90 (0.05) &   2 &  11 & 5.00\\ 
  500 & SD & yes & 0.93 (0.04) & 0.90 (0.05) &   2 &   9 & 5.00\\ 
   \hline
\end{tabular}
\caption{Interaction-related heterogeneity reduction per feature for $\xv_1$ and $\xv_2$ by mean (standard deviation) of $R^2_j$ and minimum, maximum and median number of final subspaces after applying GADGET based on different sample sizes, effect methods and with and without using PINT upfront.}
\label{tab:pint}
\end{table}

\section{Empirical Validation of the PINT Procedure}
\label{sec:sim_pint_valid}

We now empirically validate PINT to detect interactions.

\paragraph{\normalfont\textit{Hypothesis.}} It can be shown empirically that the PINT procedure is faithful w.r.t. the type I error and the power of the test as already validated theoretically in Section \ref{sec:interact_pimp}. 

\paragraph{\normalfont\textit{Experimental Setting.}}
Let $X_1$, 
$X_2$, $X_3$, $X_4$ $\sim \mathbbm{U}(-1,1)$ be four independently distributed features, from which we draw $500$ observations. The true functional relationship is given by $\mathbf y = \beta \xv_1 \xv_2 + \epsilon, \;\epsilon \sim \mathbbm{N}(0,1)$ with $\beta \in \{0,0.25,0.5,0.75,1,1.25,1.5,1.75,2,2.25,2.5,2.75,3\}$. Hence, for $\beta = 0$, no feature has an influence on $\mathbf{y}$, which is then defined only by the noise term: $\mathbf{y} = \epsilon$. For each setting, we train an SVM based on an RBF kernel with cost parameter $C=1$ and choose the inverse kernel width based on the data. We apply PINT for all three feature effect methods and with $s = 100$ and $\alpha = 0.05$. We perform the experiments $1000$ times to calculate the rejection rate for each feature and setting. 

\begin{figure}[ht]
   \centering    \includegraphics[width=0.85\textwidth]{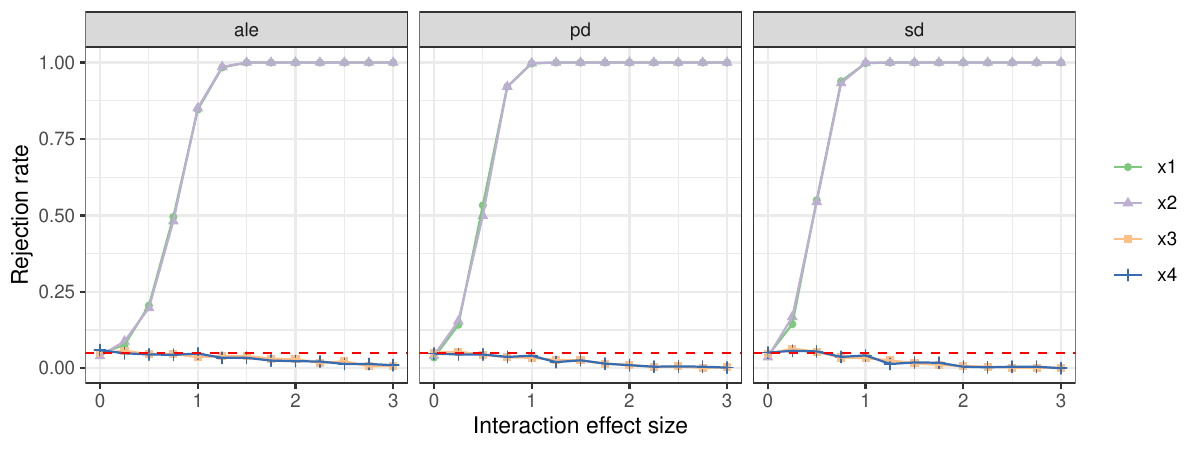}
   \caption{Rejection rates of PINT for $1000$ repetitions for PD, ALE, and SD and for varying interaction effect sizes $\beta$. For $\beta = 0$, rejection rates correspond to type I errors. For $\beta > 0$, rejection rates of $\xv_1$ and $\xv_2$ correspond to the power of the test, those of $\xv_3$ and $\xv_4$ to type I errors. The red dashed line represents the significance level of $\alpha =0.05$. Where not visible, the green line ($\xv_1$) is below the purple line ($\xv_2$).}
    \label{fig:pint_eval_error}
\end{figure}

\begin{figure}[ht]
   \centering    \includegraphics[width=0.9\textwidth]{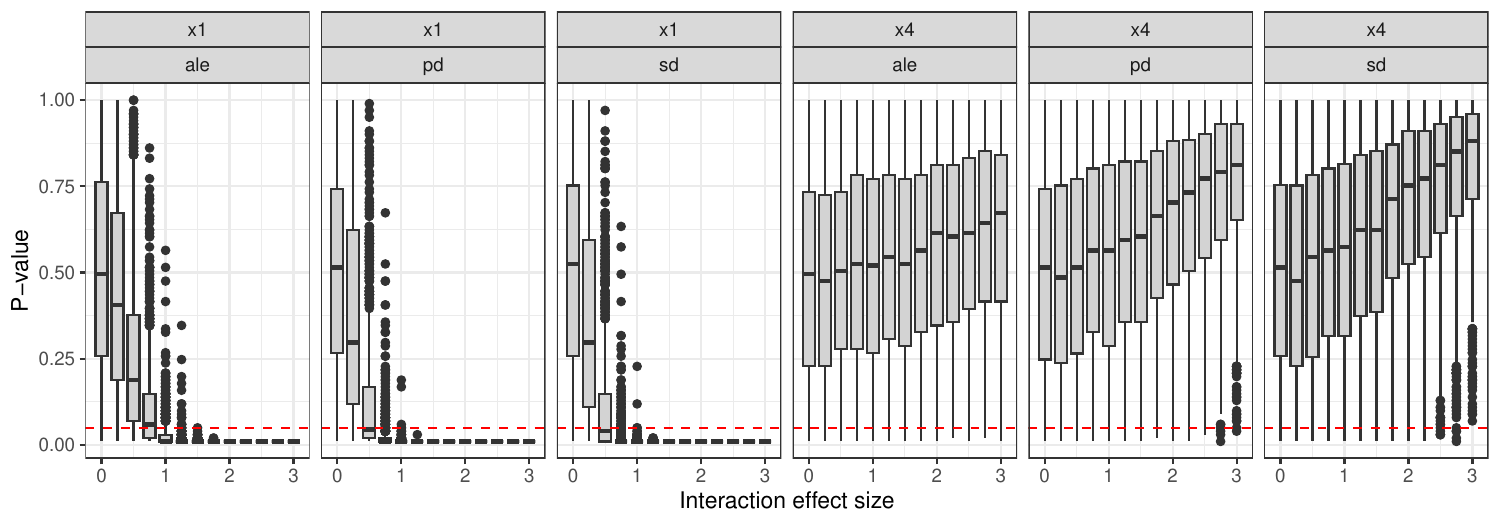}
   \caption{P-value distributions of PINT for $1000$ repetitions for $\xv_1$ and $\xv_4$ for all three feature effect methods and for varying interaction effect sizes $\beta$. The red dashed line represents the significance level of $\alpha =0.05$.}
    \label{fig:pint_eval_pval}
\end{figure}

\paragraph{\normalfont\textit{Results.}}
The rejection rate of all four features and all effect methods for $\beta = 0$ align closely with the significance level $\alpha = 0.05$ (see Figure \ref{fig:pint_eval_error}). While the power of the test approaches $1$ for increasing effect sizes of the interaction between $\xv_1$ and $\xv_2$ for all three methods, the type I error represented by the rejection rates of $\xv_3$ and $\xv_4$ goes to $0$. This trend can be explained as follows: The higher $\beta$, the higher the influence of $\xv_1$ and $\xv_2$ and thus the more the model focuses on these two features and ignores the two non-influential features. This can also be seen on the p-value distributions, which, for non-influential features such as $\xv_4$, becomes more skewed towards $1$ the higher the interaction effect, as illustrated in Figure \ref{fig:pint_eval_pval}. Thus, in the most extreme case, the model disregards the non-influential features entirely, resulting in their local effects' heterogeneity being exactly $0$. However, when permuting $\mathbf{y}$ and refitting the model, random effects may be learned for $\xv_3$ and $\xv_4$. Hence, the calculated values under the null distribution might be higher than the actual risk value, causing the observed artifact in Figure \ref{fig:pint_eval_error} where the type I error is 0 instead of showing a false rejection rate of $5\%$.

More diverse and complex settings that confirm these observations and empirically validate their correctness are discussed in Appendix \ref{app:further_examples_pint}. 

\section{Real-World Applications}
\label{sec:real_world}
We now show the usefulness of our introduced methodology on real-world applications, revealing insights into the learned regional effects, interactions, and potential biases in the data or model. We present GADGET's applicability in higher-dimensional settings and offer guidelines for handling such scenarios in Section \ref{sec:high_dim}.

\subsection{COMPAS Data Set}

Due to potential subjective judgement and the resulting bias in the decision making process of the criminal justice system \citep{blair2004influence}, ML models have been used to predict recidivism of defendants to provide a more objective guidance for judges. However, if the underlying training data are biased (e.g., different socioeconomic groups have been treated differently for the same crime in the past), the ML model might learn the underlying data bias, and due to its black-box nature, explanations for its decision-making process and a potential recourse are harder to achieve \citep{fisher2019all}.

We want to use GADGET here to obtain more information on how different characteristics of the defendant and their criminal record influence the risk of recidivism within different subgroups and if ``inadmissible'' characteristics such as ethnicity or gender cause a different risk evaluation.
For our analysis, we use the COMPAS data set to predict the risk of violent recidivism collected by ProPublica \citep{larson2016propublica}.
As \cite{fisher2019all}, we choose the three admissible features (1) age of the defendant, (2) number of prior crimes, and (3) if the crime is considered a felony versus misdemeanor, and the two ``inadmissible'' features (1) ethnicity and (2) gender of the defendant. We use the subset of African-American and Caucasian defendants and apply the data preprocessing steps suggested by ProPublica and applied in \cite{fisher2019all}, leaving us with 3373 defendants of the original pool of 4743 defendants. We consider a binary target variable that indicates a high ($=1$) or low ($=0$) recidivism risk, based on a categorization of ProPublica.
We perform our analysis on the entire data set, using a tuned SVM.\footnote{We performed proper model comparison as well to ensure that this is a well-fitting model, at least in comparison to other standard choices.}

Since the features do not show high correlations, we use ICE and PD for our analysis.\footnote{We obtained similar results by using SD instead of PD within GADGET, see Appendix \ref{app:gadget_sd_compas}.}
Figure \ref{fig:compas_root} shows that the average predicted risk visibly decreases with age, while the predicted risk first increases steeply with the number of previous crimes up to 10 and then decreases slightly for higher values. The PD values for the type of crime do not differ substantially. When considering the two ``inadmissible'' features, there is on average a slightly higher risk of recidivism for African-American versus Caucasian and female versus male defendants. For all features, we can observe highly differing local effects. Particularly, the heterogeneous ICE curves for age and number of prior crimes indicate feature interactions.   

\begin{figure}[tbh]
    \centering
    \includegraphics[width=\textwidth]{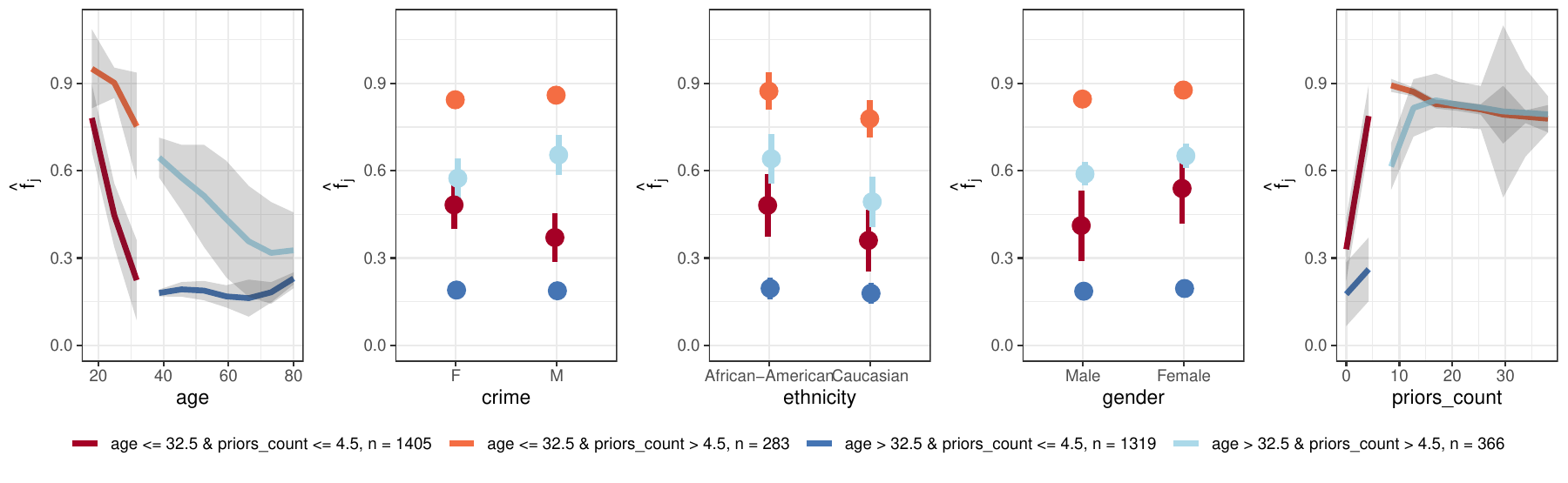}
    \caption{Regional PD plots for features of the COMPAS application after applying GADGET. Grey areas of numeric features and error bars of categorical features indicate the $95\%$ interaction-related heterogeneity interval (see Appendix \ref{app:overview_estimates}).}
    \label{fig:compas_region}
\end{figure}

We first apply PINT with PD to define our subset $S$ for GADGET. Using $s = 200$ and $\alpha = 0.05$, we approximate the null distribution using the procedure described in Section \ref{sec:interact_pimp}. All five features are significant w.r.t. the chosen significance level, so we choose $S = Z = \{age, crime, ethnicity, gender, priors\_count\}$ for GADGET.
We apply GADGET with a maximum depth of $3$ and $\gamma = 0.15$. 
\begin{figure}[tbh]
    \centering
    \includegraphics[width=\textwidth]{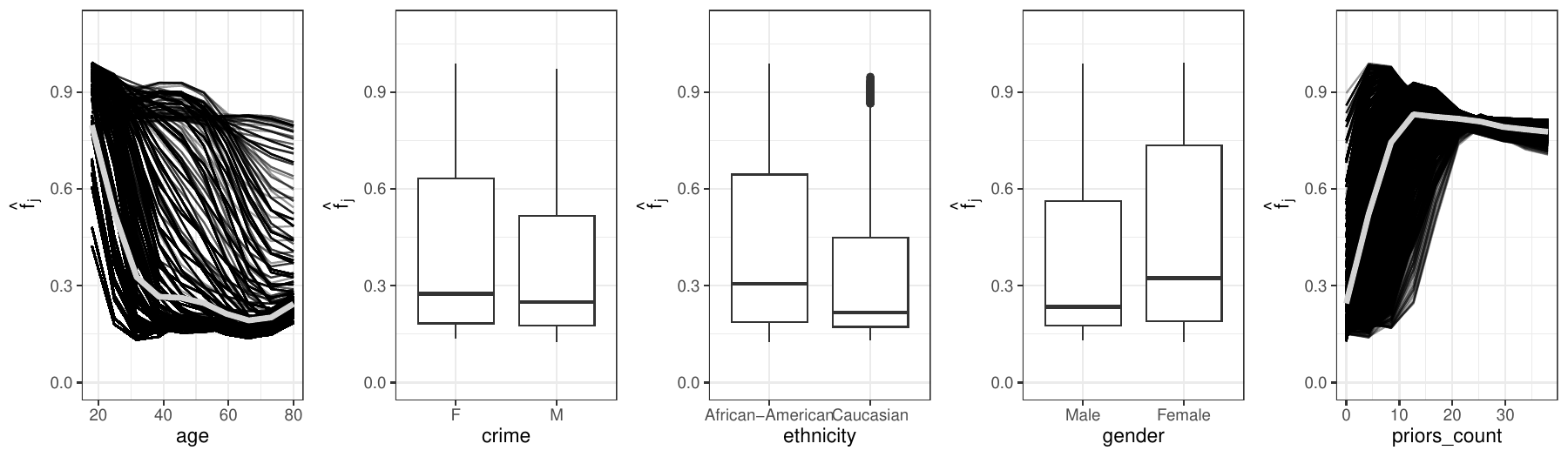}
    \caption{ICE and PD curves of considered features of the COMPAS application example.}
    \label{fig:compas_root}
\end{figure}
%
The effect plots for the four resulting regions are shown in Figure \ref{fig:compas_region}. GADGET performed the first split according to age and the splits on the second depth according to the number of prior crimes. The total reduction in interaction-related heterogeneity is $R^2_{Tot} = 0.86$. The highest reduction in heterogeneity is given by age and the number of prior crimes, which interact the most (see Figure \ref{fig:compas_region}). For defendants around 20 years of age, the predicted risk tends to be high. However, for defendants with a small number of prior crimes, the predicted risk decreases very quickly with increasing age, reaching a low risk and remaining so for people older than 32 years. In contrast, for defendants with more than four prior crimes, the predicted risk decreases only slowly with increasing age. The regional feature effects of the number of prior crimes show that the interaction-related heterogeneity is small for the subgroup of younger people, while some heterogeneity still remains for defendants older than 32 years of age. Furthermore, the interaction-related heterogeneity of the three binary features (ethnicity, gender, and severity of crime) was reduced, indicating an interaction between each of them and age as well as the number of prior crimes. Although the effect on the predicted risk only differs slightly between the categories of the three binary features for older defendants with a lower number of prior crimes and for younger defendants with a high number of prior crimes, greater differences were observed for the other two subgroups. 
The overall difference in predicted risk for the two inadmissible features of ethnicity and gender seems to be especially high for people over 32 years with a higher number of prior crimes, as well as for people younger than 32 with a lower number of prior crimes.
These differences in the predicted risk between subgroups of defendants might be due to a potentially learned bias regarding the ethnicity and gender of the defendant and thus potentially resulting in a more severe predicted risk of recidivism for some defendants than for others. 

Note that we applied GADGET to an ML model that is fitted on the COMPAS scores and not directly on COMPAS. Consequently, we are unable to draw conclusions about the learned effects of the underlying commercial black-box model. However, GADGET is model-agnostic and can be applied to any accessible black-box model. 

\subsection{Bikesharing Data Set}
We use the bikesharing data set \citep{james2022ISLR2}, to predict the hourly counts of rented bikes from the Capital bikeshare system for the years 2011 and 2012. 
We fit an RF with 10 seasonal and weather features:
the day and hour the bike was rented, if the current day is a working day, the season, number of casual bikers, normalized temperature, perceived temperature, wind speed, humidity, and weather situation (categorical, e.g., ``clear'').

\begin{figure}[tbh]
    \centering
    \includegraphics[width=0.9\textwidth]{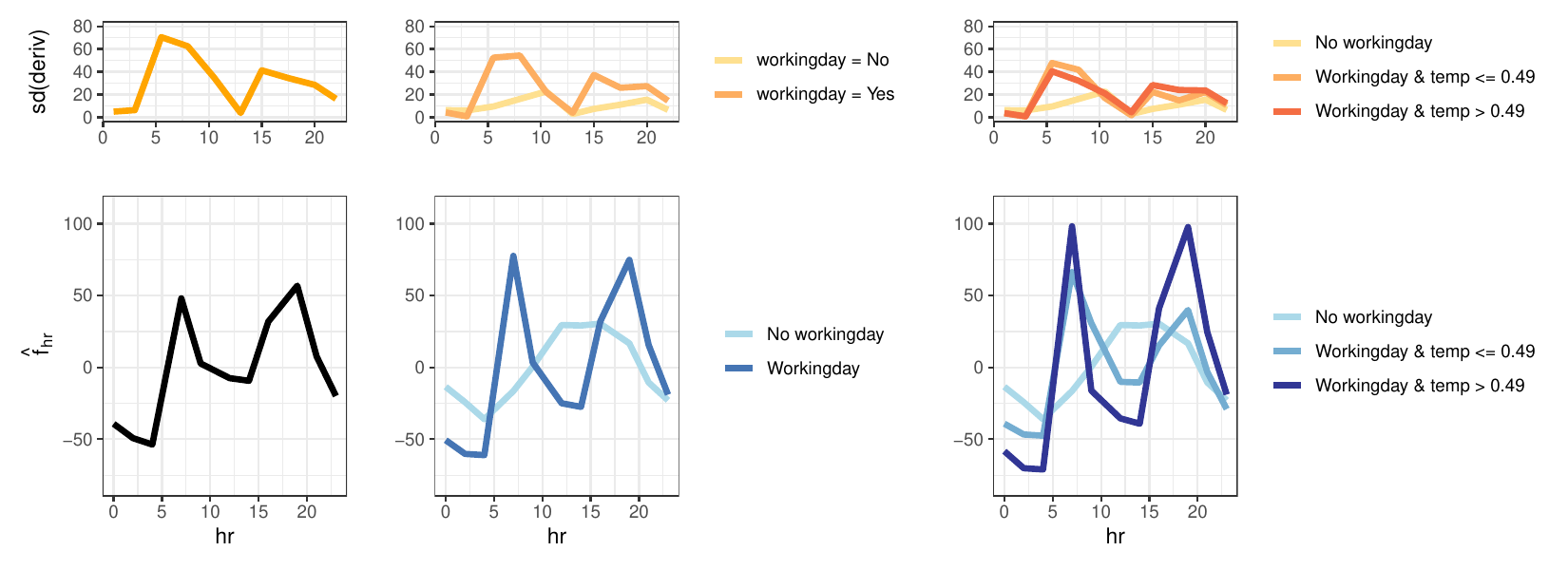}
    \caption{Global (left) and regional ALE plots after applying GADGET for feature hour using $S=Z = \{hr, workingday\}$ (middle) and using $S=Z = \{hr, workingday, temp\}$ (right) of the bikesharing data. The upper plots show the interaction-related heterogeneity as defined in Appendix \ref{app:overview_estimates}.}
    \label{fig:bike}
\end{figure}

Again, we first apply PINT with $s=200$ and $\alpha = 0.05$ on all features to define the interacting feature subset $S$ for GADGET. Since some features---such as season, temperature, and perceived temperature---are correlated, we use ALE for our analysis. Although the features ``hour'' and ``working day'' are highly significant, the p-values of all other features are close to $1$, indicating that only the heterogeneity of the local effects for hour and workingday is caused by interactions.
Hence, we set $S=Z = \{hr, workingday\}$ and apply GADGET with a maximum depth of 3 and $\gamma = 0.15$. GADGET splits once with the binary feature ``workingday'', reducing the interaction-related heterogeneity of the two features by $R^2_{Tot} = 0.88$. The middle plots of Figure \ref{fig:bike} show the regional ALE plots after applying GADGET. During working days and rush hours, prominent peaks are observed, while there is a decline during the noon and afternoon hours. However, on non-working days, the trend is the opposite. This interaction is not visible in the global ALE plot of the feature hour (left plot). 
The interaction-related heterogeneity of the feature ``hour'' is reduced compared to the global plot, although there is still some apparent variation for working days. From a domain perspective, we might also consider an interaction of the temperature with the hour and working day \citep[as done in][]{hiabu2022unifying}. Thus, we include the temperature in the feature subsets $S$ and $Z$ and apply GADGET again with the same settings as described above. The feature ``workingday'' governs the first split. However, for the region of working days, GADGET splits again according to temperature, as shown in the right plot of Figure \ref{fig:bike}. Although the interaction-related heterogeneity of the feature ``temperature'' was barely reduced within GADGET ($R^2_j = 0.03$)---which supports the results of PINT---using temperature in the subset of splitting features $Z$ further reduced the interaction-related heterogeneity of hour by $15\%$. Thus, feature interactions can be asymmetric, and extending $Z$ based on domain knowledge might be a valid option in some cases. 

\section{High-Dimensional Real-World Application with Filtering Procedure}
\label{sec:high_dim}
We have considered only low-dimensional settings with three to ten features so far. Real-world data sets are often of a higher dimension,
posing a challenge for many interpretation methods in general either due to high runtime or potentially overwhelming, incomprehensible outputs \citep{molnar2022pitfalls}. 
Here, we illustrate how this problem can be mitigated for GADGET via a two-step filtering process.



If we can quickly reduce the original feature set to the feature subset which is indeed used by the model and which is interacting, GADGET's computational complexity then only scales with the number of learned interactions, which can be much smaller than considering all potential interactions of the original feature space.
We illustrate this argumentation by the spam data set which contains $57$ numeric features to classify $4601$ e-mail samples as spam or non-spam \citep{misc_spambase_94}, based on simple text characteristics.
We train an RF with $500$ trees, achieving a misclassification error of $4.8\%$ measured by a $5$-fold cross-validation, indicating good performance \citep{misc_spambase_94}.
We apply GADGET-PD since there are only very few correlations present in the feature space. 

We suggest a two-step pre-filtering procedure for higher dimensional settings: First, filtering features based on their overall heterogeneity of local effects via Eq.~\eqref{eq:risk}. Second, filtering of the remaining features after step $1$ for interacting features with PINT according to Algorithm~\ref{alg:pint}.
For the first filtering step, we sort the features according to the heterogeneity of local effects measured by Eq.~\eqref{eq:risk} with $h$ being the mean-centered ICE values and $\mathcal{A}_g$ being the entire feature space $\mathcal{X}$. If the heterogeneity is very small, the shapes of the curves are very similar. Hence, the feature does not interact with other features in the data set and thus does not influence the prediction at all or influences the prediction only via a main effect. Therefore, we can disregard this feature for GADGET.\footnote{In the case of asymmetric feature interactions, we might consider such features in $Z$ but not in $S$, however, this might require domain knowledge (see the bikesharing example in Section \ref{sec:real_world}).}

\begin{figure}[tbh]
    \centering
    \includegraphics[width=0.8\textwidth]{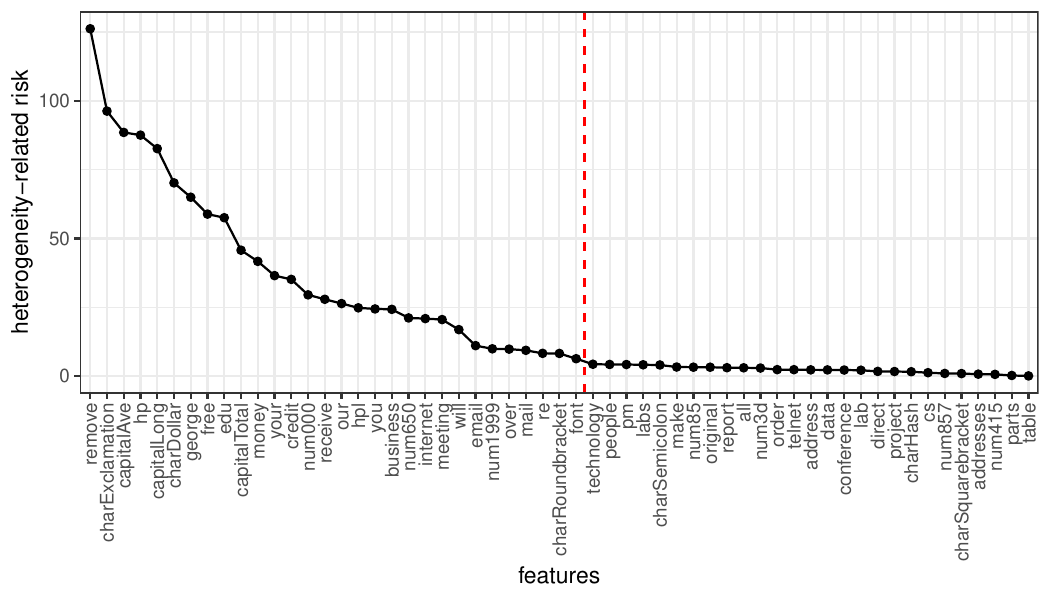}
    \caption{Risk values according to Eq.~\eqref{eq:risk}. Red line marks selected cutoff.}
    \label{fig:spam_elbow}
\end{figure}

\begin{figure}[tbh]
    \centering
    \includegraphics[width=0.8\textwidth]{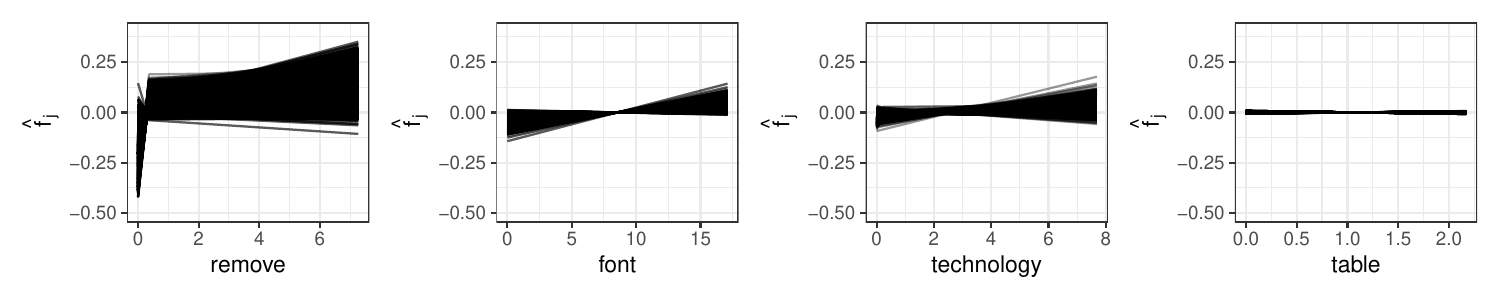}
    \caption{Mean-centered ICE curves for the feature with the highest (left) and lowest (right) risk values and the two features closest to the cutoff point (middle).}
    \label{fig:spam_heterogeneity}
\end{figure}

To decide on the cutoff point of the risk values to group features into one of the two groups: heterogeneous or homogeneous local effects, we 
visualize the risk values calculated by Eq.~\eqref{eq:risk} in decreasing order by an elbow graph as illustrated in Figure \ref{fig:spam_elbow}. The cutoff point is chosen rather conservatively as the last substantial drop in risk before all remaining values are close to $0$.\footnote{This cutoff point can also be chosen in a data-driven way, e.g., by the percentage of explained variance similarly to principal component analysis.} Figure \ref{fig:spam_heterogeneity} visualizes the respective mean-centered ICE curves for the two features closest to the cutoff point and compares them to the mean-centered ICE curves of the features with the highest and smallest risk values according to Figure \ref{fig:spam_elbow}. 
This first filtering step leaves $30$ features in the group of heterogeneous local effects.

We then apply PINT as a second filtering step, with PD, $s = 200$ permutations, and $\alpha = 0.05$ as described in Section \ref{sec:interact_pimp}. $12$ out of the $30$ features are detected as significant. 
We apply GADGET-PD with a relative improvement of $\gamma = 0.25$, a maximum tree depth of $5$ and minimum number of observations per node of $100$ as early stopping criteria.\footnote{See Appendix \ref{app:recom_stop_crit} for general recommendations on specifying these hyperparameters.} 
We define $S = Z$ as the $12$ obtained features. 
We obtain $5$ final regions when GADGET-PD is applied. With $R^2_{Tot} = 0.67$, we removed $67\%$ of the interaction-related heterogeneity within the final regions of all $12$ features in $S$. The visualizations and interpretations of the regional effect plots are provided in Appendix \ref{app:higher_dim}.

Hence, by using the two filtering steps described above, we reduced the number of plots to examine from $57$ to $12$. This increases comprehensibility and aids the user to focus on the relevant features and learned relationships.

\section{Conclusion and Discussion}
We introduced GADGET, a new framework that partitions the feature space into interpretable and distinct regions such that feature interactions of any subset of features are minimized.
GADGET can be used with any feature effect method that satisfies the \textit{local decomposability} axiom (Axiom \ref{axiom:feat_rel}). We demonstrated its use with well-known global feature effect methods---namely PD, ALE, and SD---and provided visualizations and estimates for regional feature effects and interaction-related heterogeneity. Furthermore, we introduced different measures to quantify and analyze feature interactions through the reduction of interaction-related heterogeneity within GADGET. 
To define the interacting feature subset, we introduced the novel PINT procedure to detect global feature interactions for any feature effect method used within GADGET. 

Our experiments showed that PINT detects the true subset of interacting features and that preselection leads to more meaningful and interpretable results in GADGET.
Additionally, feature effect methods based on conditional distributions, such as ALE, tend to produce more stable results compared to those based on marginal distributions, especially when features are highly correlated.  
Furthermore, due to a different weighting scheme in the decomposition of Shapely values compared to the other considered feature effect methods, higher-order terms are less likely to be detected by GADGET, especially if Shapley values are not recalculated after each partitioning step. 
This approach might be computationally expensive, which can be seen as a possible limitation. However, recent research has focused on fast approximation techniques of Shapley values \citep[e.g.,][]{lundberg_2017_unified, covert2021improving, jethani2021fastshap, lundberg2020local, chau2022rkhs, kolpaczki2023approximating} and may offer solutions to overcome this limitation. 

In general, our proposed method works well if learned feature interactions are not overly local and if observations can be grouped based on feature interactions such that local feature effects within the groups are homogeneous, while different groups exhibit heterogeneous feature effects. This approach helps to avoid the aggregation bias of global feature effect methods, provides deeper insights into learned regional effects, and detects potential biases within subgroups (as illustrated in Section \ref{sec:real_world}). 
One of the real-world examples also showed that the frequently made assumption of symmetric feature interactions (e.g., in Shapley values) is not always the case \citep[see also][for research on asymmetrical feature interactions]{masoomi_explanations_2022}. Thus, including domain knowledge to define the interacting feature subset $Z$ might sometimes be meaningful. 

Finally, GADGET is---as all other interpretation methods---affected by the Rashomon effect due to differently learned feature effects for different fitted models. Thus, the results of GADGET when applied to different models might not be the same, but might provide more insights when compared to each other. Analyzing the influence of the Rashomon effect is out-of-scope in this paper but will be considered in future work. More details and related work to this topic are provided in Appendix \ref{app:rashomon}.


\acks{This work has been partially supported by the Bavarian Ministry of Economic Affairs, Regional Development, and Energy as part of the program ``Bayerischen Verbundförderprogramms (BayVFP) – Förderlinie Digitalisierung – Förderbereich Informations- und Kommunikationstechnik'' under the grant DIK-2106-0007 // DIK0260/02 and by the German Research Foundation (DFG) under the grants 437611051 and 459360854. The authors of this work take full responsibility for its content.}


\newpage

\appendix

\section{Further Details on Functional ANOVA Decomposition} %
\label{app:fanova}

As stated in Section \ref{sec:background_fanova}, the functional ANOVA provides a decomposition of the prediction function into the main and higher-order effect of all involved features \citep{sobol1993sensitivity,stone1994fanova, hooker_discovering_2004, hooker_generalized_2007, li_general_2012}:
$$
\hat{f}(\xv) = 
 g_0 + \sum_{j = 1}^p g_j(\xv_j) + \sum_{j \neq k} g_{jk}(\xv_j,\xv_k) + \ldots + g_{12\ldots p}(\xv)
=g_0 +\sum_{k = 1}^p \; \sum_{\substack{W \subseteq \{1,\ldots,p\},\\ |W| = k}} g_W(\xv_W),
$$
where $g_0$ is a constant,
$g_j(\xv_j)$ denotes the main effect of the $j$-th feature, $g_{jk}(\xv_j,\xv_k)$ is the pure two-way interaction effect between features $\xv_j$ and $\xv_k$, and so forth. The last component $g_{12\ldots p}(\xv)$ always allows for an exact decomposition.
We discuss here the standard and the generalized functional ANOVA decomposition which provide a unique decomposition of the above-stated equation based on different assumptions.

\paragraph{\normalfont\textit{Standard Functional ANOVA Decomposition.}}

The standard functional ANOVA decomposition \citep{sobol1993sensitivity,hooker_discovering_2004} provides a unique solution in case of independent features by imposing the vanishing (or strong annihilating) condition \citep{li_general_2012, rahman_generalized_2014}: $\textstyle \int g_W(\xv_W) d\P_{X_j}(X_j) = 0, \; \forall W \neq \emptyset \text{ and } \forall j \in W$.
The vanishing condition implies a) that the component functions $g_W(\xv_W)$ ``integrate to zero w.r.t. the marginal density of each random variable'' in $W$; b) the \textit{zero means} property, i.e., $E[g_W(X_W)] = 0$; c) the \textit{orthogonality} property, i.e., $E[g_W(X_W)g_V(X_V)] = 0$ with $\emptyset \neq W \subseteq \{1,\ldots, p\}$, $\emptyset \neq V \subseteq \{1,\ldots, p\}$ and $W \neq V$; d) that each (lower-dimensional) component function of $g_W$ itself is the zero function. 
The component functions can be determined recursively by
\begin{equation}
    g_W(\xv_W) = \int_{\xv_{-W}} \left(\fh(\xv) - \sum_{V \subset W} g_V(\xv_V)\right) d\P_{X_{-W}}(\xv_{-W}).
      \label{eq:fANOVA_seq}
 \end{equation}

\paragraph{\normalfont\textit{Generalized Functional ANOVA Decomposition.}}
In the case of dependent features, the \textit{vanishing condition} does not work, as it does not imply all a)-d) above. 
Therefore, \cite{hooker_generalized_2007} introduced the generalized functional ANOVA decomposition with a \textit{relaxed vanishing} (or weak annihilating) condition: $\textstyle \int g_W(\xv_W) p_{X}(\xv) d\xv_j d \xv_{-W}= 0, \; \forall j \in W \neq \emptyset$ where $p_{X}$ is the joint density of $X$.
If this condition holds, then the hierarchical orthogonality condition $\int g_W(\xv_W)h_V(\xv_V)d\P_X(X) = 0, \; \forall V \subset W$ is satisfied. This again leads to a unique decomposition. 

With $p_X(\xv)$ now being a general probability density with its support being grid-closed and with $\fh$ being square-integrable, 
the component functions can then be uniquely determined by optimizing Eq.~\eqref{eq:fANOVA_opt}.
\begin{equation}
    g_W(\xv_W) = \argmin_{h_W \in \mathbb{L}^2(\mathbb{R}^W),W \subseteq \{1,\ldots,p\}} \int \left(\sum_{W \subseteq \{1,\ldots,p\}} h_W(\xv_W) - \hat f(\xv)\right)^2 d\P_X(\xv)
    \label{eq:fANOVA_opt}
\end{equation}

The estimation procedure can become computationally expensive with increasing dimensionality.
However, being able to decompose a prediction function into its components and hence, being able to interpret main and higher-order effects between features separately is often desired. Thus, models that are able to directly model these structures have been introduced and extended. Generalized additive models (GAMs), for example, additively decompose the prediction function into the (non-linear) main effects of each feature. 
\cite{lou_accurate_2013} extended GAMs by adding relevant two-way feature interactions to the main effect model. 
\cite{watanabe_constrained_2021} further extend this approach by also allowing feature interactions of a higher order as well as monotonic constraints. 

Recent research has focused on computing the decomposition more efficiently, e.g., by developing models that integrate the respective conditions directly in the optimization process through constraints \citep[e.g.,][]{sun2022puregam} or on purifying the resulting decomposition for specific models within the modeling process \citep[e.g.,][]{lengerich2020purifying, hu_using_2022}. 
However, most approaches require estimating the underlying data distribution which remains challenging \citep{lengerich2020purifying}, with initial attempts employing adaptive kernel methods \citep{sun2022puregam}.
Notably, finding an appropriate decomposition is only complex in the presence of feature interactions, otherwise, the prediction function can be uniquely decomposed into the main effects using models like the generalized additive model (GAM).

\section{Theoretical Evidence of GADGET}
\label{app:proofs}
We provide the proofs for the theorems of Sections \ref{sec:requirements} to \ref{sec:shap}. This includes the theoretical foundation of GADGET as well as the proofs of the applicability of the feature effect methods PD, ALE, and SD within the GADGET algorithm by defining respective local feature effect functions that fulfill Axiom \ref{axiom:feat_rel}. We also show in Appendix \ref{app:repid} that the REPID method which was introduced in Section \ref{sec:rel_work} is a special case of the GADGET algorithm.

\subsection{Proof of Theorem \ref{theorem:interact_rel}}
\label{app:proof_theorem2}
\textit{Proof Sketch}
If the function $\hat f(\xv)$ can be decomposed as in Eq.~\eqref{eq:fANOVA} and if Axiom \ref{axiom:feat_rel} holds for the local feature effect function $h$, then the main effect of feature $\xv_j$ at $x_j$ is cancelled out within the loss function in Eq. \eqref{eq:loss}. Thus, the loss function measures the interaction-related heterogeneity of feature $\xv_j$ at $x_j$, since the variability of local effects in the subspace $\mathcal{A}_g$ are only based on feature interactions between the $j$-th feature and features in $-j$.

\medskip

\begin{proof}
\begin{footnotesize}
    \begin{eqnarray*}
       \mathcal{L}_j\left(\mathcal{A}_g, x_j\right) &=&
    \sum\limits_{i: \xi \in \mathcal{A}_g}
     \left(h(x_j, \mathbf{x}_{-j}^{(i)}) - \E[h (x_j, X_{-j})|\mathcal{A}_g] \right)^2\\
     &=&   \sum\limits_{i: \xi \in \mathcal{A}_g}
     \left(g_j(x_j) + \sum\limits_{k = 1}^{p-1} \; \sum\limits_{\substack{W \subseteq -j,\\ |W| = k}} g_{W \cup j}(x_j,\xv_W^{(i)}) - \E\biggl[g_j(x_j) + \sum\limits_{k = 1}^{p-1} \; \sum\limits_{\substack{W \subseteq -j,\\ |W| = k}} g_{W \cup j}(x_j,X_W) \mid \mathcal{A}_g\biggr]\right)^2\\
     &=&  \sum\limits_{i: \xi \in \mathcal{A}_g}
     \left(\sum\limits_{k = 1}^{p-1} \; \sum\limits_{\substack{W \subseteq -j,\\ |W| = k}} g_{W \cup j}(x_j,\xv_W^{(i)}) - \E[ g_{W \cup j}(x_j,X_W)|\mathcal{A}_g]\right)^2 
   \end{eqnarray*}
\end{footnotesize}
\end{proof}

\subsection{Proof of Theorem \ref{corollary:objective}}
\label{app:proof_cor3}
\textit{Proof Sketch}
We show that the theoretical minimum of the objective is $\mathcal{I}(t^*,z^*) = 0$ if no feature in $S$ interacts with any feature in $-Z$. We apply the functional decomposition within the risk function $\mathcal{R}_j(\mathcal{A}_b^{t,z}, \tilde\xv_j)$. Since we assume that no feature in $S$ interacts with any feature in $-Z$, the second and third term---which contain interactions between these two feature subsets---are zero. Hence, the risk function is only defined by feature interactions between features in $S$ and $Z$, which are minimized by the objective in Algorithm \ref{alg:tree}.

\medskip

\begin{proof}
If the feature subset $Z$ contains all features interacting with features in $S$, and hence no feature in $-Z$ interacts with any feature in $S$, then the risk function for feature $\xv_j$ within a subspace $\mathcal{A}_b^{t,z}$ reduces to the variance of feature interactions between feature $\xv_j$ and features in $Z$:
\begin{small}
    \begin{eqnarray*}
    \mathcal{R}_j(\mathcal{A}_b^{t,z}, \tilde\xv_j) 
    &=& \sum\limits_{\substack{k: k \in \{1,\ldots, m\}\\ \wedge x_j^{(k)} \in \mathcal{A}_b^{t,z}}} \sum\limits_{i: \xi \in \mathcal{A}_b^{t,z}}
    \left(\sum\limits_{l = 1}^{p-1} \; \sum\limits_{\substack{W \subseteq -j,\\ |W| = l}} g_{W \cup j}(x_j,\xv_W^{(i)}) - \E[ g_{W \cup j}(x_j,X_W)|\mathcal{A}_b^{t,z}] \right)^2
    \end{eqnarray*}
\end{small}
We can further expand the equation above as follows: 
\begin{small}
\begin{eqnarray*}
    \mathcal{R}_j(\mathcal{A}_b^{t,z}, \tilde\xv_j) 
    &=& \sum\limits_{\substack{k: k \in \{1,\ldots, m\}\\ \wedge x_j^{(k)} \in \mathcal{A}_b^{t,z}}} \sum\limits_{i: \xi \in \mathcal{A}_b^{t,z}}
     \left(\Bigg(\sum\limits_{l = 1}^{|Z\setminus j|} \; \sum\limits_{\substack{Z_l \subseteq Z \setminus j,\\ |Z_l| = l}} g_{Z_l \cup j}(x_j,\xv_{Z_l}^{(i)}) - \E[ g_{Z_l \cup j}(x_j,X_{Z_l})|\mathcal{A}_b^{t,z}]\Bigg)\right.  \\
     &+& \Bigg(\sum\limits_{l = 2}^{p-1} \; \sum\limits_{\substack{W \subseteq -j \\ \wedge \exists Z_l \subseteq Z\setminus j: Z_l \subset W \\ \wedge  \exists -Z_l \subseteq -Z\setminus j: -Z_l \subset W,\\ |W| = l}} \underbrace{g_{W \cup j}(x_j,\xv_{Z_l}^{(i)}, \xv_{{-Z}_l}^{(i)}) - \E[ g_{W \cup j}(x_j,X_{Z_l}, X_{{-Z}_l})|\mathcal{A}_b^{t,z}]}_\text{$=0$} \Bigg) \\
      &+& \left.\Bigg(\sum\limits_{l = 1}^{|-Z\setminus j|} \; \sum\limits_{\substack{-Z_l \subseteq -Z\setminus j,\\ |-Z_l| = l}} \underbrace{g_{-Z_l \cup j}(x_j,\xv_{{-Z}_l}^{(i)}) - \E[ g_{-Z_l \cup j}(x_j, X_{{-Z}_l})|\mathcal{A}_b^{t,z}] }_\text{$=0$}\Bigg)\right)^2 \\
     &=& \sum\limits_{\substack{k: k \in \{1,\ldots, m\}\\ \wedge x_j^{(k)} \in \mathcal{A}_b^{t,z}}} \sum\limits_{i: \xi \in \mathcal{A}_b^{t,z}}
     \left(\sum\limits_{l = 1}^{|Z\setminus j|} \; \sum\limits_{\substack{Z_l \subseteq Z \setminus j,\\ |Z_l| = l}} g_{Z_l \cup j}(x_j,\xv_{Z_l}^{(i)}) - \E[ g_{Z_l \cup j}(x_j,X_{Z_l})|\mathcal{A}_b^{t,z}] \right)^2 \\
\end{eqnarray*}
\end{small}

Since the objective is defined such that it minimizes these interactions for all $j \in S$ by splitting the feature space w.r.t. features in $Z$, we can split deep enough to achieve $g_{Z_l \cup j}(x_j,\xv_{Z_l}^{(i)}) = \E[ g_{Z_l \cup j}(x_j,X_{Z_l})|\mathcal{A}_b^{t,z}]$ for all terms within the sums of the risk function and for all $j \in S$. In other words, the individual interaction effect is equal to the expected interaction effect within a subspace. Thus, the theoretical minimum is $\mathcal{I}(t^*,z^*) = 0$. 
\end{proof}

\subsection{Proof of Theorem \ref{theorem:axiom_pd}}
\label{app:proof_pd} 


Here, we define $h$ to meet Axiom \ref{axiom:feat_rel} from Section \ref{sec:requirements} for PD where ICE curves represent local feature effects. The $i$-th ICE curve of feature $\xv_j$ can be decomposed as follows:
 \begin{eqnarray*}
\hat f(\xv_j, \xv_{-j}^{(i)})
&=& 
\underbrace{g_0}_\text{constant term} + \underbrace{g_j(\xv_j)}_\text{main effect of $\xv_j$} + 
\underbrace{\sum_{k \in -j} g_k(\xv_k^{(i)})}_{\substack{\text{main effect of all other} \\ \text{features in $-j$ for observation $i$}}} \\ &+&
\underbrace{\sum\limits_{k = 1}^{p-1}  \sum\limits_{\substack{W\subseteq -j, |W| = k}} g_{{W} \cup \{j\}}(\xv_j, \xi_{W})}_{\substack{\text{$(k+1)$-order interaction between} \\ \text{$\xv_j$ and features in $-j$ for observation $i$}}}
+ \underbrace{\sum\limits_{k = 2}^{p-1}  \sum\limits_{\substack{W\subseteq -j, |W| = k}} g_{{W}}(\xi_{W})}_{\substack{\text{$k$-order interaction between} \\ \text{features in $-j$ for observation $i$}}}
\end{eqnarray*}

However, this decomposition of the local feature effect of $\xv_j$ contains not only feature effects that depend on $\xv_j$, but also other effects (e.g., $i$-th main effects of features in $-j$), and thus Axiom \ref{axiom:feat_rel} is not fulfilled by ICE curves.\\

\medskip

\paragraph{\normalfont\textit{Proof Sketch}} By mean-centering ICE curves, constant and feature effects independent of $\xv_j$ are canceled out, and thus $\hat f^c(\xv_j, \xv_{-j}^{(i)})$ can be decomposed into the mean-centered main effect of $\xv_j$ and the $i$-th mean-centered interaction effect between $\xv_j$ and features in $-j$.\\ 

\medskip

\begin{proof}
    \begin{footnotesize}
\begin{flalign}
\hat f^c(x_j, \xv_{-j}^{(i)})
&= \hat f(x_j, \xv_{-j}^{(i)}) - \E\left[\hat f(X_j, \xv_{-j}^{(i)})\right]&&\\
&= g_0 + g_j(x_j) + \sum_{k \in -j} g_k(\xv_k^{(i)}) + \sum\limits_{k = 1}^{p-1}  \sum\limits_{\substack{W\subseteq -j, \\ |W| = k}} g_{{W} \cup \{j\}}(x_j, \xi_{W}) + \sum\limits_{k = 2}^{p-1}  \sum\limits_{\substack{W\subseteq -j, \\ |W| = k}} g_{{W}}(\xi_{W}) &&\\
&-g_0 - E\left[g_j(X_j)\right] - \sum_{k \in -j} g_k(\xv_k^{(i)}) - 
 E\left[\sum\limits_{k = 1}^{p-1}  \sum\limits_{\substack{W\subseteq -j, \\ |W| = k}} g_{{W} \cup \{j\}}(X_j, \xi_{W})\right] - \sum\limits_{k = 2}^{p-1}  \sum\limits_{\substack{W\subseteq -j, \\ |W| = k}} g_{{W}}(\xi_{W})&&\\
&= g_j(x_j) - E\left[g_j(X_j)\right] + \sum\limits_{k = 1}^{p-1}  \sum\limits_{\substack{W\subseteq -j, \\ |W| = k}} g_{{W} \cup \{j\}}(x_j, \xi_{W}) - E\left[\sum\limits_{k = 1}^{p-1}  \sum\limits_{\substack{W\subseteq -j, \\ |W| = k}} g_{{W} \cup \{j\}}(X_j, \xi_{W})\right] &&  \\
&= \underbrace{g^{c}_j(x_j)}_{\substack{\text{mean-centered} \\ \text{main effect of $\xv_j$}}} + \underbrace{\sum\limits_{k = 1}^{p-1}  \sum\limits_{\substack{W\subseteq -j, \\ |W| = k}} g^{c}_{{W} \cup \{j\}}(x_j, \xi_{W})}_{\substack{\text{ mean-centered interaction effect of} \\ \text{ $\xv_j$ with $\xv_{-j}^{(i)}$}}}
\end{flalign}
    
\end{footnotesize}

Thus, Axiom \ref{axiom:feat_rel} is satisfied by mean-centered ICE curves and can be used as local feature effect $h$ within GADGET.
\end{proof}

Following from that, the mean-centered PD for feature $\xv_j$ at $x_j$ can be decomposed by:
\begin{equation}
f^{PD, c}_j(x_j) = \E[\fh^c(x_j,X_{-j})] = g^{c}_j(x_j) + \sum\limits_{k = 1}^{p-1}  \sum\limits_{\substack{W\subseteq -j, \\ |W| = k}} E\left[g^{c}_{{W} \cup \{j\}}(x_j, X_{W})\right],
\end{equation}
which is the mean-centered main effect of feature $\xv_j$ and the expected mean-centered interaction effect with feature $\xv_j$ at feature value $x_j$.

\subsection{REPID as Special Case of GADGET}
\label{app:repid}
The objective function $\mathcal{I}(t,z)$ in Algorithm \ref{alg:tree} for $h = \hat f^c(x_j, \xv_{-j}^{(i)})$ is defined by the above loss function $\mathcal{L}_j^{PD}\left(\mathcal{A}_g, x_j\right)$ as follows:
\begin{eqnarray*}
   \mathcal{I}(t,z) = \sum_{j \in S} \sum_{g \in \{l,r\}} \sum_{k:k \in \{1,\ldots,m\} \wedge \xj^{(k)} \in \mathcal{A}_g} \mathcal{L}_j^{PD}\left(\mathcal{A}_g, \xj^{(k)}\right) 
\end{eqnarray*}
For the special case where we consider one feature of interest that we want to visualize ($S = j$) and all other features as possible split features ($Z = -j$), the objective function of GADGET reduces to: 
\begin{eqnarray*}
   \mathcal{I}(t,z) = \sum_{g \in \{l,r\}} \sum_{k = 1}^m \mathcal{L}_j^{PD}\left(\mathcal{A}_g, \xj^{(k)}\right),
\end{eqnarray*}
which is the same objective used within REPID. 
Thus, for the special case where we choose mean-centered ICE curves as local feature effect method and $S = j$ and $Z = -j$, GADGET is equivalent to REPID.

\subsection{Proof of Theorem \ref{theorem:axiom_ale}}
\label{app:proof_ale} 

Here, we show the fulfillment of Axiom \ref{axiom:feat_rel} from Section \ref{sec:requirements} for ALE.\\
\\
\textit{Proof Sketch} The local feature effect method used in ALE is the partial derivative of the prediction function at $\xv_j = x_j$. Thus, we define the local feature effect function $h$ by $\textstyle h(x_j,\xv_{-j}^{(i)}):=  \frac{\partial \fh(x_j, \xv_{-j}^{(i)})}{\partial x_j}$. We can decompose $h$ such that it only depends on main and interaction effects of and with feature $\xv_j$.

\medskip

\begin{proof}
    \begin{eqnarray*}
   \frac{\partial \fh(x_j, \xv_{-j}^{(i)})}{\partial x_j} &=& \frac{\partial \left( g_0 + \sum_{j = 1}^p g_j(x_j) + \sum_{j \neq k} g_{jk}(x_j,\xv_k^{(i)}) + \ldots + g_{12\ldots p}(\xv^{(i)})\right)}{\partial x_j}\\
   &=& \frac{\partial g_j(x_j)}{\partial x_j} + \sum\limits_{k = 1}^{p-1} \; \sum\limits_{\substack{W \subseteq -j,\\ |W| = k}} \frac{\partial g_{W \cup j}(\xv_j,\xv_W^{(i)})}{\partial x_j}
\end{eqnarray*}
\end{proof}
Taking the conditional expectation over the local feature effects (partial derivatives) at $x_j$ yields the (conditional) expected (i.e., global) feature effect at $x_j$: 
\begin{equation}
\E\Bigg[  \frac{\partial \fh(X_j, X_{-j})}{\partial x_j} \bigg \vert X_j = x_j \Bigg]   = \frac{\partial g_j(x_j)}{\partial x_j} + \sum\limits_{k = 1}^{p-1} \; \sum\limits_{\substack{W \subseteq -j,\\ |W| = k}} \E\left[\frac{\partial g_{W \cup j}(X_j,X_W)}{\partial x_j}\Bigg\vert X_j = x_j\right]
\end{equation}

\medskip
\subsection{Proof of Theorem \ref{theorem:axiom_sd}}
\label{app:proof_sd} 
Here, we show the fulfillment of Axiom \ref{axiom:feat_rel} defined in Section \ref{sec:requirements} for Shapley values, which are the underlying local feature effect in SD plots.\\
\\
\textit{Proof Sketch} The local feature effect function in the SD plot is the Shapley value. We define $\textstyle h(x_j,\xv_{-j}^{(i)}):=  \phi_j^{(i)}(x_j)$ to be the Shapley value for the $i$-th local feature effect at a fixed value $x_j$, which is typically the $i$-th feature value of $\xj$ (i.e., $x_j = x_j^{(i)}$). In Eq.~\eqref{eq:shapley}, we defined $\phi_j^{(i)}(x_j)$ according to \cite{herren_statistical_2022} which only depends on main and interaction effects of $\xv_j$.\\

\medskip

\begin{proof}
    \begin{eqnarray*}
   \phi_j^{(i)}(x_j) &=& \sum_{k=0}^{p-1}\frac{1}{k+1}\sum\limits_{\substack{W\subseteq -j:\\ |W| = k}} \left(\E[\hat f(x_j, X_{-j})|X_W = \xv_W^{(i)}] - \sum_{V \subset \{W\cup j\}} \E[\hat f(X)|X_V = \xv_V^{(i)}]\right)\\
   &=& g_j^c(x_j) + \sum_{k=1}^{p-1}\frac{1}{k+1}\sum_{W\subseteq -j: |W| = k} g_{W\cup j}^c(x_j,\xv_W^{(i)}),\\
\end{eqnarray*}
with $g_{W\cup j}^c(x_j,\xv_W^{(i)}) = \E[\hat f(x_j, X_{-j})|X_W = \xv_W^{(i)}] - \sum_{V \subset \{W\cup j\}} \E[\hat f(X)|X_V = \xv_V^{(i)}]$.\\
Hence, we can decompose $h$ such that it only depends on main effects of and interaction effects with feature $\xv_j$.
\end{proof}
Taking the expectation over the local feature effects $h = \phi_j$ at $x_j$ yields the expected (i.e., global) feature effect of Shapley values at $\xv_j = x_j$. 
\begin{equation}
\E_{X_W}[\phi_j(x_j)]   = g_j^c(x_j) + \sum_{k=1}^{p-1}\frac{1}{k+1}\sum_{W\subseteq -j: |W| = k} \E[g_{W\cup j}^c(x_j,X_W)]
\end{equation}
\\


    

\section{Further Characteristics of Feature Effect Methods}
In this section, we cover further characteristics of the different feature effect methods used within GADGET. As illustrated for PD in Section \ref{sec:pdp}, we also show here for ALE and SD that the joint feature effect (and possibly the prediction function) within the final regions of GADGET can be approximated by the sum of univariate feature effects. Hence, the joint feature effect can be additively decomposed into the features' main effects within the final regions. Furthermore, we provide an overview on estimates and visualization techniques for the regional feature effects and interaction-related heterogeneity for the different feature effect methods. We also explain how categorical features are handled within GADGET, depending on the underlying feature effect method.

\subsection{Decomposability of PD}
\label{app:decomp_pd}

According to \cite{friedman_greedy_2001} the usage of (multivariate) PD functions as additive components in a functional decomposition can recover the prediction function up to a constant. 
Thus, the prediction function can be decomposed into an intercept $g_0$ and the sum of mean-centered PD functions
similar to the recursive procedure of the standard functional ANOVA decomposition:
\begin{equation}
    \fh(\xv) = g_0 + \sum_{k=1}^p\sum_{\substack{W \subseteq \{1,\ldots,p\},\\ |W| = k}} \left( f_W^{PD,c}(\xv_W)  - \sum_{V \subset W} f_V^{PD,c}(\xv_V) \right),
    \label{eq:pd_fanova}
\end{equation}
with $\textstyle f_W^{PD,c}(\xv_W) = f_W^{PD}(\xv_W) - \int f_W^{PD}(\xv_W) d\xv_W$.
\cite{tan_considerations_2021} note that if the prediction function can be written as a sum of main effects, it can be exactly decomposed by an intercept plus the sum of all mean-centered 1-dimensional PD functions.

Based on this decomposition and if $Z$ contains all features interacting with features in $S$ and if GADGET is applied such that the theoretical minimum of the objective function is reached, then according to Theorem \ref{corollary:objective}, the joint mean-centered PD function $f^{PD, c}_{S|\mathcal{A}_g}$ within each final subspace $\mathcal{A}_g$ can be decomposed into the respective 1-dimensional mean-centered PD functions:  
\begin{equation}
   f^{PD, c}_{S|\mathcal{A}_g}(\xv_S) = \sum_{j \in S} f^{PD, c}_{j|\mathcal{A}_g} (\xv_j).
   \label{eq:decomp_pdp}
\end{equation}
Eq.~\eqref{eq:decomp_pdp} is justified by the assumption that no more interactions between features in S and other features are present in the final regions \citep{friedman_predictive_2008}.

Furthermore, if the subset $-S$ is the subset of features that do not interact with any other features (local feature effects are homogeneous), then according to Eq.~\eqref{eq:decomp_pdp} and Eq.~\eqref{eq:pd_fanova}, the prediction function $\fh_{\mathcal{A}_g}$ within each final subspace $\mathcal{A}_g$ can be decomposed into the 1-dimensional mean-centered PD functions of all $p$ features plus a constant value $g_0$:
\begin{equation}
   \fh_{\mathcal{A}_g}(\xv) = g_0 + \sum_{j =1}^p f^{PD, c}_{j|\mathcal{A}_g} (\xv_j). 
   \label{eq:decomp_pred_pdp}
\end{equation}
Thus, depending on how we choose the subsets $S$ and $Z$ and the extent to which we are able to minimize the interaction-related heterogeneity of feature effects by recursively applying Algorithm \ref{alg:tree}, we might be able to approximate the prediction function by an additive function of main effects of all features within the final regions.

This can also be shown for the simulation example illustrated in Figure \ref{fig:pdp_dt}, where $\fh^{PD, c}_{2|\mathcal{A}_g} (\xv_2) = 0$ (the regional effect of feature $\xv_2$ after the split is still $0$ and a has low interaction-related heterogeneity).
Instead, the regional effects of $\xv_1$ and $\xv_3$ vary compared to the root node and strongly reduce the interaction-related heterogeneity. Since the regional effects of $\xv_1$ and $\xv_3$ are approximately linear, we can estimate the prediction function within each subspace by
 $$\fh_{\mathcal{A}_l}(\xv) = g_0 + \frac{-3.04}{1.05}\xv_1 + \frac{0.99}{0.94}\xv_3 = g_0 - 2.9 \xv_1 + 1.05 \xv_3$$  
 and 
 $$\fh_{\mathcal{A}_r}(\xv) = g_0 + \frac{3.08}{1.05}\xv_1 + \frac{0.89}{0.83}\xv_3 = g_0 + 2.93 \xv_1 + 1.07 \xv_3,$$  
which is a close approximation to the true functional relationship of the data-generating process and thus provides a better understanding of how the features of interest influence the prediction function compared to only considering the global PD plots.

Note that besides the decomposability property in Eq.~\eqref{eq:decomp_pdp}, each global PD of features in $S$ is a weighted additive combination of the final regional PDs. Thus, each global PD can be additively decomposed into regional PD.\footnote{This decomposition is not in general true for mean-centered PD curves, since the mean-centering constant might be changed in the partitioning process.}

\subsection{Decomposability of ALE}
\label{app:decomp_ale}
\begin{figure}[tbh]
    \centering
      \includegraphics[width=0.49\linewidth]{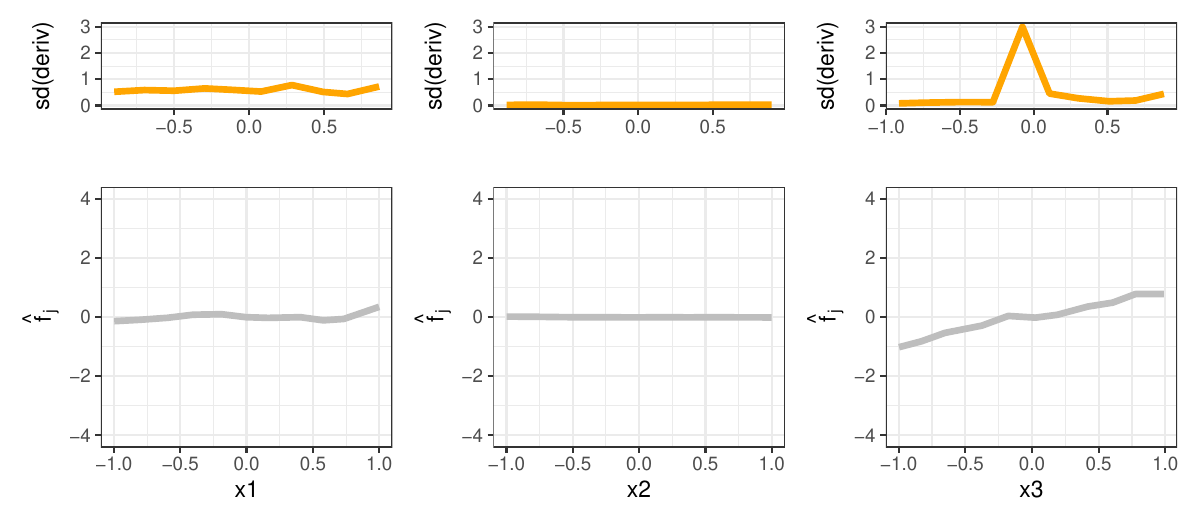}
      \scalebox{1}{
      \hspace{15pt} 
      \begin{tikzpicture}
      \usetikzlibrary{arrows}
        \usetikzlibrary{shapes}
         \tikzset{treenode/.style={draw}}
         \tikzset{line/.style={draw, thick}}
        \node [treenode](a0) {} ; [below=1pt,at=(4,0)]  {};
         \node [treenode, below=0.3cm, at=(a0.south), xshift=-3.0cm]  (a1) {};
         \path [line] (a0.south) -- + (0,-0.2cm) -| (a1.north) node [midway, above] {};
         \node[ at=(a0.south), xshift=-1.5cm] {$x_3 \leq -0.003$};
      \end{tikzpicture}
      \hspace{35pt}
      \begin{tikzpicture}
      \usetikzlibrary{arrows}
        \usetikzlibrary{shapes}
         \tikzset{treenode/.style={draw}}
         \tikzset{line/.style={draw, thick}}
        \node [treenode] (a01) {};[below=5pt,at=(node1.south) , xshift=3.5cm]
         \node [treenode, below=0.3cm, at=(a01.south), xshift=3.0cm]  (a2) {};
         \path [line] (a01.south) -- +(0,-0.2cm) -|  (a2.north) node [midway, above] {};
         \node[at=(a01.south), xshift=1.5cm]{$x_3 > -0.003$};
      \end{tikzpicture}
      }
    \includegraphics[width=0.49\linewidth]{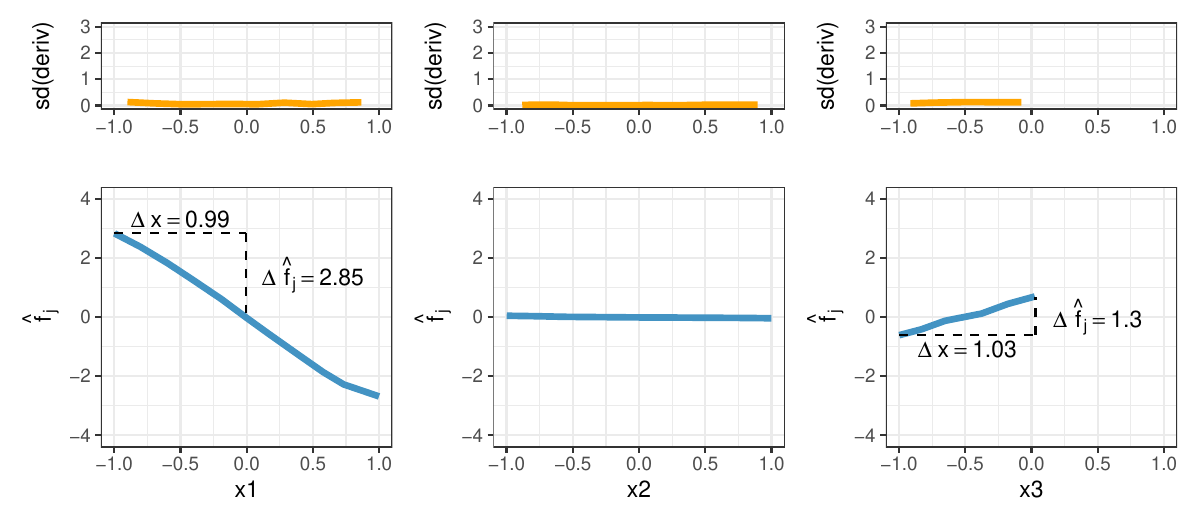}
    \includegraphics[width=0.49\linewidth]{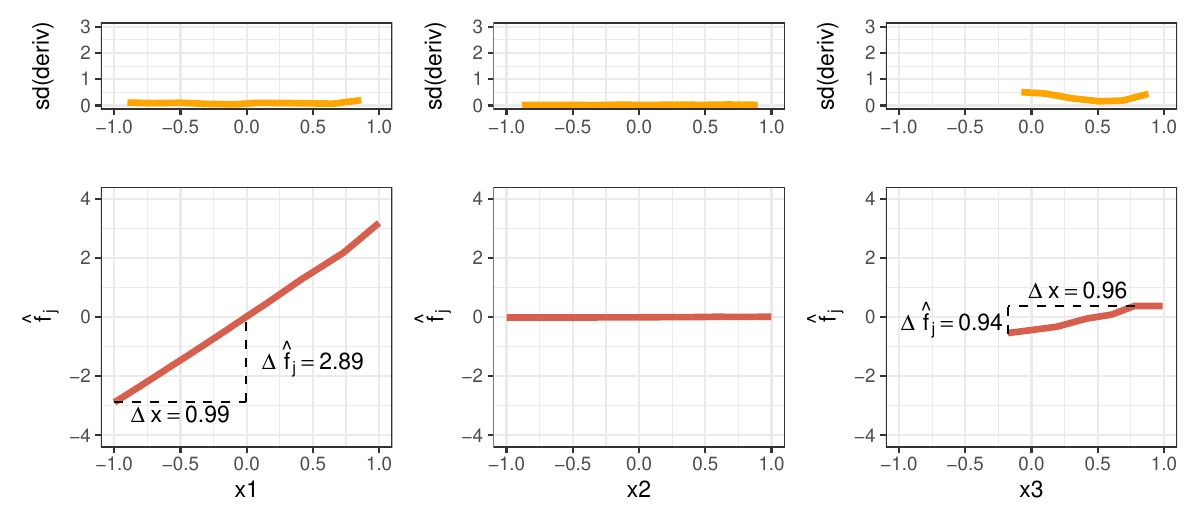}
    \vspace{.2in}
    \caption{Visualization of applying GADGET with $S = Z = \{1,2,3\}$ to derivatives of ALE for the uncorrelated simulation example of Section \ref{sec:pdp} with $Y = 3X_1\mathbbm{1}_{X_3>0} - 3X_1\mathbbm{1}_{X_3\leq 0}  + X_3 + \epsilon$ with $\epsilon \sim \mathbb{N}(0,0.09)$. The upper plots show the standard deviation of the derivatives (yellow) and the ALE curves (grey) on the entire feature space, while the lower plots represent the respective standard deviation of the derivatives and regional ALE curves after partitioning the feature space w.r.t. $\xv_3 = -0.003$.
    }
    \label{fig:ale_dt}
\end{figure}

Figure \ref{fig:ale_dt} shows the effect plots when GADGET-ALE is applied to the uncorrelated simulation example of Section \ref{sec:pdp} with $S = Z = \{1, 2, 3\}$. The grey curves before the split illustrate the global ALE curves which typically do not show the interaction-related heterogeneity of the underlying local effects.
Hence, we added a plot that visualizes this heterogeneity by providing the standard deviation of the underlying derivatives within each interval as a (yellow) curve along the range of $\xv_j$, similar to the idea of \cite{goldstein_peeking_2015} for derivative ICE curves. This shows that the local effects for the feature $\xv_2$ are very homogeneous across its entire range, while $\xv_1$ exhibits consistently heterogeneous local effects, and $\xv_3$ displays a high heterogeneity specifically around $\xv_3 = 0$. GADGET chooses $\xv_3 = -0.003$ as the best split point, significantly reducing the interaction-related heterogeneity among the three features. Thus, the ALE curves in the subspaces more accurately represent the underlying individuals.

Similar to PD plots, ALE plots include an additive recovery and can therefore be decomposed additively into main and interaction effects, as defined in Eq.~\eqref{eq:ale_fanova}. \cite{apley_visualizing_2020} defined the $W-$th order ALE function by the pure $W$-th order (interaction) effect similar to the functional ANOVA decomposition from Eq.\eqref{eq:fANOVA}. 
Furthermore, \cite{apley_visualizing_2020} show that the ALE decomposition has an orthogonality-like property, which guarantees (similar to the generalized functional ANOVA) that the following decomposition is unique (for details on estimating higher-order ALE functions, see \cite{apley_visualizing_2020}): 
\begin{equation}
    \fh(\xv) = g_0 + \sum_{\substack{W \subseteq \{1,\ldots,p\},\\ |W| = k}}  \fh_W^{ALE,c}(\xv_W) 
     \label{eq:ale_fanova}.
 \end{equation}

Furthermore, if $Z$ includes all features interacting with those in $S$ and GADGET achieves the theoretical minimum of the objective function, then according to Theorem \ref{corollary:objective}, the joint mean-centered ALE function $f^{ALE, c}_{S|\mathcal{A}_g}$ within each final subspace $\mathcal{A}_g$ can be decomposed into the 1-dimensional mean-centered ALE functions of features in $S$. Consequently, with no further interactions between features in $S$ and other features in the final regions, $f^{ALE, c}_{S|\mathcal{A}_g}$ can be uniquely decomposed into the mean-centered main effects of features in $S$---just as in PD functions \citep{apley_visualizing_2020}: 
\begin{equation}
   f^{ALE, c}_{S|\mathcal{A}_g}(\xv_S) = \sum_{j \in S} f^{ALE, c}_{j|\mathcal{A}_g} (\xv_j) 
   \label{eq:decomp_ale}
\end{equation}

Moreover, let $-S$ be the subset of features that do not interact with any other features. Then, according to Eq.~\eqref{eq:decomp_ale} and Eq.~\eqref{eq:pd_fanova}, the prediction function $\fh_{\mathcal{A}_g}$ within the region $\mathcal{A}_g$ can be decomposed into the 1-dimensional mean-centered ALE functions of all $p$ features, plus some constant value $g_0$:
\begin{equation}
   \fh_{\mathcal{A}_g}(\xv) = g_0 + \sum_{j =1}^p f^{ALE, c}_{j|\mathcal{A}_g} (\xv_j). 
   \label{eq:decomp_pred_ale}
\end{equation}
We can again illustrate this decomposition in Figure \ref{fig:ale_dt}, where $\xv_2$ shows an effect of $0$ with low heterogeneity before and after the split. The feature effects of $\xv_1$ and $\xv_3$ show high heterogeneity before the split, which is almost completely minimized after the split w.r.t. $\xv_3$. Hence, the resulting regional (linear) ALE curves are representative estimates for the underlying local effects. Therefore, we can approximate the prediction function within each subspace by
 $$\fh_{\mathcal{A}_l}(\xv) = g_0 + \frac{-2.85}{0.99}\xv_1 + \frac{1.3}{1.03}\xv_3 = g_0 - 2.89 \xv_1 + 1.26 \xv_3$$  
 and 
 $$\fh_{\mathcal{A}_r}(\xv) = g_0 + \frac{2.89}{0.99}\xv_1 + \frac{0.94}{0.96}\xv_3 = g_0 + 2.92 \xv_1 + 0.98 \xv_3.$$


\paragraph{\normalfont\textit{Particularities of ALE Estimation.}}\label{sec:abruptinteract}
As seen for the continuous feature $\xv_3$ in the simulation example presented here, abrupt interactions (``jumps'')\footnote{With abrupt interaction, we mean interactions that lead to an abrupt change of the influence of one feature ($\xv_1$) based on the influence of another feature at a specific (``jump'') point ($\xv_3 = 0$) like the feature interaction between $\xv_1$ and $\xv_3$ in the here presented simulation example: $Y = 3X_1\mathbbm{1}_{X_3>0} - 3X_1\mathbbm{1}_{X_3\leq 0}  + X_3 + \epsilon$ with $\epsilon \sim \mathbb{N}(0,0.09)$.} might be difficult to estimate for models that learn smooth effects, such as NNs (used here) or SVMs---especially when compared to models such as decision trees. Hence, depending on the model, these type of feature interactions can lead to very high partial derivatives in a region around the ``jump'' point instead of a high partial derivative at exactly the one specific ``jump'' point (here: $\xv_3 = 0$), thus leading to non-reducible heterogeneity (see upper right plot in Figure \ref{fig:ale_dt}). The standard deviation of the derivatives of $\xv_3$ are very high in the region around and not exactly at $\xv_3 = 0$. This interaction-related heterogeneity should be (almost) completely reduced when splitting w.r.t. $\xv_3 = 0$. However, high values may persist if the model did not perfectly capture this kind of interaction. To account for this issue in the estimation and partitioning process within GADGET, we use the following procedure for continuous features: 
In the two new subspaces after a split, if the derivatives of feature values close to the split point vary at least twice as much (measured by the standard deviation) as those of other observations within each subspace, we replace these derivatives close to the split point with values drawn from a normally distributed random variable where mean and variance are estimated by the derivatives of the remaining observation within each subspace.  

\subsection{Decomposability of SD}
\label{app:decomp_shap}

\paragraph{\normalfont\textit{Recalculation Versus No Recalculation of Shapley Values.}} 
In Section \ref{sec:shap}, we argued that 
Shapley values must be recalculated after each partitioning step to ensure that each new subspace 
to receive SD effects in the final subspaces that are representative of the underlying main effects within each subspace. 
Meanwhile, the unconditional expected value (i.e., the feature interactions on the entire feature space) are minimized without recalculating the conditional expected values. 

\begin{figure}[tbh]
    \centering
      \includegraphics[width=0.49\linewidth]{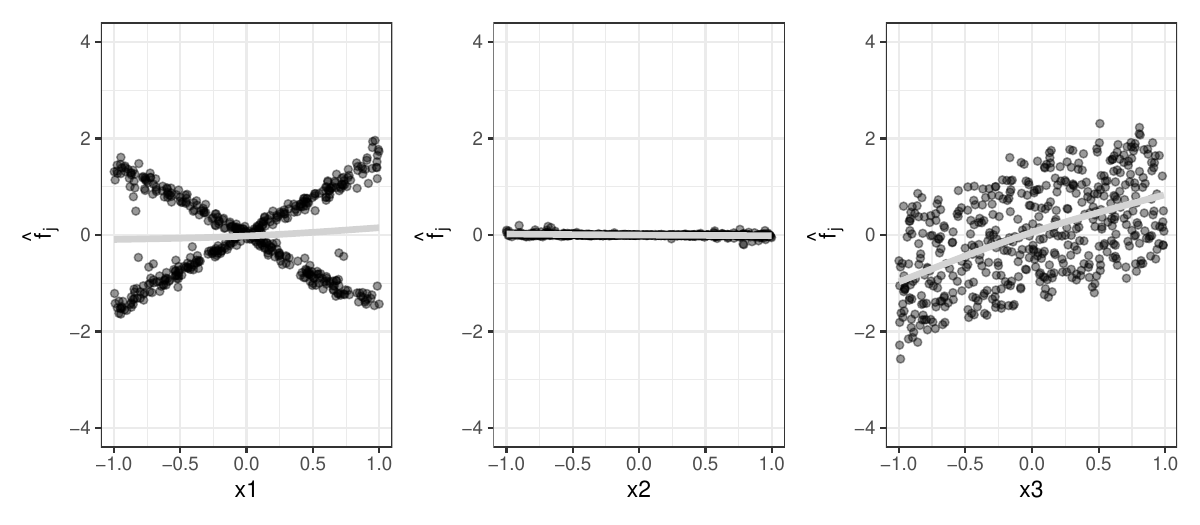}
      \scalebox{1}{
      \hspace{15pt} 
      \begin{tikzpicture}
      \usetikzlibrary{arrows}
        \usetikzlibrary{shapes}
         \tikzset{treenode/.style={draw}}
         \tikzset{line/.style={draw, thick}}
        \node [treenode](a0) {} ; [below=1pt,at=(4,0)]  {};
         \node [treenode, below=0.3cm, at=(a0.south), xshift=-3.0cm]  (a1) {};
         \path [line] (a0.south) -- + (0,-0.2cm) -| (a1.north) node [midway, above] {};
         \node[at=(a0.south), xshift=-1.5cm]{$x_3 \leq 0.007$};
      \end{tikzpicture}
      \hspace{35pt}
      \begin{tikzpicture}
      \usetikzlibrary{arrows}
        \usetikzlibrary{shapes}
         \tikzset{treenode/.style={draw}}
         \tikzset{line/.style={draw, thick}}
        \node [treenode] (a01) {};[below=5pt,at=(node1.south) , xshift=3.5cm]
         \node [treenode, below=0.3cm, at=(a01.south), xshift=3.0cm]  (a2) {};
         \path [line] (a01.south) -- +(0,-0.2cm) -|  (a2.north) node [midway, above] {};
         \node[at=(a01.south), xshift=1.5cm]{$x_3 > 0.007$};
      \end{tikzpicture}
      }
    \includegraphics[width=0.49\linewidth]{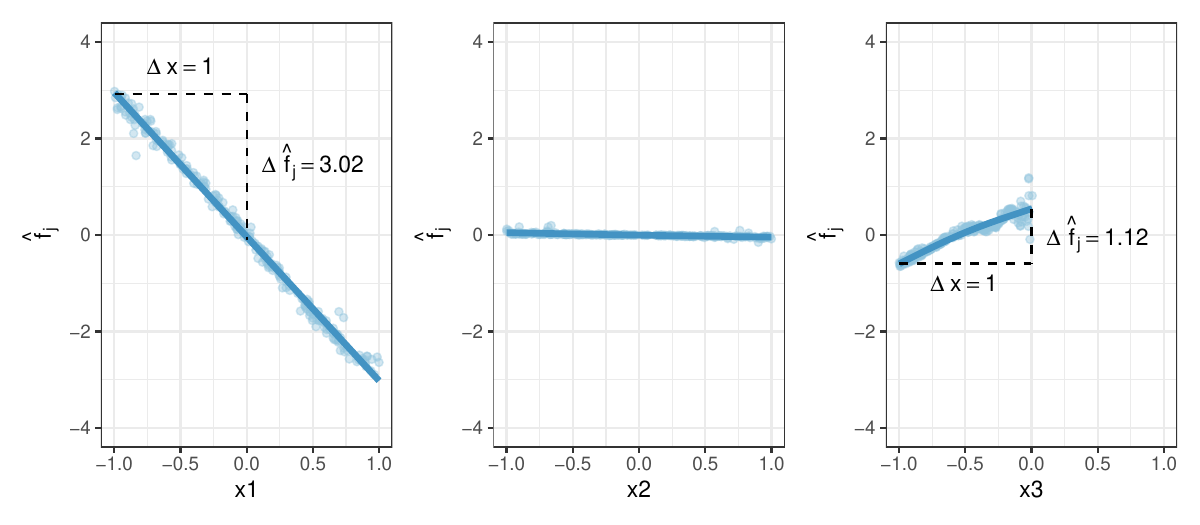}
    \includegraphics[width=0.49\linewidth]{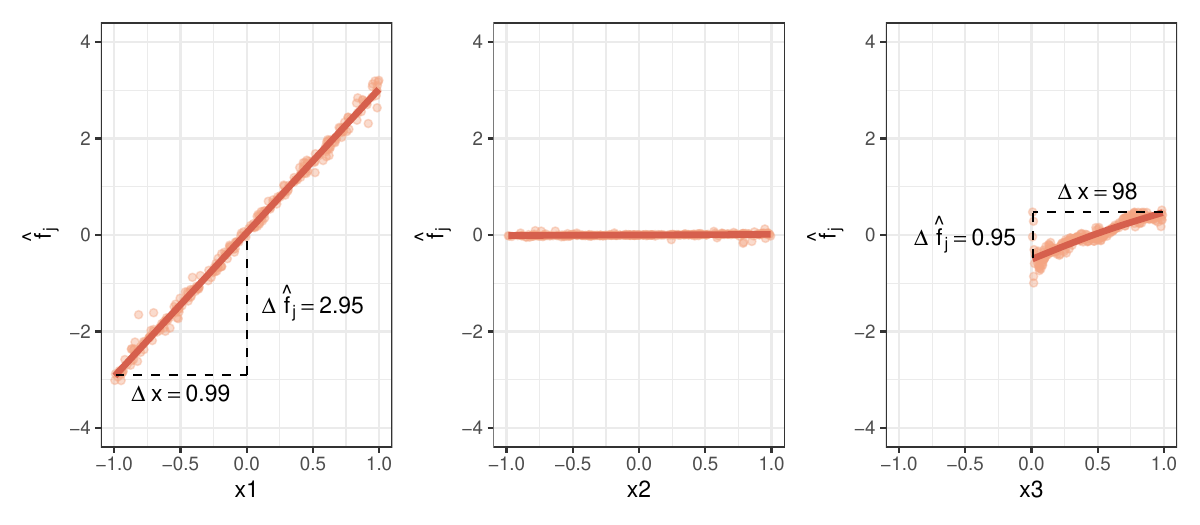}
    \vspace{.2in}
    \caption{Visualization of applying GADGET with $S = Z = \{1,2,3\}$ to Shapley values of the uncorrelated simulation example of Section \ref{sec:pdp} with $Y = 3X_1\mathbbm{1}_{X_3>0} - 3X_1\mathbbm{1}_{X_3\leq 0}  + X_3 + \epsilon$ with $\epsilon \sim \mathbb{N}(0,0.09)$. The upper plots show the Shapley values and the global estimated SD curve on the entire feature space, while the lower plots represent the respective Shapley values and regional SD curves after partitioning the feature space w.r.t. $\xv_3 = 0.007$.
    }
    \label{fig:shap_dt}
\end{figure}

The difference in the final feature effects within the subspaces is illustrated when comparing the left lower plot of Figure \ref{fig:repid_extrapol} (split without recalculation) with the respective plots of feature $\xv_1$ of Figure \ref{fig:shap_dt}. Without recalculation, the effect of feature $\xv_1$ is still regarded as an interaction effect between $\xv_1$ and $\xv_3$, and hence only half of the joint interaction effect is assigned to $\xv_1$ (i.e., the respective slope within the regions is $1.5$ and $-1.5$ instead of $3$ and $-3$), and the other half of the joint interaction effect is assigned to $\xv_3$. When Shapley values are recalculated after the first partitioning step within each subspace, we can see in Figure \ref{fig:shap_dt} that no more interactions are present between $\xv_1$ and $\xv_3$ within each subspace, due to the split w.r.t. $\xv_3$. Hence, the effect of $\xv_1$ is recognized as the main effect, with a slope closely matching the true functional relationship. Furthermore, recalculation also reduces the heterogeneity of feature effects of $\xv_3$ due to interactions with $\xv_1$. 

\textit{Note:} If the feature we use for partitioning the feature space ($z \in Z$) coincides with the features of interest ($S$), then the Shapley values should be recalculated in Algorithm \ref{alg:tree} to find the best split point (at least, if we choose the approach with recalculation after each partitioning step). The reason is that if $z\in S$, we also want to reduce the interaction-related heterogeneity within $z$ that is not accounted for if we do not recalculate the Shapley values within the new subspace. For example, in Figure \ref{fig:shap_dt}, we split according to $\xv_3$, which is also a feature of interest (here: $\{3\} \in S$). If we do not recalculate the Shapley values for $\xv_3$ within the splitting process, then the sum of the risk of any two subspaces for $\xv_3$ will always be approximately the same as the risk of the parent node, and thus the heterogeneity reduction for $\xv_3$ (see regional plots in Figure \ref{fig:shap_dt}) is not considered in the objective of Algorithm \ref{alg:tree}.

\paragraph{\normalfont\textit{Decomposition.}}
\cite{herren_statistical_2022} showed that the Shapley value of feature $\xv_j$ at feature value $x_j$ can be decomposed according to the functional ANOVA decomposition into main and interaction effects\footnote{Similar decompositions have been introduced in \cite{hiabu2022unifying} and \cite{bordt2022shapley}.}:
\begin{eqnarray}
    \phi_j^{(i)}(x_j) &=& \sum_{k=0}^{p-1}\frac{1}{k+1}\sum\limits_{\substack{W\subseteq -j:\\ |W| = k}} \left(\E[\hat f(x_j, X_{-j})|X_W = \xv_W^{(i)}] - \sum_{V \subset \{W\cup j\}} \E[\hat f(X)|X_V = \xv_V^{(i)}]\right)\nonumber\\
    &=& g_j^c(x_j) + \sum_{k=1}^{p-1}\frac{1}{k+1}\sum_{W\subseteq -j: |W| = k} g_{W\cup j}^c(x_j,\xv_W^{(i)}).
    \label{eq:shapley}
 \end{eqnarray}
Thus, in the case of interventional (i.e., marginal-based) Shapley values, Eq.~\ref{eq:shapley} are given by weighted PD functions.
Hence, if the global SD feature effect as defined in Eq.~\eqref{eq:shap_global} is considered, the same decomposition rules as defined for PD plots apply.
In other words, if $Z$ contains all features that interact with features in $S$ and if GADGET is applied such that the theoretical minimum of the objective function is reached, then according to Theorem \ref{corollary:objective}, the following decomposition in 1-dimensional global SD effect functions of features in $S$ holds:  
\begin{equation}
   f^{SD}_{S|\mathcal{A}_g}(\xv_S) = \sum_{j \in S} f^{SD}_{j|\mathcal{A}_g} (\xv_j). 
   \label{eq:decomp_shap}
\end{equation}

If all features containing heterogeneous effects (feature interactions) are included in the subset $S$, and the subset $Z$ consists of all features that interact with features in $S$, then according to Eq.~\eqref{eq:decomp_shap} and Eq.~\eqref{eq:pd_fanova}, the prediction function $\fh_{\mathcal{A}_g}$ within the region $\mathcal{A}_g$ can be uniquely decomposed into the 1-dimensional global SD effect functions of all $p$ features, plus some constant value $g_0$:
\begin{equation}
   \fh_{\mathcal{A}_g}(\xv) = g_0 + \sum_{j =1}^p f^{SD}_{j|\mathcal{A}_g} (\xv_j). 
   \label{eq:decomp_pred_shap}
\end{equation}

Again, we can derive this decomposition from Figure \ref{fig:ale_dt} in the same way we did for PD and ALE plots. Hence, we can approximate the prediction function within each subspace by
 $\fh_{\mathcal{A}_l}(\xv) = g_0 - 3.02 \xv_1 + 1.12 \xv_3$  
 and 
 $\fh_{\mathcal{A}_r}(\xv) = g_0 + 2.98 \xv_1 + 1.03 \xv_3.$

\paragraph{\normalfont\textit{Equivalence of SD and Mean-Centered PD.}}
According to \cite{herren_statistical_2022}, the Shapley value $\phi_j^{(i)}(x_j)$ of the $i$-th observation at $\xv_j = x_j$ can be decomposed as in Eq.~\eqref{eq:shapley}:
\begin{eqnarray*}
   \phi_j^{(i)}(x_j) &=& \sum_{k=0}^{p-1}\frac{1}{k+1}\sum\limits_{\substack{W\subseteq -j:\\ |W| = k}} \left(\E[\hat f(x_j, X_{-j})|X_W = \xv_W^{(i)}] - \sum_{V \subset \{W\cup j\}} \E[\hat f(X)|X_V = \xv_V^{(i)}]\right)\\
   &=& g_j^c(x_j) + \sum_{k=1}^{p-1}\frac{1}{k+1}\sum_{W\subseteq -j: |W| = k} g_{W\cup j}^c(x_j,\xv_W^{(i)}),
\end{eqnarray*}
with $g_{W\cup j}^c(x_j,\xv_W^{(i)}) = \E[\hat f(x_j, X_{-j})|X_W = \xv_W^{(i)}] - \sum_{V \subset \{W\cup j\}} \E[\hat f(X)|X_V = \xv_V^{(i)}]$.
As in Eq.~\eqref{eq:shap_global}, the global feature effect (SD) of feature $\xv_j$ at $x_j$ is then defined by
$$
f^{SD}_j(x_j) = \E_{X_W}[\phi_j]   = g_j^c(x_j) + \sum_{k=1}^{p-1}\frac{1}{k+1}\sum_{W\subseteq -j: |W| = k} \E[g_{W\cup j}^c(x_j,X_W)]
$$

Hence, if Theorem \ref{corollary:objective} is satisfied, if the joint global SD effect of features in $S$ can be decomposed into the univariate SD effects as in Eq.~\eqref{eq:decomp_shap}, and if the interventional approach for Shapley calculation is used, then all feature interactions are zero, and the global SD effect of feature $\xv_j$ at $x_j$ is given by
\begin{eqnarray*}
    f^{Shap}_j(x_j) &=&  g_j^c(x_j) + \sum_{k=1}^{p-1}\frac{1}{k+1}\sum_{W\subseteq -j: |W| = k} \E[g_{W\cup j}^c(x_j,X_W)]\\
    &\stackrel{\text{T. \ref{corollary:objective}}}{=}& g_j^c(x_j)\\
    &=& \E[\hat f(x_j, X_{-j})] - \E[\hat f(X)],
\end{eqnarray*}
which is equivalent to the mean-centered PD of feature $\xv_j$ at $x_j$.

\subsection{Overview on Estimates and Visualizations}
\label{app:overview_estimates}
We provide here an overview on the estimates and visualization techniques for PD, ALE, and SD within GADGET that we introduced in Sections \ref{sec:pdp}-\ref{sec:shap}.

\paragraph{\normalfont\textit{Local Effect.}}
The local effect $h$ for a feature $\xv_j$ at feature value $x_j$ used within GADGET is estimated by
\begin{itemize}
    \item \textbf{PD}: mean-centerd ICE $\hat h^{(i)} = \fh^c(x_j, \xi_{-j})$
    \item \textbf{ALE}: partial derivatives estimated by prediction differences within $k$-th interval $\hat h^{(i)} = \fh(z_{k-1, j}, \xi_{-j}) - \fh(z_{k-1, j}, \xi_{-j})$ where $x_j \in ]z_{k-1, j}, z_{k, j}]$ 
    \item \textbf{SD}: Shapley value $\hat h^{(i)} = \hat \phi_j^{(i)}$
\end{itemize}

\paragraph{\normalfont\textit{Regional Effect.}}
The feature effect for a feature $\xv_j$ at feature value $x_j$ within a subspace/region $\mathcal{A}_g$ of GADGET is estimated by
\begin{itemize}
    \item \textbf{PD}: mean-centered regional PD $\hat f^{PD,c}_{j|\mathcal{A}_g}(x_j) = \frac{1}{|N_g|}\sum_{i \in N_g}\fh^c(x_j, \xi_{-j})$ with $N_g$ being the index set of all $i: \xi \in \mathcal{A}_g$. 
    \item \textbf{ALE}: regional ALE $
  \hat{{f}}^{ALE}_{j|\mathcal{A}_g}(x_j) = \sum_{k = 1}^{k_j(x_j)}\frac{1}{|N_g(k)|}\sum_{i \in N_g(k)}\left[\fh(z_{k, j}, \xi_{-j}) -\fh(z_{k-1, j}, \xi_{-j})\right]$ with $N_g(k)$ being the index set of all $i: \; x_j \in \; ]z_{k-1, j}, z_{k, j}] \wedge \xi \in \mathcal{A}_g$.
    \item \textbf{SD}: regional SD $\hat{{f}}^{SD}_{j|\mathcal{A}_g}(x_j)$ is estimated by fitting a GAM on $\{x_j^{(i)}, \hat \phi_j^{(i)}\}_{i: \xi \in \mathcal{A}_g}$.
\end{itemize}
The regional effect for feature $\xv_j$ is visualized for all $x_j \in \mathcal{A}_g$. Therefore, the respective GAM curve is plotted in the case of SD, while we linearly interpolate between the grid-wise/interval-wise estimates of PD/ALE to receive the regional effect curves.

\paragraph{\normalfont\textit{Interaction-Related Heterogeneity.}}
The interaction-related heterogeneity for a feature $\xv_j$ at feature value $x_j$ within a subspace/region $\mathcal{A}_g$ of GADGET is estimated by the loss function in Eq. \eqref{eq:loss}, which quantifies the variance of local effects at $x_j$ and is visualized by
\begin{itemize}
    \item \textbf{PD}: $95\%$ interval for (mean-centered) regional PD $\Big[\hat f^{PD,c}_{j|\mathcal{A}_g}(x_j) \pm 1.96\cdot \sqrt{\hat{\mathcal{L}_j}^{PD}(\mathcal{A}_g,x_j)}\Big]$. 
    \item \textbf{ALE}: standard deviation of local effects $\sqrt{\hat{\mathcal{L}_j}^{ALE}(\mathcal{A}_g,x_j)}$.
    \item \textbf{SD}: Shapley values are recalculated within each region $\mathcal{A}_g$ and plotted with the fitted GAM for the regional SD effect to visualize the variation of local feature effects aka interaction-related heterogeneity.
\end{itemize}

For each feature $j \in S$, we generate one figure showing the regional effect curves of all final regions we obtain after applying GADGET. For PD, the regional effect curves are accompanied with intervals showing how much interaction-related heterogeneity remains in the underlying local effects (see, e.g., Figures \ref{fig:compas_region}). For ALE, a separate plot visualizes the interaction-related heterogeneity via the standard deviation of local effects, which is inspired by the derivative ICE plots of \cite{goldstein_peeking_2015} (see, e.g., Figure \ref{fig:bike}). For SD, the Shapley values that were recalculated conditioned on each subspace $\mathcal{A}_g$ are visualized with the regional effect curve (see, e.g., Figure \ref{fig:compas_region_shap}).

Note that we can also visualize the non-centered PD ($\fh^{PD}_{j|\mathcal{A}_g}$) instead of the mean-centered PD, which might provide more insights regarding interpretation. However, the interaction-related heterogeneity must be estimated by the mean-centered ICE curves to only represent heteroegeneity induced by feature interactions (see Appendix \ref{app:proof_pd}).

\subsection{Handling of Categorical Features}
\label{app:categ_feat}
In this section, we will summarize the particularities of categorical features. Compared to numeric features, we have a limited number of $K$ values (categories). 
%
Hence, compared to numeric features, we find split points by dividing the $K$ categories into the two new subspaces. Since GADGET is based on the general concept of a CART (decision tree) algorithm \citep{breiman:1984} that can handle categorical features, the splitting itself follows the same approach as for a common CART algorithm. 
If categorical features occur in $S$, the calculation of the objective function remains unchanged, and the handling of categorical features depends only on the underlying feature effect method:

\paragraph{\normalfont\textit{PD Plot.}}
Compared to numeric features, the grid points for categorical features are limited to the number of categories. Otherwise, the calculation of the loss and the risk (see, Eq.~\ref{eq:loss_pd} and Eq.~\ref{eq:risk}) works the same as for numeric features.\footnote{Since we calculate the loss point-wise at each grid point and sum it up over all grid points, the order of the category does not make a difference for the objective of GADGET.}

\paragraph{\normalfont\textit{ALE Plot.}}
ALE builds intervals based on quantiles for numeric features to calculate prediction differences between neighboring interval borders for all observations falling within this interval. For binary features, the authors solve this as follows: for all observations falling in each of the categories, the prediction difference when changing it to the other category is calculated. For more categories, they suggest a sorting algorithm.\footnote{For more information, we refer to \cite{apley_visualizing_2020}.} Hence, we still receive the needed derivatives for GADGET for each category to calculate the loss and risk function for GADGET.

\paragraph{\normalfont\textit{SD Plot.}}
Compared to numeric features, the x-axis of the SD plot is a grid of size $K$. For each of these grid points (categories), the Shapley values for the observations belonging to the specific category are calculated. Hence, instead of a spline to quantify the expected value in Eq.~\eqref{eq:shap_global}, we use the arithmetic mean within each category (similar to PD plots) and, thus, sum up the variance of Shapley values for each category over all categories (within the respective subspace).

In general, if we apply GADGET and split w.r.t. a categorical feature such that only one category is present within a subspace (e.g., we split the feature sex such that all individuals are male in one resulting subspace), then the interaction-related heterogeneity vanishes to zero since only an additive shift for the feature sex is left in this subspace.

\textit{Note:} If a categorical feature $\xv_j$ is not only considered for splitting ($j \in Z$) but is also a feature of interest ($j \in S$), the different splitting possibilities of categories of $\xv_j$ prompt recalculation for ALE, since derivatives are only calculated for pre-sorted neighboring categories. In our implementations, we only split w.r.t. the pre-sorted categories and considered them as integer values to reduce the computational burden of the calculations. 

\section{Definition and Specification of Stop Criteria for GADGET}
\label{app:recom_stop_crit}
%
The question of how many partitioning steps should be performed depends generally on the underlying research question. 
If the user is more interested in reducing the interaction-related heterogeneity as much as possible, they might split rather deeply, depending on the complexity of interactions learned by the model. However, this might lead to many regions that are more challenging to interpret. If the user is more interested in a small number of regions, they might prefer a shallow tree, thus reducing only the heterogeneity of the features that interact the most.

\subsection{Definition of Stop Criteria}


We propose the following stopping criteria to control the number of partitioning steps in GADGET: 
First, we can use common decision tree hyperparameters such as the tree depth or the minimum number of observations per leaf node. Alternatively, an early stop mechanism can be applied based on the interaction-related heterogeneity reduction.
According to our proposed split-wise measure, we further split only if the relative improvement is at least $\gamma \in [0,1]$ times the total relative heterogeneity reduction of the previous split: $\gamma \times \frac{\sum_{j \in S} \left(\mathcal{R}_j(\mathcal{A}_P, \tilde{\mathbf{x}}_j)\right) - \mathcal{I}(\hat{t}, \hat{z})}{\sum_{j \in S} \left(\mathcal{R}_j(\mathcal{X}, \tilde{\mathbf{x}}_j)\right)}$. Another option is to stop splitting once a pre-defined total reduction of heterogeneity ($R^2_{Tot}$) is achieved. Generally, higher $\gamma$ values and lower thresholds for $R^2_{Tot}$ result in fewer partitioning steps, and vice versa.

\subsection{Recommendations to Specify Stop Criteria}

GADGET performs binary splits, making it easier to minimize abrupt interactions (i.e., interactions where categorical or discretized features are involved, see also Section~\ref{sec:abruptinteract}). Conversely, minimizing smooth interactions (i.e., interactions between continuous features) may require multiple splits. 
%
%
We recommend a maximum tree depth between 3 and 6 to maintain some degree of interpretability and setting \(\gamma\) (the minimum relative risk reduction for a split) between 0.1 and 0.25 to further limit the number of terminal regions. 
%
%
For abrupt interactions (with few interactions), the risk reduction for the first split is usually quite high. Therefore, to find further feature interactions, we recommend to set $\gamma$ towards the lower end of the recommended range (e.g., 0.1 to 0.2). 
In the case of smooth interactions, the risk reduction for the first split is usually rather low compared to subgroup-specific feature interactions. Therefore, setting $\gamma$ to a small value leads to a vanishing improvement which prevents the desired stop mechanism. However, the value should also not be too high to prevent the first split.
In summary, a tree depth of 3 to 6 and \(\gamma\) between 0.1 and 0.25 worked well in our examples. To precisely quantify feature interactions or to completely minimize them, more splits and a smaller $\gamma$ may be needed, possibly using $R^2_{Tot}$ as a stopping criterion.

\section{Theoretical Evidence of PINT}
\label{app:pint_proof}

Here, we provide the proofs to the Theorems defined in Section \ref{sec:interact_pimp}.

\subsection{Proof of Theorem \ref{thm:typeI}}
\label{app:theorem_pint}

\textit{Proof Sketch}
In this proof we show that the type I error $\P(\mathcal{R}_j(\widehat f_{\mathcal{D}_n}, \tilde\xv_j, \mathcal{X}) > z_{1 - \alpha})$ of the test is constrained by $\alpha + \epsilon$ with $\epsilon = \max_{t > 0} [\P(\mathcal{R}_j(\widehat f_{\mathcal{D}_n}, \tilde\xv_j, \mathcal{X}) > t) - \P( \mathcal{R}_j(\widehat f_{\mathcal{D}_n^\Pi}, \tilde\xv_j, \mathcal{X})  > t)]$. \\
\begin{proof}
	The type I error probability is
	\begin{align}
		&\quad  \P\biggl(\mathcal{R}_j(\widehat f_{\mathcal{D}_n}, \tilde\xv_j, \mathcal{X}) > z_{1 - \alpha}\biggr) \\
		&= \P\biggl( \mathcal{R}_j(\widehat f_{\mathcal{D}_n^\Pi}, \tilde\xv_j, \mathcal{X})  >  z_{1 - \alpha}\biggr) +  \biggl[\P\biggl(\mathcal{R}_j(\widehat f_{\mathcal{D}_n}, \tilde\xv_j, \mathcal{X}) >  z_{1 - \alpha}\biggr) - \P\biggl( \mathcal{R}_j(\widehat f_{\mathcal{D}_n^\Pi}, \tilde\xv_j, \mathcal{X})  >  z_{1 - \alpha}\biggr)\biggr] \\
		&\le \P\biggl( \mathcal{R}_j(\widehat f_{\mathcal{D}_n^\Pi}, \tilde\xv_j, \mathcal{X})  >  z_{1 - \alpha}\biggr) + \max_{t > 0} \biggl[\P\biggl(\mathcal{R}_j(\widehat f_{\mathcal{D}_n}, \tilde\xv_j, \mathcal{X}) > t\biggr) - \P\biggl( \mathcal{R}_j(\widehat f_{\mathcal{D}_n^\Pi}, \tilde\xv_j, \mathcal{X})  > t\biggr)\biggr] \\
		&\le \alpha + \epsilon,
	\end{align}	
where the last step follows from the definition of $z_{1- \alpha}$ and independence  of permutations. 
\end{proof}

\subsection{Proof of Theorem \ref{thm:typeII}}
\label{app:corollary_pint}

\textit{Proof Sketch}
Here, we show that the type II error of the PINT procedure strives towards $0$. Recall that a risk of 0 means that no interactions are learned by the model. If Condition \eqref{cond:2-1} holds---i.e., the risk under the permuted data set is less than every $\epsilon > 0$ with high probability---the type II error of the PINT procedure strives towards $0$. The last step follows from the assumption that feature interactions have been learned by the model fitted on the original data set. It follows that the power of the test strives towards $1$.\\

\begin{proof}
	Condition \eqref{cond:2-1} implies
	\begin{align}
		\P\biggl(\mathcal{R}_j(\widehat f_{\mathcal{D}_n^\Pi}, \tilde\xv_j, \mathcal{X}) > \epsilon\biggr) \to 0 \notag
	\end{align}
	for every $\epsilon > c > 0$ and thus $	\P(z_{1- \alpha} \ge c) \to 0$.
	The type II error probability is 
	\begin{align}
		  &\quad \P\biggl(\mathcal{R}_j(\widehat f_{\mathcal{D}_n},\tilde\xv_j, \mathcal{X}) \le z_{1 - \alpha}\biggr)  \\
		&\le \P\biggl(\mathcal{R}_j(\widehat f_{\mathcal{D}_n}, \tilde\xv_j, \mathcal{X}) \le z_{1 - \alpha}, z_{1- \alpha} \ge c\biggr) +  \P\biggl(\mathcal{R}_j(\widehat f_{\mathcal{D}_n}, \tilde\xv_j, \mathcal{X}) \le z_{1 - \alpha}, z_{1- \alpha} < c\biggr) \notag \\
		&\le \P\biggl( z_{1- \alpha} \ge c\biggr) + \P\biggl(\mathcal{R}_j(\widehat f_{\mathcal{D}_n}, \tilde\xv_j, \mathcal{X}) <  c\biggr)  \\
            &\to 0.  
	\end{align}
\end{proof}

\section{Further Empirical Validation of PINT}
\label{app:pint}


\subsection{Example for Failure of PINT}
\label{app:sim_pint_invalid}

\begin{figure}[tbh]
   \centering    \includegraphics[width=\textwidth]{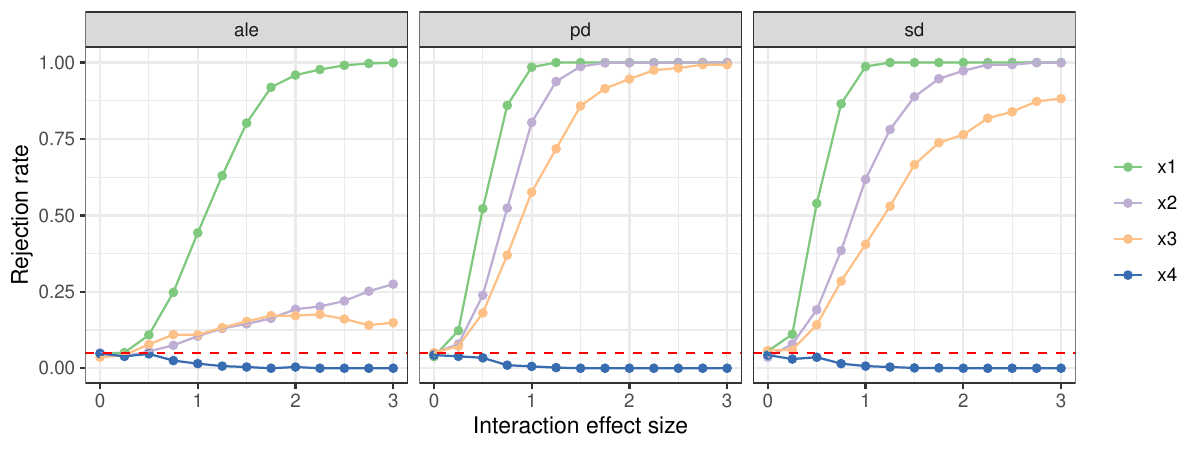}
   \caption{Rejection rates of PINT for $1000$ repetitions for all three feature effect methods and for varying interaction effect sizes $\beta$. For $\beta = 0$, all rejection rates correspond to type I errors. For $\beta > 0$,  the rejection rates of $\xv_1$ and $\xv_2$ correspond to the power of the test, those of $\xv_3$ and $\xv_4$ to type I errors. The red dashed line in the left plot represents the significance level of $\alpha =0.05$.}
    \label{fig:pint_eval_counter}
\end{figure}

Note that if the condition \eqref{cond:2-1} in Section \ref{sec:pint_validation} is not fulfilled, the type I error is not constrained by the significance level $\alpha$. This can be illustrated with the following example: Let $X_1,X_2,X_4 \sim \mathbbm{U}(-1,1)$ and $X_3 = \exp(X_2) + \delta$ with $\delta \sim \mathbbm{N}(0, 0.01)$. We draw $500$ observations and define the true functional relationship by $y = \beta \xv_1 \exp(\xv_2) + \epsilon$ with $\epsilon \sim \mathbbm{N}(0,1)$ and $\beta \in \{0,0.25,0.5,0.75,1,1.25,1.5,1.75,2,2.25,2.5,2.75,3\}$. We again fit an SVM with RBF kernel and apply PINT using the same hyperparameter settings as before. We repeat the experiment $1000$ times. In this setting, $\xv_3$ almost perfectly reflects the influence of $\exp(\xv_2)$ and therefore it is possible to learn the simpler linear $\beta \xv_1 \xv_3$ instead of the true more complex non-linear interaction effect. Hence, it is very likely that $\fh(\xv)$ does not accurately approximate $f(\xv)$ since $\fh(\xv)$ most likely learns an interaction between $\xv_1$ and the non-influential feature (in $f(\xv)$) $\xv_3$. Hence, this can lead to a higher type I error for $\xv_3$ than $\alpha$ as illustrated in Figure \ref{fig:pint_eval_counter}.

\subsection{Further Empirical Validation of PINT}
\label{app:further_examples_pint}

Here, we provide further evaluations and empirical validation of using PINT to select truly interacting features. The experiments in this section are based on the simulation study of Section \ref{sec:sim_pint} where we analyzed the robustness of PINT concerning potential spurious interactions and compared it to the H-Statistic. We now analyze in more detail the sensitivity of PINT for the three feature effect methods considering varying correlations of the features, a linear and non-linear true underlying function, differing effect sizes of the interaction effect and how the noise term of the true underlying function influences the rejection rate of PINT.
For all experiments, we consider drawing $500$ observations of the considered random variables and an SVM based on an RBF kernel with the hyperparameters Section \ref{sec:sim_pint}.
More specifically we consider the following combination of different simulation settings:

\paragraph{\normalfont\textit{Linear and non-linear functional relationships.}}
We consider two different true underlying functions. One with linear feature effects (as in Section \ref{sec:sim_pint}) and one more complex one with non-linear effects. Both of them include feature interactions between $\xv_1$ and $\xv_2$ and potential spurious interactions. We argue that the linear setting that we considered so far is rather simple leading to a model fit that is (almost) perfect without too much model uncertainty. Therefore, considering a more complex setting might lead to a worse and less certain model fit.

\begin{itemize}
    \item Linear true functional relationship: 
$\mathbf{y} = f_{lin}(\xv) + \epsilon_k$ with $f_{lin}(\xv) = \xv_1 + \xv_2 + \xv_3 - \beta\xv_1 \xv_2$ and $k \in \{1,2\}$.
\item Non-linear true functional relationship:
$\mathbf{y} = f_{nlin}(\xv) + \epsilon_k$ with $f_{nlin}(\xv) = \xv_1^2 + \xv_2^3 + \exp(\xv_3) - \beta\xv_1 \xv_2 + \epsilon_k$ with $k \in \{1,2\}$. 
\end{itemize}

\paragraph{\normalfont\textit{Correlations and spurious interactions.}}
We consider four features $X_1, X_2, X_4 \sim \mathbbm{U}(-1,1)$
and $X_3 = X_2 + \delta$ with $\delta \sim \mathbbm{N}(0,0.09)$ for the setting defined in Section \ref{sec:sim_pint} with a high linear correlation of $\rho_{23} \approx 0.9$ between $\xv_3$ and $\xv_2$. This will very likely lead to a spurious interaction between $\xv_1$ and $\xv_3$ as shown in the analysis of the mentioned section. To investigate how sensitive PINT is concerning this phenomenon, we consider a second smaller correlation which reduces the risk of spurious interactions. Therefore, we take into account a medium correlation by defining $X_3$ as a combination of $X_2$ and a uniformly distributed variable (similarly to Section \ref{sec:sim_extrapol}): $X_3 = 0.37 X_2 + 0.63 u$ with $u \sim \mathbbm{U}(-1,1)$ which leads to a linear correlation of $\rho_{12} \approx 0.5$. 
As a plausibility check and baseline, we consider all features to be independent and thus $X_3 \sim \mathbbm{U}(-1,1)$.

\paragraph{\normalfont\textit{Noise term.}} Regarding the noise term $\epsilon_k$ we take into consideration two alternatives: no noise term, i.e., $\epsilon_1 = 0$. This is the setting we used in Section \ref{sec:sim_pint}. Therefore, the noise did not influence the modeling approach and thus did also not influence the interaction detection with PINT. To analyze the influence of a noise term on PINT, we consider one that accounts for ten percent of the standard deviation of $f(X)$ (similar to the other simulations in this paper), i.e., $\epsilon_2 \sim \mathbbm{N}(0, 0.01 \cdot var(f(X)))$.

\paragraph{\normalfont\textit{Interaction effect size.}} Both functional relationships are also defined for different coefficient values $\beta$ for the interaction terms. In Section \ref{sec:sim_pint} we chose $\beta = 2$. The analysis showed that the interaction can be detected very well. Hence, we take into consideration further smaller values for $\beta$ to see how it affects the power of the test. Therefore, we consider $\beta \in \{0.75, 1, 1.25, 1.5, 1.75, 2\}$.

\begin{figure}[h]
    \centering
    \includegraphics[width=\textwidth]{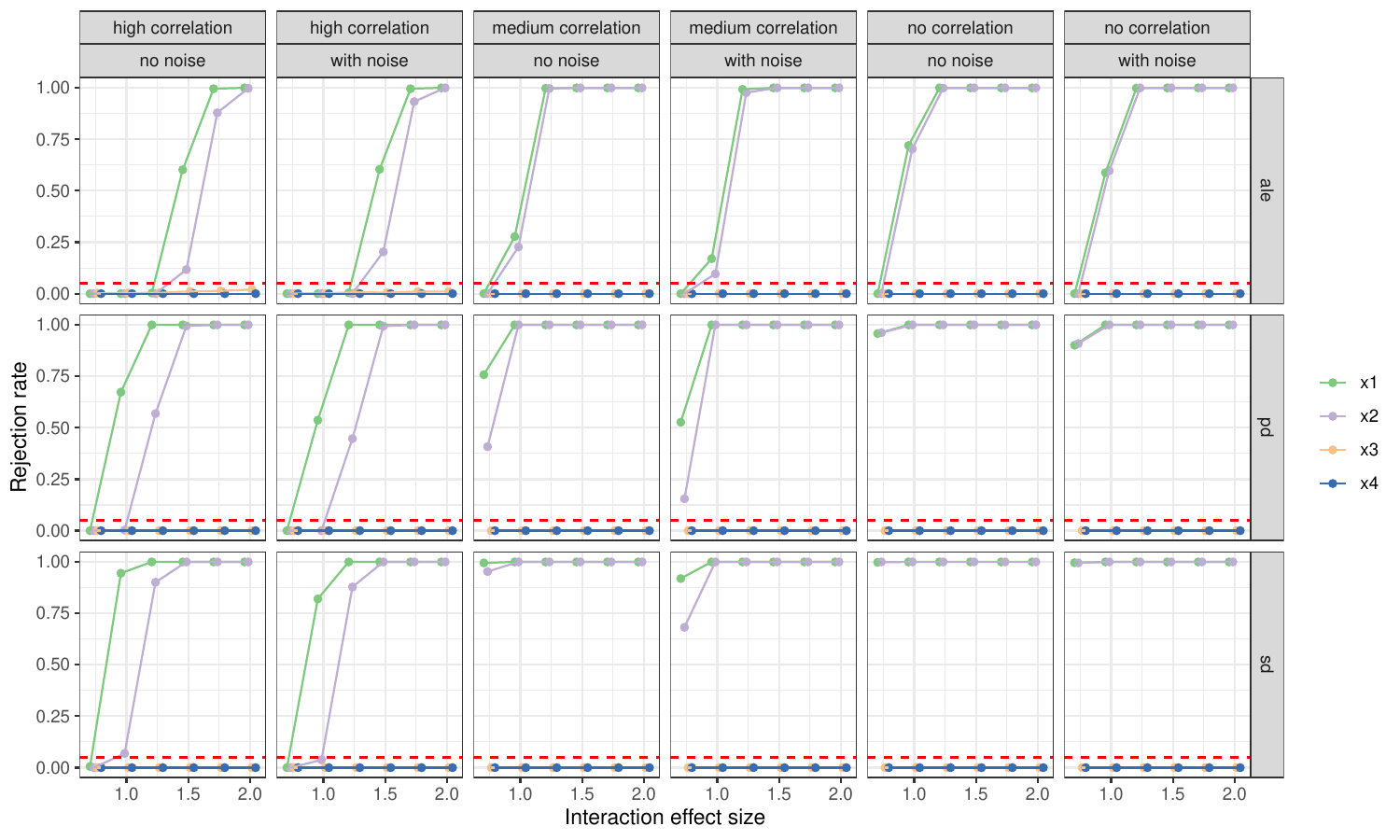}
    \caption{Rejection rates across all runs for the linear functional relationship and for each feature effect method shown in the rows. The columns are defined by the correlation between $\xv_2$ and $\xv_3$ and if a noise term is included or not. For each of these settings varying interaction effect sizes between $0.75$ and $2$ are considered.}
    \label{fig:rejection_lin}
\end{figure}

For each combination of the described settings, we perform $1000$ repetitions. The rejection rates over these repetitions for the different settings of the linear functional relationship are shown in Figure \ref{fig:rejection_lin}. It shows that when SD is used within PINT for the uncorrelated setting the type I and type II error are always $0$ independent of the interaction effect size. This can be similarly observed for PD, however, for ALE the interaction is not detected for $\beta = 0.75$. Also, the other settings show a higher power for PD and SD. A reason for this might be the definition of the intervals for ALE since the results for ALE are more sensitive concerning the size of the intervals than PD for the number of grid points. It is also observable that the power of the test is in most cases slightly higher for the settings without noise than with noise, however, the differences are rather small. In general, the interaction effect is harder to detect for smaller effect sizes if the correlation increases. The type I error does never exceed the significance level $\alpha = 0.05$, also not for $\xv_3$ which might have been considered in a spurious interaction with $\xv_1$ for the medium and high correlation settings.

\begin{figure}[h]
    \centering
     \includegraphics[width=\textwidth]{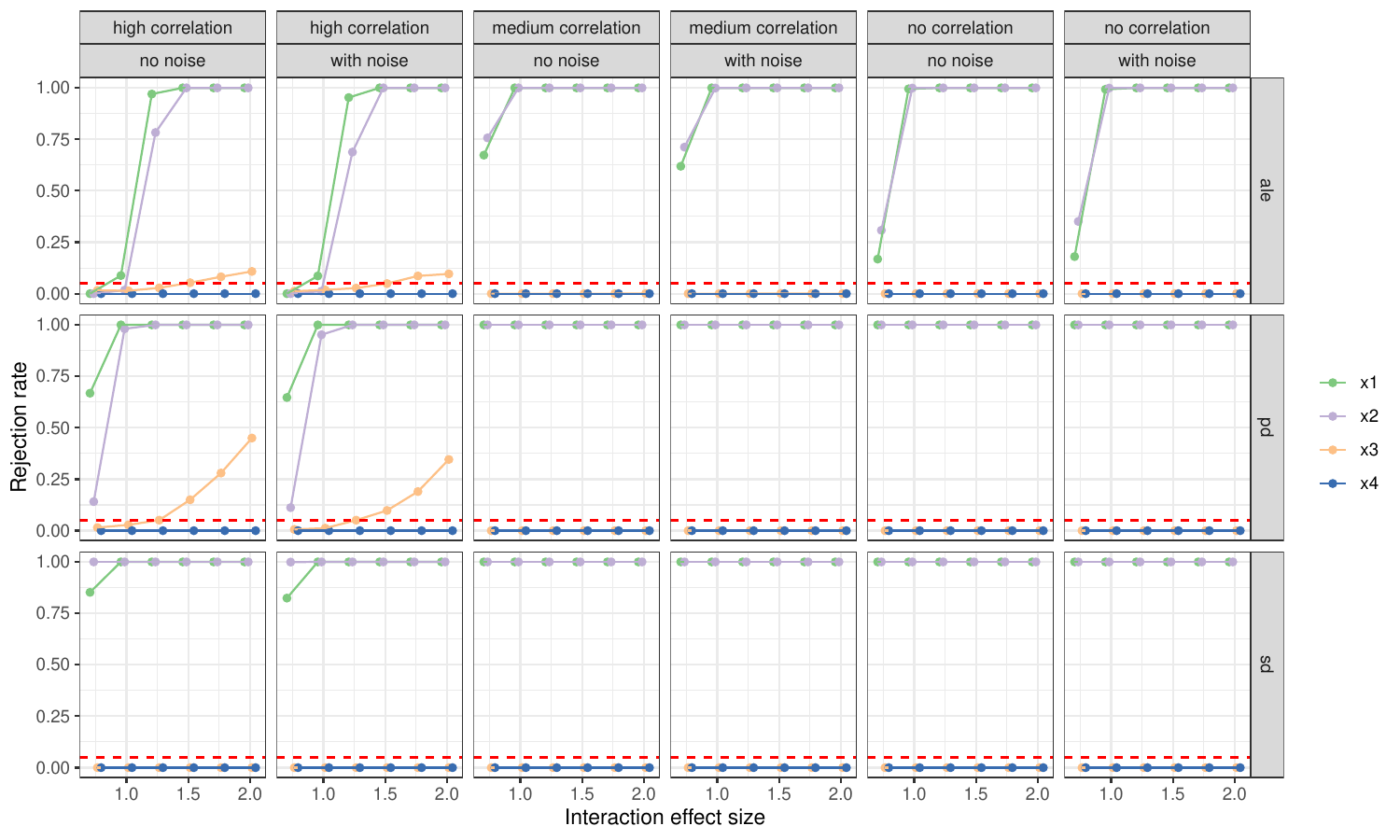}
    \caption{Rejection rates across all runs for the non-linear functional relationship and for each feature effect method shown in the rows. The columns are defined by the correlation between $\xv_2$ and $\xv_3$ and if a noise term is included or not. For each of these settings varying interaction effect sizes between $0.75$ and $2$ are considered.}
    \label{fig:rejection_nonlin}
\end{figure}

For the non-linear setting (see Figure \ref{fig:rejection_nonlin}), we can observe type I and II errors of $0$ for the settings without and with medium correlations if PD or SD is used within PINT for all interaction effect sizes. Compared to the linear setting, a type I error that exceeds the significance level is observable for the high correlation settings. The type I error is therefore particularly high for increasing effect sizes when PD is considered. Learning non-linear effects leads to more oscillating behavior when extrapolating into unseen regions of the feature space. As discussed and illustrated in Section \ref{sec:illustr_comp}, this problem is particularly severe for PD plots leading to high heterogeneity and, therefore, could also affect PINT.


\section{GADGET-SD Results for COMPAS}
\label{app:gadget_sd_compas}
\paragraph{\normalfont\textit{COMPAS Data Set.}}

In addition to the results of GADGET based on PD presented in Section \ref{sec:real_world}, we also applied GADGET with the same settings based on SD. 
\begin{figure}[h]
    \centering
    \includegraphics[width=\textwidth]{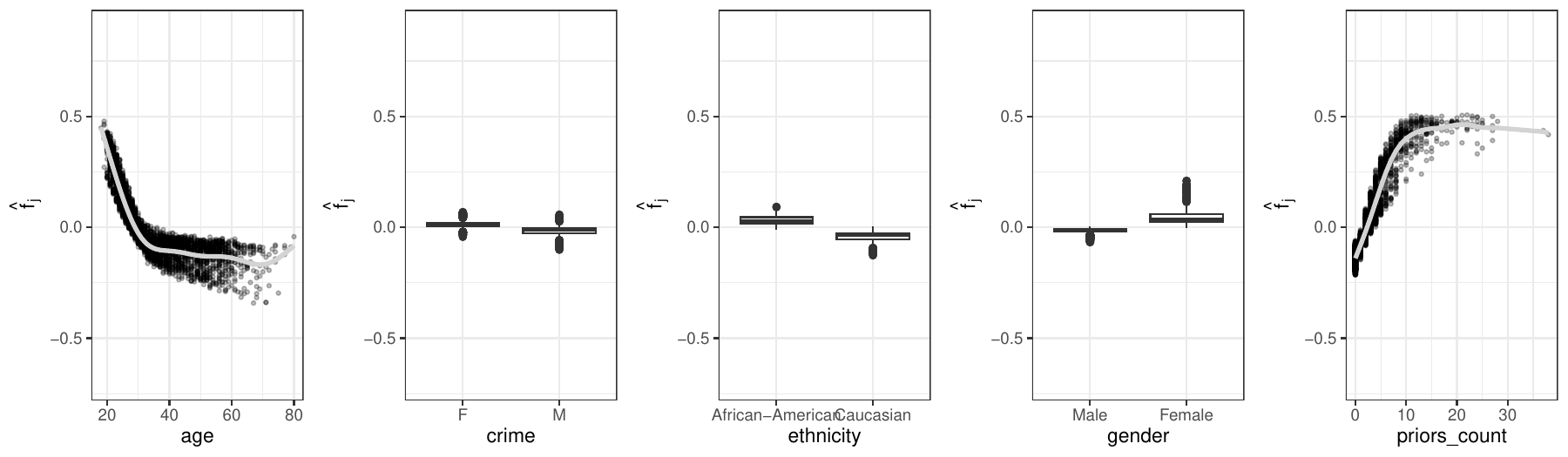}
    \caption{Global SD curves and Shapley values of considered features of the COMPAS application example.}
    \label{fig:compas_root_shap}
\end{figure}

\begin{figure}[h]
    \centering
    \includegraphics[width=\textwidth]{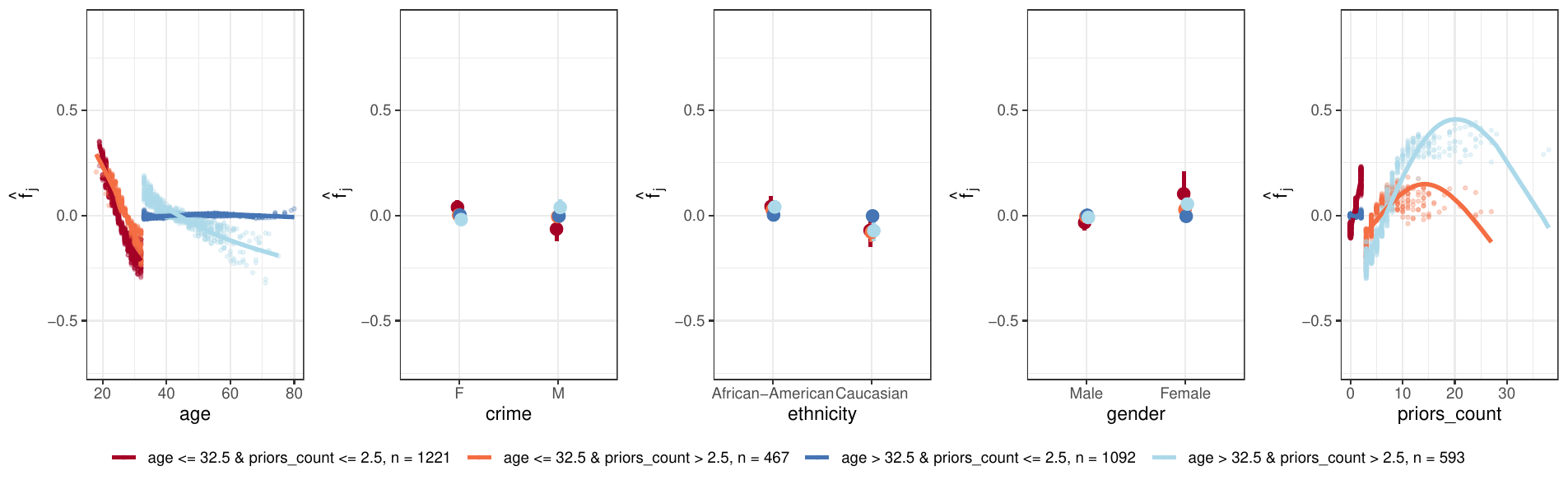}
    \caption{Regional SD plots for considered features of the COMPAS application after applying GADGET. Shapley values within each region are recalculated and visualize the interaction-related heterogeneity within each region.}
    \label{fig:compas_region_shap}
\end{figure}

The effect plots for the four resulting regions are shown in Figures \ref{fig:compas_root_shap} and
\ref{fig:compas_region_shap}. 
GADGET based on SD finds the same first split as for PD. The second split is also executed according to the number of prior crimes, but the split value is lower than for PD (at $2.5$ instead of $4.5$). The total interaction-related heterogeneity reduction ($R^2_{Tot} = 0.87$) is also similar. Note that Shapley values explain the difference between the actual and average prediction. Hence, SD plots are centered, while Figure \ref{fig:compas_region} shows the uncentered regional PD plots. 

\section{Further Details on Higher Dimensional Settings}
\label{app:higher_dim}

In Section \ref{sec:high_dim}, we consider the spam data set as an example of a high-dimensional data set. 
The data set contains $4601$ samples of e-mails of $57$ numeric features and a binary target (i.e., spam (1) or non-spam (0)). 
The features \textit{capitalLong}, \textit{capitalAve}, and \textit{capitalTotal} reflect the length of the longest, the average length and the total length of uninterrupted sequences of capital letters in an e-mail, respectively. The remaining $48$ features refer to the percentage of the respective word or character appearing within an e-mail. 

\begin{figure}[tbh]
    \centering
    \includegraphics[width=0.7\textwidth]{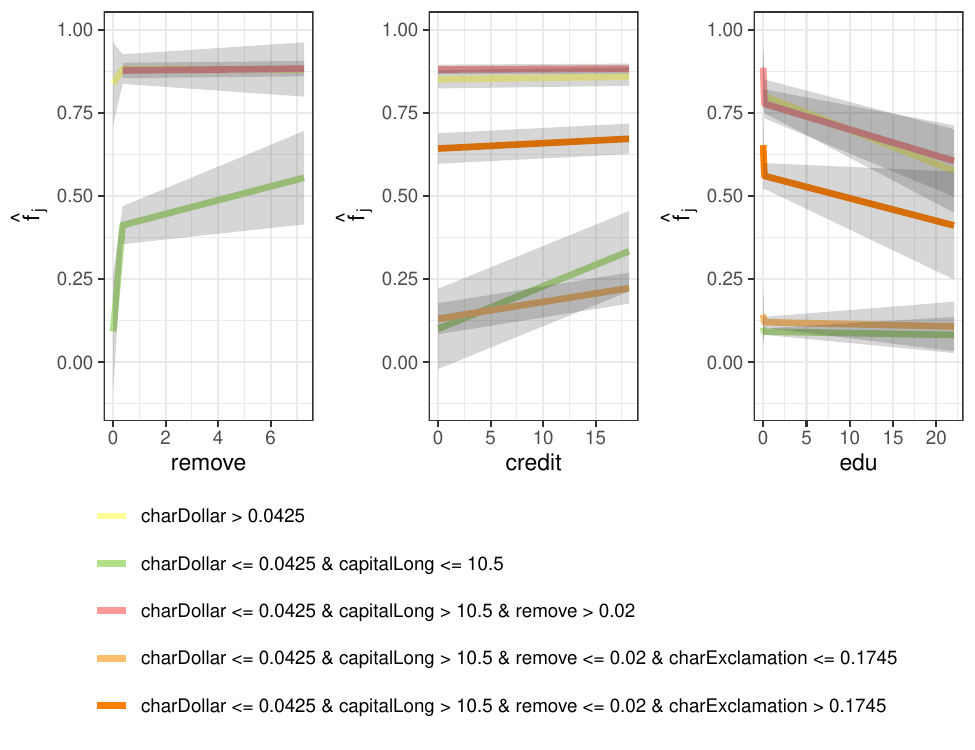}
     \vspace*{-0.7cm}
     \caption{Regional PD plots for \textit{remove}, \textit{credit} and \textit{edu} of the spam data set after applying GADGET when only interacting features are selected for $S$ and $Z$ according to the pre-filtering steps. The grey areas indicate the $95\%$ interaction-related heterogeneity interval as defined in Appendix \ref{app:overview_estimates}.}
     \label{fig:spam_filtered}
 \end{figure}

While we explained in Section \ref{sec:high_dim} the filtering procedure to handle high-dimensional data, we show here the results after applying the filtering steps and GADGET-PD on the spam data set.
Therefore, we apply GADGET on the $12$ remaining features after the filtering procedure described in Section \ref{sec:high_dim}. Figure \ref{fig:spam_filtered} shows the regional effects plots for $3$ of the $12$ features. The first split feature chosen by GADGET is \textit{charDollar} which indicates the percentage of the dollar sign within an e-mail. A high number is an indicator for spam e-mails. This region represented by the yellow curves is highly influenced by the number of occurrences of the word \textit{edu}. 
The combination of a lower number of dollar signs with small values for \textit{capitalLong}, decreases the probability of being classified as spam to most likely being classified as non-spam. However, in combination with a higher frequency of the words \textit{remove} or \textit{credit} the classification as spam becomes more likely.
For a small number of dollar signs but higher values of \textit{capitalLong}, also the frequencies of the word \textit{remove} and of \textit{charExclamation} are used to define the remaining three regions in Figure \ref{fig:spam_filtered}.

\section{Rashomon Effect}
\label{app:rashomon}
The Rashomon effect in ML states that there are often various models of a function class with almost equally good performance that are constructed differently \citep{breiman2001statistical, semenova2022existence, molnar2022pitfalls}. This set of models is called the Rashomon set. Since these models are constructed differently, they might identify different patterns in the data and rely on different features, which can lead to different results in the interpretation methods. For example, \cite{muller2023empirical} evaluate empirically, how the Rashomon effect influences different feature attribution methods. Their findings provide empirical support for both the disagreement problem of different interpretation methods and the sensitivity of these methods concerning their hyperparameter configurations. Hence, the Rashomon effect can also appear on the level of the interpretation method. A common example is seen in counterfactual explanations, where different counterfactuals may provide the same desired change in the outcome but based on different changes in the feature space. To address this issue, \citet{dandl2020multi} recommended to consider multiple counterfactuals instead of only one counterfactual.
Furthermore, the decomposition of the prediction function of a fitted model $\fh$ into effects of varying orders is typically not unique.
As a result, many different exact decompositions are possible and can lead to differing interpretations \citep{lengerich2020purifying}. One possibility to obtain a unique decomposition is the functional ANOVA decomposition. However, this decomposition might still differ across other models, as it is specifically calculated for one particular fitted model. 
Admittedly, the Rashomon effect remains an important unsolved problem in interpretable ML and a further discussion is outside of the scope of this work, but we refer the reader to \cite{fisher2019all} or \cite{donnelly2023rashomon} for potential approaches to deal with it.

\section{Marginal- and Conditional-based Feature Effect Methods}
\label{app:marg_vs_cond}
Feature effect methods can be grouped into approaches based on marginal or conditional distributions. While the PD plot is exemplary for the former and the M plot as well as the ALE plot are based on the latter, Shapley values can be defined either way.

While marginal-based approaches are usually simple to define and efficiently computable, they are known to extrapolate into unseen regions of the feature space when features are highly correlated, which might lead to misleading interpretations as illustrated in Section \ref{sec:illustr_comp} for PD plots \citep{hooker_generalized_2007, molnar2022pitfalls}. Conditional-based approaches solve the extrapolation problem by considering the true underlying data distribution \citep{aas_explaining_2021}. However, they require estimating the conditional data distribution, which is challenging, particularly in high-dimensional settings. The estimation influences the quality of the derived interpretation \citep{aas_explaining_2021, sundararajan_many_2020, janzing2020feature}. 
While both approaches come with different merits and weaknesses, the derived interpretation often differs as well. For example, PD plots visualize the influence of a feature of interest on the prediction function irrespective of the underlying feature correlations. In contrast, the M plot accounts for the correlation structure and thus also takes into account the influence of the features correlated with the feature of interest. Therefore, the two approaches answer different questions: 
Marginal-based approaches primarily offer insights into the inner workings of the ML model without explicitly considering the underlying data correlations, making them particularly useful for model audit and debugging.
Conditional-based approaches are used to derive interpretations that reflect the given data distribution and thus are preferable for questions related to model inference \citep{chen_true_2020, freiesleben2022scientific, watson2022conceptual}.

To conclude, \cite{chen_true_2020} argue that the marginal-based approach is not generally wrong and can be useful if we want to derive explanations that are true to the model, while the conditional-based approach should be used when we want to extract interpretations that are true to the data. 
The framework that we introduce in this paper 
works with both marginal-based (e.g., PD plots as illustrated in Section \ref{sec:pdp}) as well as conditional-based (e.g., ALE as shown in Section \ref{sec:ale}) approaches. Hence, the user needs to decide on a suitable feature effect method depending on their desired interpretation.

\vskip 0.2in
\bibliography{bib}

\end{document}